\pdfoutput = 1  
\documentclass[letterpaper, 10 pt, conference]{IEEEtran}  

\IEEEoverridecommandlockouts                              

\usepackage{graphics} 
\usepackage{epsfig} 
\usepackage{subfigure}
\usepackage{amsmath} 
\usepackage{amssymb}  
\usepackage{epstopdf}
\usepackage{hyperref}
\usepackage{lipsum,amsmath,multicol}
\usepackage{float}
\usepackage{algorithm}
\usepackage{algpseudocode}
\usepackage{wrapfig}
\usepackage{sidecap}

\usepackage{amsthm}
\newtheorem{theorem}{Theorem}
\usepackage{comment}
\usepackage{array}
\usepackage{glossaries}
\makeglossaries
\newcolumntype{L}[1]{>{\raggedright\let\newline\\\arraybackslash\hspace{0pt}}m{#1}}
\newcolumntype{C}[1]{>{\centering\let\newline\\\arraybackslash\hspace{0pt}}m{#1}}
\newcolumntype{R}[1]{>{\raggedleft\let\newline\\\arraybackslash\hspace{0pt}}m{#1}}

\allowdisplaybreaks
 
\begin{document}
%
\title{Solving Chance Constrained Optimization under Non-Parametric Uncertainty Through Hilbert Space Embedding.}
\author{Bharath Gopalakrishnan, Arun Kumar Singh, K.Madhava Krishna and Dinesh Manocha
}
\maketitle

\begin{abstract}
In this paper, we present an efficient algorithm for solving a class of chance constrained optimization under non-parametric uncertainty. Our algorithm is built on the possibility of representing arbitrary distributions  as functions in Reproducing Kernel Hilbert Space (RKHS). We use this foundation to formulate chance-constrained optimization as one of minimizing the distance between a desired distribution and the distribution of the constraint functions in the RKHS. We provide a systematic way of constructing the desired distribution based on a notion of scenario approximation. Furthermore, we use the kernel trick to  show that the computational complexity of our reformulated optimization problem is comparable to solving a deterministic variant of the chance-constrained optimization.
We validate our formulation on two important robotic/control applications: (i) reactive collision avoidance of mobile robots in uncertain dynamic environments and (ii) inverse dynamics based path tracking of manipulators under perception uncertainty. In both these applications, the underlying chance constraints are defined over highly non-linear and non-convex functions of the uncertain parameters and possibly also decision variables. We also benchmark our formulation with the existing approaches in terms of sample complexity and the achieved optimal cost  highlighting significant improvements in both these metrics.  

\end{abstract}

\section{Introduction}
Consider the following optimization problem in terms of a scalar variable $u$.

\begin{subequations}
\begin{align}
\min J(u)\label{cost}\\
P(f(\textbf{w}_1, \textbf{w}_2, u)\leq 0) \geq \eta\label{chance}\\
u\in \mathcal{C}\label{feasible_set}
\end{align}
\end{subequations}

where, $J(u)$ is a user defined cost function, $P(.)$ represents probability and $f(.)$ is the constraint function which depends on the decision variable $u$ and  uncertain parameters, $\textbf{w}_1, \textbf{w}_2$. The dependence of $f(.)$ on both $\textbf{w}_1, \textbf{w}_2$ and $u$ could possibly be highly non-linear and non-convex. The inequality (\ref{chance}) can be generalized to include any number of uncertain parameters and multiple chance constraints. Further, multiple optimization variables can also be accommodated. However, for easier exposition, we first restrict  our analysis to the simple case described above. Extensions to a more general case are straightforward and we discuss those later. The set $\mathcal{C}$ represents the feasible space of $u$ and is assumed to be convex for simplicity. Optimizations such as (\ref{cost})-(\ref{feasible_set}) are called chance-constrained optimizations and are used extensively for decision making under uncertainty. In robotics and control applications, they form the backbone of the robust Model Predictive Control (MPC) frameworks. For example, see  \cite{aerial_chance1}, \cite{chance_ad1}, \cite{chance_ad2}, \cite{chance_ad3}.

\newtheorem{remark}{Remark}

%
%
%
%
%

\begin{figure}[!t]
\centering
\includegraphics[width= 8.35cm, height=4.2cm] {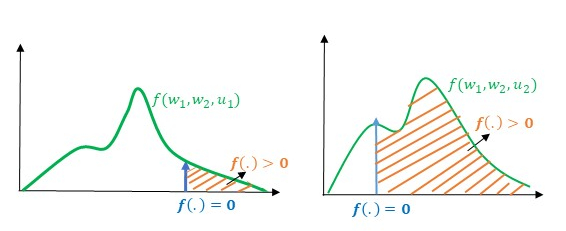}
\caption{An illustration of the observations made in Remark \ref{remark1}. The shape of the distribution can be manipulated by $u$. An appropriate shape is one where most of the mass lies to the left of $f(.) = 0$}
\label{intro_dist_fig}
\end{figure}

\begin{remark}\label{remark1}
At an intuitive level, chance-constrained optimizations can be interpreted as
a problem of ensuring that a specific portion of the mass of the distribution  $f(\textbf{w}_1, \textbf{w}_2, u )$ lie to the left of $f(.)=0$ (refer to Fig. \ref{intro_dist_fig}). For  given uncertain parameters $\textbf{w}_1, \textbf{w}_2$, the distribution is parametrized by the decision variable $u$ and can therefore be used to manipulate the location of a specified portion of its mass. However, each choice of $u$ incurs a cost $J(u)$.
\end{remark}

\begin{remark}\label{remark_eta}
The chance constraint probability $\eta$ has a direct correlation with the amount of mass of the distribution $f(\textbf{w}_1, \textbf{w}_2, u )$ lying to the left of $f(.)=0$. A Larger mass amounts to a higher $\eta$.
\end{remark}

Chance-constrained  optimizations are known to be very difficult to  solve.
The complexity increases further even when the uncertainty is non-parametric, that is, the analytical functional form of the probability distribution of $\textbf{w}_1, \textbf{w}_2$ are not known. Chance constraints are easy to solve when  $\textbf{w}_1$, $\textbf{w}_2$ are assumed to have a Gaussian distribution and the constraint function $f(.)$ is affine with respect to $u$ for given $\textbf{w}_1$, $\textbf{w}_2$ \cite{boyd_chance}, \cite{nemirovski}. However, in general, optimization problems where chance constraints are defined over non-linear and non-convex functions and the underlying uncertainty cannot be represented in any parametric form are known to be computationally intractable. Thus, various approximations and reformulations are proposed in existing literatures to tackle chance-constrained optimization problems.

\begin{table}[!h]
\centering
\caption{Important Symbols }
\begin{tabular}{|p{1.5cm}|p{5cm}|p{5cm}|p{5cm}|}\hline
$f(.)$ & Constraint function    \\ \hline
$\eta$ & Chance constraint probability  \\ \hline
$P_f(u)$    & Distribution of the chance constraints  \\ \hline
$P_f^d$     & Desired distribution\\ \hline
$\textbf{w}_1, \textbf{w}_2$ & uncertain parameters \\ \hline
$\textbf{w}_1^i, \textbf{w}_2^j$ & $i^{th}, j^{th}$ sample of uncertain parameters \\ \hline
${^i}\textbf{w}_2$ & $i^{th}$ variant of the uncertain parameter $\textbf{w}_2$. \\ \hline
$k(.,.)$ & Kernel function\\ \hline
$\mu_{P_f}$    & Kernel Mean of the distribution $P_f$\\ \hline
$\mu_{P_f^d}$    & Kernel Mean of the distribution $P_f^d$\\ \hline
$E[f(.)]$   & Expectation of a function $f(.)$ with respect to its random arguments \\ \hline
$Var[f(.)]$   & Variance of a function $f(.)$ with respect to its random arguments \\ \hline 
\end{tabular}
\end{table}

A popular approximation called the scenario approach \cite{scenario1}, \cite{scenario2} starts with drawing $n$ samples (or scenarios) of $\textbf{w}_1, \textbf{w}_2$ from their distribution and then replaces (\ref{chance}) with $n^2$ constraints of the form $f_i(\textbf{w}_1^i, \textbf{w}_2^j, u)\leq 0, \forall i, j$. The scenario approach has a very interesting set of pros and cons. On the one hand, it is conceptually simple and is applicable even when the parametric form of the distribution of uncertain parameters is not known and just their samples are given. On the other hand, the naive implementation of the scenario approach is known to be overly conservative. To be precise, the cost $J(.)$ increases with $n$, although the solution becomes more robust at the same time. Works like \cite{scenario_rejection1} provide algorithms for rejection sampling to reduce the conservativeness of the scenario approach.

An alternate class of approach relies on replacing chance constraints with a deterministic surrogate \cite{boyd_chance}, \cite{musigma1}, \cite{sample_average_shapiro}. For example, (\ref{saa}) represents the robust variant of the so called \emph{sample average approximation} (SAA) \cite{sample_average_shapiro}, where, $\textbf{w}_1^i, \textbf{w}_2^j $ represents the $i^{th}, j^{th}$ samples of $\textbf{w}_1, \textbf{w}_2 $ and  $\mathcal{I}_f$ represents an indicator function. The variable $\gamma$ is similar but not necessarily the same as the chance constraint probability $\eta$. A strong advantage of SAA (\ref{saa}) is that it provides a very tight approximation resulting in a low cost solution. However, (\ref{saa}) represents an extremely difficult non-smooth and non-convex constraint. Thus, the reformulated chance-constrained optimization itself becomes very difficult. Our experimentation has shown that it is possible to solve SAA based reformulations  of single variable chance constrained optimization with an exhaustive search. However, such an approach is unlikely to scale to problems with multiple decision variables.

\begin{equation}
P(f(\textbf{w}_1^i,\textbf{w}_2^j), u)\geq \eta \approx \frac{1}{n^2}\sum_i\sum_j \mathcal{I}_f \geq \gamma
\label{saa}
\end{equation}

\begin{equation}\nonumber
\mathcal{I}_f = \begin{cases}
    1, & \text{if $f(\textbf{w}_1^i, \textbf{w}_2^j, u)\leq 0$}.\\
    0, & \text{otherwise}.
  \end{cases}
 \end{equation}
 
 A simpler surrogate constraint \ref{meanvar} proposed in \cite{mean_var_approach} has been used in works like \cite{musigma1}, \cite{musigma2}, \cite{gaussian_chance}.

\begin{equation}
E[f(\textbf{w}_1, \textbf{w}_2, u)]+\epsilon \sqrt{Var[f(\textbf{w}_1, \textbf{w}_2, u)]}\leq 0, \epsilon >0.
\label{meanvar}
\end{equation}

\noindent where, $E[.]$, $Var[.]$  represent the mean and variance of $f(.)$,  taken with respect to random variables $\textbf{w}_1, \textbf{w}_2$. Using, Cantelli's inequality, it can be shown that the satisfaction of (\ref{meanvar}) ensures that chance constraints are satisfied with $\eta\geq \frac{\epsilon}{1+\epsilon^2}$. However, it should be noted that this bound can be rather loose. The attractive feature of (\ref{meanvar}) is that  it is applicable for a wide class of chance constraints. However, its efficiency is predicated on how easy it is to compute  analytical expressions for $E[.]$ and $Var[.]$ . For example, if $f(.)$ is highly non-linear or/and the parametric form of $\textbf{w}_1, \textbf{w}_2$ is not known, then  computing an accurate analytical expression for $E[.]$ and $Var[.]$ becomes a very challenging problem. A workaround has been proposed in works like \cite{musigma2}, \cite{chance_multi} \cite{chance_surrogate} where the analytical expressions for $E[.]$ and $Var[.]$ are approximated through Monte Carlo sampling. However we should note two key bottlenecks of such approaches . First, the sample complexity is poor and our experimentation shows that it usually requires around $10^6$ samples to get to a reasonable approximation. Second, for a given sample size, it is difficult to estimate how well the surrogate constraints are approximated. This in turn, makes it difficult to accurately infer the feasibility of the original chance constraints.

\subsection{Contributions}
\begin{figure}[!t]
\centering
\includegraphics[width= 8.35cm, height=6.2cm] {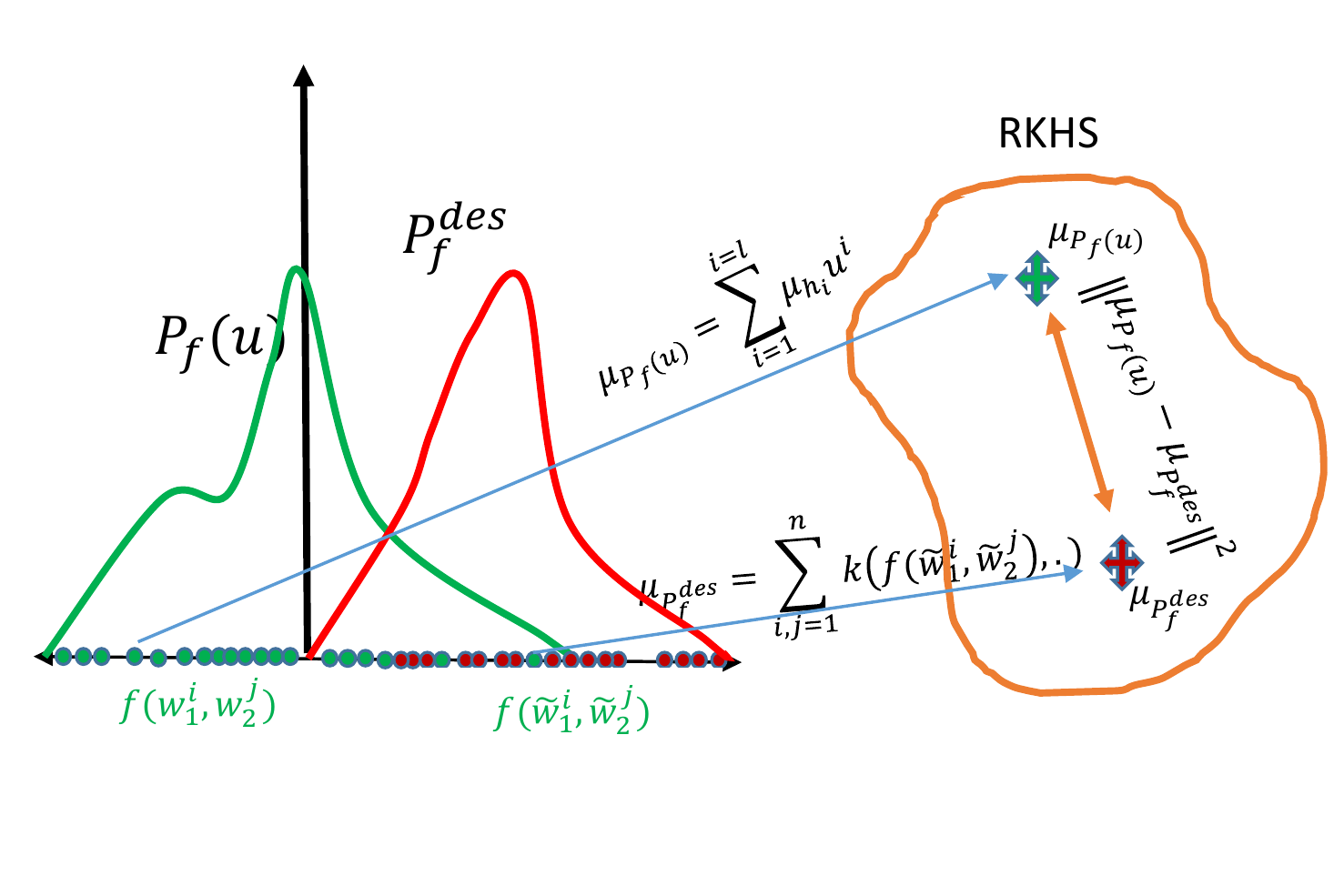}
\caption{Pdfs in the physical space can be represented as functions (or points) in RKHS.}
\label{intro_fig}
\end{figure}

\noindent In this paper, we present a novel approach built on the fact that any arbitrary distribution ($f(\textbf{w}_1, \textbf{w}_2, u)$ in our case) can be embedded as a function (or a point, refer to Fig.~\ref{intro_fig}) in Reproducing Kernel Hilbert Space (RKHS). The embedded function is generally referred to as the Kernel Mean and thus RKHS embedding is also known as Kernel Mean Embedding (KME) in the existing literature \cite{scholkopf}. A few key advantages of RKHS embedding are worth pointing out. First, the embedding can be achieved even when the parametric form of the underlying uncertainty ($\textbf{w}_1, \textbf{w}_2$ in our case) is not known. Second, the embedding only requires  point-wise evaluations of $f(.)$ and is thus not influenced by its algebraic complexity. Finally, it opens avenues for the use of established \emph{reduced set methods} to achieve a good sample complexity. Intuitively, \emph{reduced set methods} provides a systematic way of choosing a subset of samples while still retaining as much information as possible from the original sample size by re-weighting the importance of those samples.

In the current work, we build on the concept of RKHS embedding and put forward the following contributions:

\begin{itemize}
\item We interpret chance-constrained optimization as a problem of matching  higher order moments of two given distributions. The two distributions in consideration are the distribution of the constraint functions and a certain "desired distribution", which we show, can be systematically constructed, borrowing notions from scenario approximation. Although conceptually simple, to the best of our knowledge, there are no other works based on this interpretation.

\item We further reformulate moment matching as a problem of minimizing the distance between the RKHS embeddings of the constraint function and the desired distribution. We use the so called Kernel trick to show that the complexity of the RKHS embedding based reformulation is comparable to solving a deterministic variant of the chance-constrained optimization (\ref{cost}-\ref{feasible_set}), obtained by replacing (\ref{chance}) with a single deterministic constraint of the form $f(\textbf{w}_1, \textbf{w}_2, u)\leq 0$. To be precise, if $f(.)$ is polynomial in $u$ of order $l$, then the reformulated problem is also a polynomial optimization problem of order $2l$. 

\item  We benchmark  our formulation with  the existing approaches based on two metrics :  sample complexity and obtained optimal cost. In particular, we highlight the following results: First, we show that our formulation significantly outperforms scenario approximation in both the metrics. Second, our formulation and the SAA approach based on surrogate constraints (\ref{saa}) results in similar optimal cost. However, our formulation leads to a simpler optimization problem and enjoy better sample complexity. Finally, our formulation also outperforms approaches based on surrogate constraints (\ref{meanvar}).

\item We apply our formulation on two challenging motion planning/control problems. The first problem involves navigating a mobile robot in dynamic and uncertain environments. Herein, we consider noise arising from both perception and ego-motion during formulating  the chance constraints for ensuring probabilistic collision avoidance. Our second problem implements a stochastic variant of inverse dynamics based path tracking for manipulators. We assume that the concerned manipulator has noise-less motions but noisy state estimation. Consequently, the manipulator should compute the necessary torque commands while considering the state estimation uncertainty to ensure that the probability of exerting a torque that violates the specified bounds is under some threshold. This requirement can be naturally put in the form of chance constraints.

\end{itemize}

\noindent 

\section{Embedding Distribution in RKHS}

\subsection{RKHS}
\noindent RKHS is a Hilbert space with a positive definite function $k(.) : \Re^n \times \Re^N \rightarrow \Re$ called the Kernel. Let, $\textbf{x}$ denote an observation in physical space (say Euclidean). It is possible to embed this observation in the RKHS by defining the following kernel based function whose first argument is always fixed at $\textbf{x}$.

\begin{equation}
\phi(\textbf{x}) = k(\textbf{x},.)
\end{equation}

An attractive feature of RKHS is that it is  endowed with an inner product which, in turn, can be used to model the distance between two functions in RKHS. Furthermore, the distance can be formulated in terms of kernel function in the following manner

\begin{equation}
\langle\phi(\textbf{x}_i)\phi(\textbf{x}_j)\rangle = k(\textbf{x}_i, \textbf{x}_j)
\label{kernel_trick}
\end{equation}

\noindent Equation (\ref{kernel_trick}) is called the "kernel trick" and its strength lies in the fact that the inner product can be computed by only point wise evaluation of the kernel function.

\subsection{Distribution Embedding}
\noindent The projection to RKHS can also be generalized to distributions. Let, $\textbf{w}^1$, $\textbf{w}^2$, $\textbf{w}^3$.....$\textbf{w}^n$ be samples drawn from an unknown probability distribution $P$. This distribution $P$ can be represented in the RKHS through a function called the Kernel Mean, which is described in the following manner

\begin{equation}
\mu_P[\textbf{w}] = \sum_{j=1}^{n}\alpha_jk(\textbf{w}^j,.)
\label{kernel_mean_samples}
\end{equation}

\noindent where, $\alpha_j$ is the weight associated with $\textbf{w}^j$. For example, if the samples are i.i.d then, $\alpha_j=\frac{1}{n}, \forall j$. The estimator (\ref{kernel_mean_samples}) is consistent, i,e, the estimation improves as the number of samples increases.

Following \cite{scholkopf},  equation (\ref{kernel_mean_samples}) can also be used to embed functions of random variables like $f(\textbf{w}_1, \textbf{w}_2, u)$.

\begin{equation}
\mu_f(u) =  \sum_{i=1}^{n}\sum_{j=1}^{n}\alpha_i\beta_j k(f(\textbf{w}_1^i, \textbf{w}_2^j, u),.)
\label{kernel_mean_chance}
\end{equation} 

\noindent An important thing to note from (\ref{kernel_mean_chance}) is that for given samples of $\textbf{w}_1$, $\textbf{w}_2$, the Kernel Mean given by (\ref{kernel_mean_chance}) is  dependent on the variable $u$.


\subsection{Maximum Mean Discrepancy (MMD)}\label{MMD}
\noindent Given two distributions $P$, $Q$, MMD refers to the distance between their RKHS embeddings $\mu_{P}$, $\mu_Q$. That is:

\begin{eqnarray}
MMD: \Vert \mu_P-\mu_Q \Vert^2 = \langle \mu_P-\mu_Q\rangle \nonumber \\ = \langle \mu_P, \mu_P\rangle -2\langle\mu_P, \mu_Q\rangle+\langle \mu_Q, \mu_Q\rangle \nonumber\\
= \sum_{i=1}^{i=n}\sum_{j=1}^{j=n}\alpha_i\alpha_j k(\textbf{w}^i_1,\textbf{w}^j_1)-2\sum_{i=1}^{i=n}\sum_{j=1}^{j=n}\alpha_i\beta_j k(\textbf{w}^i_1,\textbf{w}^j_2) \nonumber  \\
+\sum_{i=1}^{i=n}\sum_{j=1}^{j=n} \beta_i\beta_j k(\textbf{w}^i_2,\textbf{w}^j_2)\label{mmd}
\end{eqnarray}

\noindent An important thing to note from (\ref{mmd}) is how the kernel trick allows us to express MMD only in terms of the point-wise evaluation of the kernel function.

\subsection{Reduced Set Methods}\label{reduced_set}
\noindent Consider a vector  described in terms of weighted combination of basis functions. Reduced set methods are a class of algorithms which allows us to compute an optimal approximation of the vector using a highly reduced number of basis functions \cite{scholkopf3}. Interestingly, the same class of algorithms can be applied to improve the sample complexity of RKHS embedding as well. The process can be described as follows. Let $\hat{\textbf{w}}_1^1, \hat{\textbf{w}}_1^2..\textbf{w}_1^N$ and $\hat{\textbf{w}}_2^1, \hat{\textbf{w}}_2^2..\textbf{w}_2^N$  represent $N$ i.i.d samples of $\textbf{w}_1$, $\textbf{w}_2$ respectively. Further, let  $\textbf{w}_1^1, \textbf{w}_1^2..\textbf{w}_1^n$ and $\textbf{w}_2^1, \textbf{w}_2^2..\textbf{w}_2^n$ represent a subset (reduced set) of the i.i.d samples. It is implied that $n<<N$. Now, intuitively, a reduced set method would re-weight the importance of each sample from the reduced set such that they retain as much as information of the original i.i.d samples. The weights $\alpha_i$, $\beta_i$ associated with $\textbf{w}_1^i$ and $\textbf{w}_2^i$  are computed through the following optimization problems. 

\small
\begin{equation}
\alpha_i = \arg\min\Vert \frac{1}{N}\sum_{i=1}^{i=N}k(\hat{\textbf{w}_1}^i,.)-\sum_{i=1}^{i=n}\alpha_ik(\textbf{w}_1^i,.)\Vert_2, s.t \sum \alpha_i =1 
\end{equation}
\vspace{-0.5cm}
\begin{equation}
\beta_i = \arg\min\Vert \frac{1}{N}\sum_{i=1}^{i=N}k(\hat{\textbf{w}_2}^i,.)-\sum_{i=1}^{i=n}\beta_ik(\textbf{w}_2^i,.)\Vert_2, s.t \sum \beta_i =1 
\end{equation}
\normalsize

\section{Main Results}
In this section, we derive our main result which is a reformulation of chance-constrained optimization (\ref{cost})-(\ref{feasible_set}) into a much simpler minimization problem. The following are our key assumptions:

\begin{itemize}
\item We assume that the uncertainty is non-parametric, which in our case means that the probability distribution functions associated with $\textbf{w}_1, \textbf{w}_2$ are not known. Rather, we have access to their $n$ discrete samples. These samples could come from a simulator which mimics a very generalized distribution with arbitrary order of moments.

\item We assume that the analytical form for the constraint functions are known.

\end{itemize} 

\subsection{Algebraic Form of the Constraint Function}

\noindent In this paper, we consider the  chance constraints defined over the following class of  constraint functions.

\begin{equation}
f(\textbf{w}_1, \textbf{w}_2, u) = \sum_{i=0}^{l}h_i(\textbf{w}_1, \textbf{w}_2)u^i
\label{general_chance}
\end{equation}

\noindent where, $h_i(\textbf{w}_1, \textbf{w}_2), \Re^n\rightarrow R$ is a generic possibly non-linear function of $\textbf{w}_1, \textbf{w}_2$, while $u^i$ represents a monomial of order $i$. The definition (\ref{general_chance}) is very general and has the famous affine class of chance constraints as a special case with $l=1$ and $h_0(\textbf{w}_1, \textbf{w}_2)=0$, $h_1(\textbf{w}_1, \textbf{w}_2) = \textbf{w}_1$. It can be seen that even if the uncertain parameters, $\textbf{w}_1$, $\textbf{w}_2$ are Gaussian, the chance constraints defined over $f(\textbf{w}_1, \textbf{w}_2, u)$  may still be too complex to get an analytical characterization for the distribution of $f(\textbf{w}_1, \textbf{w}_2, u)$.

Let, $P_f(u)$ represent the distribution of $f(\textbf{w}_1, \textbf{w}_2,u)$parametrized in terms of $u$. Its RKHS embedding can be computed using (\ref{kernel_mean_chance}) in the following manner:

\begin{equation}
\mu_{P_f}(u) = \sum_{i=0}^{i=l}\mu_{h_i}u^i
\label{rkhs_chance1}
\end{equation}

\begin{equation}
\mu_{h_i} = \sum_{i=1}^{i=n}\sum_{j=1}^{j=n}\alpha_i\beta_jk(h_i(\textbf{w}_1^i, \textbf{w}_2^j),.)
\label{rkhs_chance2}
\end{equation}

\subsection{Desired Distribution}
\noindent The notion of desired distribution is derived from the observations made in Remark \ref{remark1}. To recap, we want to ensure that the distribution $f(\textbf{w}_1, \textbf{w}_2, u )$ achieves an appropriate shape. To this end, desired distribution acts as a benchmark for $f(\textbf{w}_1, \textbf{w}_2, u )$; in other words, a distribution that $f(\textbf{w}_1, \textbf{w}_2, u )$ should resemble as closely as possible  for an appropriately chosen $u$. We formalize the notion of desired distribution with the help of the following definitions:

\newtheorem{definition}{Definition}

\begin{definition}\label{def_nominal_sol}
$u_{nom}$ refers to any solution of the optimization (\ref{cost})-(\ref{feasible_set}) that is associated with a low optimal cost $J({u}_{nom})$. 
\end{definition}

\begin{definition}
Let $\widetilde{\textbf{w}}_1$, $\widetilde{\textbf{w}}_2$ be random variables which represent the same entity as $\textbf{w}_1, \textbf{w}_2$ but belong to some known distributions $P_{\textbf{w}_1}^{des}$, $P_{\textbf{w}_2}^{des}$. Further, when  $\widetilde{\textbf{w}}_1 \approx P^{des}_{\textbf{w}_1} $ and $\widetilde{\textbf{w}}_2,\approx P^{des}_{\textbf{w}_2} $, then,  $f(\widetilde{\textbf{w}}_1, \widetilde{\textbf{w}}_2, u_{nom})\approx P^{des}_{f}$. In such a case, $P^{des}_{f}$ is called the desired distribution if the following holds:

\begin{equation}
P(f(\widetilde{\textbf{w}}_1, \widetilde{\textbf{w}}_2, u_{nom})\leq 0)\approx 1.0, \widetilde{\textbf{w}}_1\approx P_{\textbf{w}_1}^{des}, \widetilde{\textbf{w}}_2\approx P_{\textbf{w}_2}^{des}
\label{desired_chance}
\end{equation}
\end{definition}

Equation (\ref{desired_chance}) suggests that if the uncertain parameters belong to the distribution $P^{des}_{\textbf{w}_1}$, $P^{des}_{\textbf{w}_2}$, then the entire mass of the distribution, $f(\widetilde{\textbf{w}}_1, \widetilde{\textbf{w}}_2, u)$ can be manipulated to lie almost completely to the left of $f(.)=0$ by choosing $u=u_{nom}$.  This setting represents an ideal case because we have constructed uncertainties appropriately, so that we can manipulate the distribution of the chance constraints while incurring a
nominal cost.

\noindent {\textbf{Constructing the Desired Distribution}}:

\noindent We now describe how distributions $P_{\textbf{w}_1}^{des}$, $P_{\textbf{w}_2}^{des}$ and $P^{des}_{f}$ can be constructed. While exact computations may be intractable, in this section, we provide a simple way of constructing an approximate estimate of these distributions. The basic procedure is as follows.

Given $n$ samples of $\textbf{w}_1, \textbf{w}_2$ we construct two sets  $\mathcal{C}_{\widetilde{\textbf{w}_1}}$, $\mathcal{C}_{\widetilde{\textbf{w}_2}}$ respectively containing $n_{\textbf{w}_1}$ samples of $\textbf{w}_1$ and $n_{\textbf{w}_2}$ samples of $\textbf{w}_2$. For clarity of exposition, we choose $\widetilde{\textbf{w}}_1$, $\widetilde{\textbf{w}}_2$ to identify samples from set $\mathcal{C}_{\widetilde{\textbf{w}_1}}$, $\mathcal{C}_{\widetilde{\textbf{w}_2}}$. Now, assume that the following holds.

\begin{equation}
f(\widetilde{\textbf{w}}_1^i, \widetilde{\textbf{w}}_2^j, u_{nom})\leq 0, \forall \widetilde{\textbf{w}}_1^i\in \mathcal{C}_{\widetilde{\textbf{w}_1}},  \widetilde{\textbf{w}}_2^j\in \mathcal{C}_{\widetilde{\textbf{w}_2}}
\label{desired_sample_distri}
\end{equation}  

By comparing (\ref{desired_chance}) and (\ref{desired_sample_distri}), it can be inferred that the sets $\mathcal{C}_{\widetilde{\textbf{w}_1}}$, $\mathcal{C}_{\widetilde{\textbf{w}_2}}$ are in fact sample approximations of the distributions $P_{\textbf{w}_1}^{des}$ and $P_{\textbf{w}_2}^{des}$ respectively. Furthermore, a set $\mathcal{C}_f$ containing $n_{\textbf{w}_1}*n_{\textbf{w}_2}$ samples of $f(\widetilde{\textbf{w}}_1^i, \widetilde{\textbf{w}}_2^j, u_{nom})$ can  be taken as the sample approximation of the desired  distribution $P^{des}_{f}$.

One last piece of puzzle remains. We still do not know, however which $n_{\textbf{w}_1}$ samples of $\textbf{w}_1$ and $n_{\textbf{w}_2}$ samples of $\textbf{w}_2$ should be chosen to construct sets $\mathcal{C}_{\widetilde{\textbf{w}_1}}$, $\mathcal{C}_{\widetilde{\textbf{w}_2}}$. 
In particular, we need to ensure that the assumption (\ref{desired_sample_distri}) holds for the chosen samples. To this end, we follow the following process. We arbitrarily choose $n_{\textbf{w}_1}$ samples of $\textbf{w}_1$ and $n_{\textbf{w}_2}$ samples of $\textbf{w}_2$ and correspondingly obtain a suitable $u_{nom}$ as a solution to the following optimization problem:

\begin{subequations}
\begin{align}
u_{nom} = \arg\min J(u)\label{cost_scenario}\\
f(\widetilde{\textbf{w}}_1^i, \widetilde{\textbf{w}}_2^j, u)\leq 0, \forall i = {1,2..n_{\textbf{w}_1}},j = {1,2..n_{\textbf{w}_2}} \label{scenario_const}\\
u\in \mathcal{C}\label{feasible_scenario}
\end{align}
\end{subequations}

Note that satisfaction of (\ref{scenario_const}) ensures that the assumption (\ref{desired_sample_distri}) holds. Few points are worth noting about the above optimization. First, it is a deterministic problem whose complexity  primarily depends on the algebraic nature of $f(.)$. Second, the desired distribution  can always be constructed if we have access to sets $\mathcal{C}_{\widetilde{\textbf{w}_1}}$, $\mathcal{C}_{\widetilde{\textbf{w}_2}}$. The construction of these two sets is guaranteed as long as we can obtain a feasible solution to (\ref{cost_scenario})-(\ref{feasible_scenario}). Third, the computational burden of solving the optimization problem can be significantly reduced by some clever sampling. For example, in our implementation, we compute the left hand side of (\ref{scenario_const}) for different combination of samples and then choose the set which leads to the least violation of the constraints (\ref{scenario_const}). Finally, (\ref{cost_scenario})-(\ref{feasible_scenario}) is precisely the so-called scenario approximation for chance constrained optimization (\ref{cost})-(\ref{feasible_set}). Conventionally, scenario approximation is solved with a large $n_{\textbf{w}_1}, n_{\textbf{w}_2}$ (typically $10^4$ ) in order to obtain a solution that satisfy chance constraints (\ref{chance}) with a high $\eta$ ($\approx 0.90$). In contrast, we use (\ref{cost_scenario})-(\ref{feasible_scenario}) to estimate the desired distribution and thus for our purpose, a small sample size in the range of $n_{\textbf{w}_1}=n_{\textbf{w}_2} \approx 20$ proves to be sufficient in practice.

The RKHS embedding of these distributions can be obtained in the following manner:

\begin{equation}
\mu_{P^{des}_{{\textbf{w}}_1}} = \sum_{i=1}^{i=n_{\textbf{w}_1}}\lambda_i k(\widetilde{\textbf{w}}^i_1,.), \hspace{0.2cm} \widetilde{\textbf{w}}_1^i \in \mathcal{C}_{\widetilde{\textbf{w}}_1}
\label{rkhs_desired1}
\end{equation}

\begin{equation}
\mu_{P^{des}_{\textbf{w}_2}} = \sum_{i=1}^{i=n_{\textbf{w}_2}} \xi_i k(\widetilde{\textbf{w}}^i_2.,), \hspace{0.2cm} \widetilde{\textbf{w}}_2^i \in \mathcal{C}_{\widetilde{\textbf{w}}_2}
\label{rkhs_desired2}
\end{equation}

\small
\begin{equation}
\mu_{P^{des}_f} = \sum_{i=1}^{i=n_{\textbf{w}_1}}\sum_{j=1}^{j=n_{\textbf{w}_2}}\lambda_i \xi_j k(f(\widetilde{\textbf{w}}^i_1, \widetilde{\textbf{w}}_2^j,u_{nom}),.),\hspace{-0.1cm}
\widetilde{\textbf{w}}_1^i, \widetilde{\textbf{w}}_2^j \in \mathcal{C}_{\widetilde{\textbf{w}}_1},\mathcal{C}_{\widetilde{\textbf{w}}_2}
\label{rkhs_desired3}
\end{equation}
\normalsize

\noindent Where, $\lambda_i, \xi_j$ are constants derived from the reduced set methods described in Section \ref{reduced_set}.

\subsection{Chance-Constrained Optimization as a Moment Matching Problem}
In this section, we reformulate the chance-constrained optimization (\ref{cost})-(\ref{feasible_set}) as a moment matching problem. Our key idea builds upon the following theorem from \cite{moment_bound_lindsay}.

\begin{theorem}\label{th_moment_bound}
 $\Vert P_f(u)- P_f^{des}\Vert \leq B(d)$, $B(d)\rightarrow 0$, $d\rightarrow \infty$ 
\end{theorem}

\noindent where, $d$ refers to the order upto  which the moments of $P_f(u)$ and $P_f^{des}$ are similar. The above theorem suggests that the difference between two distributions can be bounded by a non-negative function $B(d)$ which decreases with an increasing order of moment $d$. Authors in \cite{moment_bound_lindsay} also show that this bound is particularly tight near the tail end of the distribution. Now, recalling that  almost the entire mass of $P_f^{des}$ lies to the left of $f(.)=0$, it is clear that as we make the tail of  $P_f^{des}$ and  $P_f(u)$ similar by matching higher order moments, we ensure that more and more of the mass of $P_f(u)$ gets shifted to the left of $f(.)=0$. This, leads to the satisfaction of chance constraints (\ref{chance}) with a higher $\eta$. Theorem \ref{th_moment_bound} lays the foundation for the following optimization problem which can  act as a substitute for the original chance-constrained optimization (\ref{cost})-(\ref{feasible_set}).

\begin{subequations}
\begin{align}
 \arg\min \rho_1 \mathcal{L}_{mom}(P_f(u), P_f^d, d)+\rho_2 J(u)\label{cost_mom_matching}\\
u\in \mathcal{C}\label{feasible_mom_matching}
\end{align}
\end{subequations}

\noindent where, $\mathcal{L}_{mom}(.)$ is a cost function that measures the similarity between the first $d$ moments of $P_f(u)$ and $P_f^{des}$. A low value of $\mathcal{L}_{mom}$ would imply that the first $d$ moments of $P_f(u)$ and $P_f^{des}$ are very similar.

\noindent \textbf{Accommodating Chance Constraint Probability $\eta$:}  Optimization (\ref{cost_mom_matching})-(\ref{feasible_mom_matching}),  accommodates the chance constraint probability $\eta$ in an implicit manner. Thus, the process of obtaining  solutions with different level of  robustness based on $\eta$ is more indirect and involved than the original optimization (\ref{cost})-(\ref{feasible_set}). In (\ref{cost_mom_matching})-(\ref{feasible_mom_matching}), the similarity between the tail of $P_f(u)$ and $P_f^{des}$  not only depends on the residual of $\mathcal{L}_{mom}(.)$ but also on the moment order $d$ used to construct $\mathcal{L}_{mom}(.)$.  Fixing weights $\rho_1$ and $\rho_2$ and increasing $d$ increases the similarity near the tail end and  thus leads to the satisfaction of chance constraints with higher $\eta$. A similar goal can be achieved by fixing $d$ and $\rho_2$ and increasing $\rho_1$.

\subsection{ Reformulating Distribution/Moment Matching through RKHS Embedding} 
\noindent The optimization  (\ref{cost_mom_matching})-(\ref{feasible_mom_matching}) is still challenging to solve as it is not clear how to derive a suitable analytical form for $\mathcal{L}_{mom}(.)$. To the best of our knowledge, there is no mapping that directly quantify the similarity between the first $d$ moments of two given distributions. Here, we present a workaround based on the concept of RKHS embedding and  MMD distance. Our key idea is based on the following results from \cite{scholkopf}, \cite{scholkopf2}

Let $\mu_{P_f}, \mu_{p_f^{des}}$ represent the RKHS embedding of the distributions $P_f, P_f^{des}$ respectively. If the embedding is constructed through polynomial kernels, then the following theorem holds \cite{scholkopf2}, (\cite{scholkopf4}, pp-15).

\begin{theorem}\label{th_poly_mmd}
If $\Vert \mu_{P_f}(u)-\mu_{P^{des}_{f}} \Vert \rightarrow 0$, then moments of $P_f(u)$  and  $P_f^{des}$ upto order $d$ become similar.
\end{theorem}

\noindent That is, decreasing the residual of MMD distance becomes a way of matching the first $d$ moments of the distribution $P_f(u)$ and $P_f^{des}$.  Theorem  \ref{th_poly_mmd} suggest that the MMD distance can be used  either as a measure of similarity between the first $d$ moments of the two distributions. In other words, MMD with polynomial kernel can act as a surrogate for $\mathcal{L}_{mom}(.)$. Using this insight, we present the following optimization problem which can act as a surrogate for  (\ref{cost_mom_matching})-(\ref{feasible_mom_matching}).

\begin{subequations}
\begin{align}
\arg\min \rho_1\overbrace{\Vert \mu_{P_f}(u)-\mu_{P^{des}_{f}}\Vert^2}^{MMD}+\rho_2J(u)\label{cost_reform}\\
u\in \mathcal{C}\label{feasible_reform}
\end{align}
\end{subequations}

\subsection{Simplification Based on Kernel Trick}\label{simplification}
\noindent We now use the so called "kernel trick" to obtain a simplified form for the optimization (\ref{cost_reform})-(\ref{feasible_reform}) and highlight that the computational complexity of solving (\ref{cost_reform})-(\ref{feasible_reform}) is comparable to solving a deterministic variant of the original chance-constrained optimization. For  ease of exposition, we consider a specific instance from the definition of constraint function (\ref{general_chance}) with $l=2$ i.e. $f(.) = h_0(.)+h_1(.)u+h_2u^2$.

We have

\begin{eqnarray}\nonumber
\Vert \mu_{P_f}(u)-\mu_{P^{des}_{f}}\Vert^2 = \langle\mu_{P_f}(u)-\mu_{P^{des}_{f}}\rangle\\
=\langle \mu_{h_0}+\mu_{h_1}u+\mu_{h_2}u^2, \mu_{h_0}+\mu_{h_1}u+\mu_{h_2}u^2\rangle\nonumber\\-2\langle \mu_{h_0}+\mu_{h_1}u+\mu_{h_2}u^2,\mu_{P^{des}_{f}}\rangle+\langle  \mu_{P^{des}_{f}},\mu_{P^{des}_{f}}\rangle\label{mmd_expansion}
\end{eqnarray}

\noindent Expanding $\langle \mu_{h_0}+\mu_{h_1}u+\mu_{h_2}u^2, \mu_{h_0}+\mu_{h_1}u+\mu_{h_2}u^2\rangle$, we get

\small
 \begin{eqnarray}
u^4\langle\sum_{i=1}^{n}\sum_{j=1}^{n}\alpha_i\beta_jk(h_2(\textbf{w}_1^i,\textbf{w}_2^j),.),\sum_{i=1}^{n}\sum_{j=1}^{n}\alpha_i\beta_jk(h_2(\textbf{w}_1^i,\textbf{w}_2^j),.)\rangle \nonumber \\
+2u^3\langle \sum_{i=1}^{n}\sum_{j=1}^{n}\alpha_i\beta_jk(h_2(\textbf{w}_1^i,\textbf{w}_2^j),.),\sum_{i=1}^{n}\sum_{j=1}^{n}\alpha_i\beta_jk(h_1(\textbf{w}_1^i,\textbf{w}_2^j),.\rangle \nonumber \\
+2u^2\langle \sum_{i=1}^{n}\sum_{j=1}^{n}\alpha_i\beta_jk(h_2(\textbf{w}_1^i,\textbf{w}_2^j),.),\sum_{i=1}^{n}\sum_{j=1}^{n}\alpha_i\beta_jk(h_0(\textbf{w}_1^i,\textbf{w}_2^j),.)\rangle \nonumber \\+u^2\langle \sum_{i=1}^{n}\sum_{j=1}^{n}\alpha_i\beta_jk(h_1(\textbf{w}_1^i,\textbf{w}_2^j),.),\sum_{i=1}^{n}\sum_{j=1}^{n}\alpha_i\beta_jk(h_1(\textbf{w}_1^i,\textbf{w}_2^j),.)\rangle \nonumber\\
+2u\langle\sum_{i=1}^{n}\sum_{j=1}^{n}\alpha_i\beta_jk(h_1(\textbf{w}_1^i,\textbf{w}_2^j),.),\sum_{i=1}^{n}\sum_{j=1}^{n}\alpha_i\beta_jk(h_0(\textbf{w}_1^i,\textbf{w}_2^j),.)\rangle \nonumber \\
+\langle\sum_{i=1}^{n}\sum_{j=1}^{n}\alpha_i\beta_jk(h_0(\textbf{w}_1^i,\textbf{w}_2^j),.),\sum_{i=1}^{n}\sum_{j=1}^{n}\alpha_i\beta_jk(h_0(\textbf{w}_1^i,\textbf{w}_2^j),.)\rangle
 \end{eqnarray}
\label{KME_first_term}
\normalsize

\noindent Using the kernel trick, (\ref{kernel_trick}) reduces to the following expression

\small
 \begin{eqnarray}
  u^4 \textbf{c}_{\alpha\beta} \textbf{K}_{h_2h_2}\textbf{c}_{\alpha\beta}^T +2u^3\textbf{c}_{\alpha\beta} \textbf{K}_{h_2h_1}\textbf{c}_{\alpha\beta}^T 
 +2u^2\textbf{c}_{\alpha\beta} \textbf{K}_{h_2h_0}\textbf{c}_{\alpha\beta}^T \nonumber \\
  + u^2 \textbf{c}_{\alpha\beta} \textbf{K}_{h_1h_1}\textbf{c}_{\alpha\beta}^T      
 +2u\textbf{c}_{\alpha\beta} \textbf{K}_{h_1h_0}\textbf{c}_{\alpha\beta}^T+ \textbf{c}_{\alpha\beta} \textbf{K}_{h_0h_0}\textbf{c}_{\alpha\beta}^T\label{mmd_first_term}
\end{eqnarray}
  
where, 

\small
\begin{equation}
\textbf{c}_{\alpha\beta}=[\alpha_1\beta_1,\alpha_1\beta_2 ,\alpha_1\beta_3,...\alpha_n\beta_n]_{1X(n*n)} 
 \end{equation}
 \normalsize

\begin{equation}
\textbf{K}_{h_ih_j} = \begin{pmatrix}
\textbf{K}_{h_i,h_j}^{11} & \textbf{K}_{h_i,h_j}^{12} &\textbf{K}_{h_i,h_j}^{13}&..&..&\textbf{K}_{h_i,h_j}^{1n}  \\
\textbf{K}_{h_i,h_j}^{21} & \textbf{K}_{h_i,h_j}^{22} &\textbf{K}_{h_i,h_j}^{23}&..&..&\textbf{K}_{h_i,h_j}^{2n}\\
. &.&.&..&..&.\\
. &.&.&..&..&.\\
\textbf{K}_{h_i,h_j}^{n1} & \textbf{K}_{h_i,h_j}^{n2} &\textbf{K}_{h_i,h_j}^{n3}&..&..&\textbf{K}_{h_i,h_j}^{nn}
\end{pmatrix}
\label{kernel_matrix_main}
\end{equation}

$\textbf{K}_{h_ih_j}^{ab} =$
\small
\begin{equation*}\label{kernel_submatrix}
 \begin{pmatrix}
k(h_i(\textbf{w}_1^a,\textbf{w}_2^1),h_j(\textbf{w}_1^b,\textbf{w}_2^1)),&...&k(h_i(\textbf{w}_1^a,\textbf{w}_2^1),h_j(\textbf{w}_1^b,\textbf{w}_2^n))\\
k(h_i(\textbf{w}_1^a,\textbf{w}_2^2),h_j(\textbf{w}_1^b,\textbf{w}_2^1)),&...&k(h_i(\textbf{w}_1^a,\textbf{w}_2^2),h_j(\textbf{w}_1^b,\textbf{w}_2^n))\\
.,&..,&..\\
k(h_i(\textbf{w}_1^a,\textbf{w}_2^n),h_j(\textbf{w}_1^b,\textbf{w}_2^1)),&...&k(h_i(\textbf{w}_1^a,\textbf{w}_2^n),h_j(\textbf{w}_1^b,\textbf{w}_2^n))\\
\end{pmatrix}_{n\times n}
\end{equation*}
\normalsize

Following a similar process, the second term, $2\langle \mu_{h_0}+\mu_{h_1}u+\mu_{h_2}u^2,\mu_{P^{des}_{f}}\rangle$ reduces to

 \begin{eqnarray}
 2(\textbf{c}_{\alpha\beta} \textbf{K}_{h_2f}\textbf{c}_{\lambda\xi}^T u^2+ \textbf{c}_{\alpha\beta} \textbf{K}_{h_1f}\textbf{c}_{\lambda\xi}^T u+
  \textbf{c}_{\alpha\beta} \textbf{K}_{h_0f}\textbf{c}_{\lambda\xi}^T)
  \end{eqnarray}

Where,
\begin{subequations}\nonumber
\begin{align}
\textbf{c}_{\alpha\beta}=[\alpha_1\beta_1,\alpha_1\beta_2,\alpha_1\beta_3,...\alpha_n\beta_n]_{1X(n*n)}\\ 
\textbf{c}_{\lambda\xi}=[\lambda_1\xi_1,\lambda_1\xi_2,\lambda_1\xi_3,...\lambda_{n_{\textbf{w}_1}} \xi_{n_{\textbf{w}_2}}]_{1X({n_{\textbf{w}_1}}*{n_{\textbf{w}_2}})} 
\end{align}
\end{subequations}

\small
\begin{equation}
\textbf{K}_{h_if} = \begin{pmatrix}
\textbf{K}_{h_i,f}^{11} & \textbf{K}_{h_i,f}^{12} &\textbf{K}_{h_i,f}^{13}&..&..&\textbf{K}_{h_i,f}^{1n_{\textbf{w}_1}}  \\
\textbf{K}_{h_i,f}^{21} & \textbf{K}_{h_i,f}^{22} &\textbf{K}_{h_i,f}^{23}&..&..&\textbf{K}_{h_i,f}^{2n_{\textbf{w}_1}}\\
. &.&.&..&..&.\\
. &.&.&..&..&.\\
\textbf{K}_{h_i,f}^{n1} & \textbf{K}_{h_i,f}^{n2} &\textbf{K}_{h_i,f}^{n3}&..&..&\textbf{K}_{h_i,f}^{nn_{\textbf{w}_1}}
\end{pmatrix}_{n^2\times(n_{\textbf{w}_1}*n_{\textbf{w}_2})}
\end{equation}
\normalsize

\begin{figure*}
\small
\begin{equation*}
 \textbf{K}_{h_if}^{ab} =\begin{pmatrix}
k(h_i(\textbf{w}_1^a,\textbf{w}_2^1),f(\widetilde{\textbf{w}}_1^b,\widetilde{\textbf{w}}_2^1, u_{nom})),&...&k(h_i(\textbf{w}_1^a,\textbf{w}_2^1),f(\widetilde{\textbf{w}}_1^b,\widetilde{\textbf{w}}_2^{n_{\textbf{w}_2}}, u_{nom}))\\
k(h_i(\textbf{w}_1^a,\textbf{w}_2^2),f(\widetilde{\textbf{w}}_1^b,\widetilde{\textbf{w}}_2^1, u_{nom})),&...&k(h_i(\textbf{w}_1^a,\textbf{w}_2^2),f(\widetilde{\textbf{w}}_1^b,\widetilde{\textbf{w}}_2^{n_{\textbf{w}_2}}, u_{nom}))\\
\dots&\dots&\dots\\
k(h_i(\textbf{w}_1^a,\textbf{w}_2^n),f(\widetilde{\textbf{w}}_1^b,\widetilde{\textbf{w}}_2^1, u_{nom})),&...&k(h_i(\textbf{w}_1^a,\textbf{w}_2^n),f(\widetilde{\textbf{w}}_1^b,\widetilde{\textbf{w}}_2^{n_{\textbf{w}_2}}, u_{nom}))\\
\end{pmatrix}_{n\times n_{w_2}}
\end{equation*}
\normalsize
\end{figure*}

Finally, the last term, $\langle  \mu_{P^{des}_{f}},\mu_{P^{des}_{f}}\rangle$ in (\ref{mmd_expansion}) can be handled in a similar manner and thus, optimization (\ref{cost_reform})-(\ref{feasible_reform}) can be expressed as the following non-linear optimization problem

\begin{subequations}
\begin{align}
J = \rho_1(a_1u^4+a_2u^3+a_3u^2+a_4u+a_5)+\rho_2 J(u) \label{cost3}\\
u\in \mathcal{C}\label{feasible_reform3}
\end{align} 
\end{subequations}

Where,
\begin{subequations}\nonumber
\begin{align}
 a_1=\textbf{c}_{\alpha\beta} \textbf{K}_{h_2h_2} \textbf{c}_{\alpha\beta}^T, a_2=2\textbf{c}_{\alpha\beta} \textbf{K}_{h_2h_1} \textbf{c}_{\alpha\beta}^T \\
   a_3=2 \textbf{c}_{\alpha\beta} \textbf{K}_{h_2h_0} \textbf{c}_{\alpha\beta}^T +\textbf{c}_{\alpha\beta} \textbf{K}_{h_1h_1} \textbf{c}_{\alpha\beta}^T-2\textbf{c}_{\alpha\beta} \textbf{K}_{h_2f} \textbf{c}_{\lambda\xi}^T \\
  a_4=2\textbf{c}_{\alpha\beta} \textbf{K}_{h_1h_0} \textbf{c}_{\alpha\beta}^T - 2\textbf{c}_{\alpha\beta} \textbf{K}_{h_1f}\textbf{c}_{\lambda\xi}^T \\
   a_5=\textbf{c}_{\alpha\beta} \textbf{K}_{h_0h_0}\textbf{c}_{\alpha\beta}^T -2\textbf{c}_{\alpha\beta} \textbf{K}_{h_0f}\textbf{c}_{\lambda\xi}^T+\textbf{c}_{\lambda\xi}\textbf{K}_{ff}\textbf{c}_{\lambda\xi}^T
\end{align}  
\end{subequations}

\noindent \textbf{Computational Complexity}
The computational complexity of our proposed algorithm has two specific parts. The first part stems from the complexity of constructing the kernel matrix like (\ref{kernel_matrix_main}) used to formulate the cost function (\ref{cost3}). This in turn depends on the number of samples of the uncertain parameters $\textbf{w}_1$, $\textbf{w}_2$. In the worst case, we require $n^2$ samples. However, as explained in the Section \ref{reduced_set}, the value of $n$ can be optimized using the reduced set methods.

The second part of the complexity stems from how difficult it is to solve the optimization (\ref{cost3})-(\ref{feasible_reform3}). To understand this further, consider a deterministic variant of (\ref{cost}-(\ref{feasible_set})), where we replace the chance constraint (\ref{chance}) with a deterministic constraint of the form $f(\textbf{w}_1, \textbf{w}_2, u)\leq 0$. If $f(.)$ is quadratic in $u$, then the result would be a non-linear optimization problem with a quadratic constraint. In comparison, for the same form of the constraint function, our RKHS based reformulation takes the form of a non-linear optimization with a quartic polynomial of $u$ in the cost function (see (\ref{cost3})). An optimization with quartic polynomials can now be converted to that with a quadratic polynomial with a simple change of variables. Thus, it can be seen that the computational complexity of (\ref{cost3})-(\ref{feasible_reform3}) is comparable to solving a deterministic variant of the chance-constrained optimization problem. In general if $f(.)$ is a polynomial of order $l$, then our reformulation would involve a polynomial of order $2l$ in the cost function.

\subsection{Convergence}
\noindent Our proposed algorithm inherits the strong convergence guarantees associated with RKHS embedding of functions of random variables. For the individual functions $h_i(\textbf{w}_1, \textbf{w}_2)$ (see \ref{general_chance}), used to construct the chance constraints, the convergence of the RKHS embedding solely depends on the constants $\alpha_i$, $\beta_i$. To be precise, let $\hat{\mu}_{h_i}$ represent the true RKHS embedding, then the 
following holds \cite{scholkopf}.

\begin{subequations}
\begin{align}
\text{Assumption:} \Sigma \alpha_i=1, \Sigma \beta_i=1, \Sigma \alpha_i^2, \Sigma \beta_i^2 \rightarrow 0, n\rightarrow \infty \label{assumption_convergence}\\
\Vert \hat{\mu}_{h_i}-\mu_{h_i}\Vert = O_p(\sqrt{\Sigma \alpha_i^2}+\sqrt{\sum\beta_i^2})  \label{convergence}
\end{align}
\end{subequations}

\noindent Where, $O_p$ represents convergence in probability. The assumptions (\ref{assumption_convergence}) are trivially satisfied if we use i.i.d samples of $\textbf{w}_1, \textbf{w}_2$ to construct $\mu_{h_i}$. Furthermore, the assumptions are  also satisfied when working with the reduced set samples computed by the approach described in Section \ref{reduced_set}. 

The convergence of $\mu_{P_f}$ in RKHS also follow similar arguments. However, the convergence rate in this case also depends on the value of the decision variable $u$ besides $\alpha_i$ and $\beta_i$ (see (\ref{kernel_mean_chance})). Intuitively, the reason for this can be understood in the following manner: The shape of the distribution $P_f(u)$ depends on $u$, and thus for some $u$ it can attain a very peculiar shape, which would require a larger number of samples for accurate enough estimation.

\section{Applications}
In this section, we consider two robotic/control applications and  model them in the form of the chance-constrained optimization (\ref{cost})-(\ref{feasible_set}) and also present their RKHS reformulations.

\subsection{Dynamic Obstacle Avoidance along a Given Path}\label{obst_avoid}
\begin{figure}[!tbh]
  \centering
\subfigure[]{
\includegraphics[width= 4.0cm, height=3.0cm]{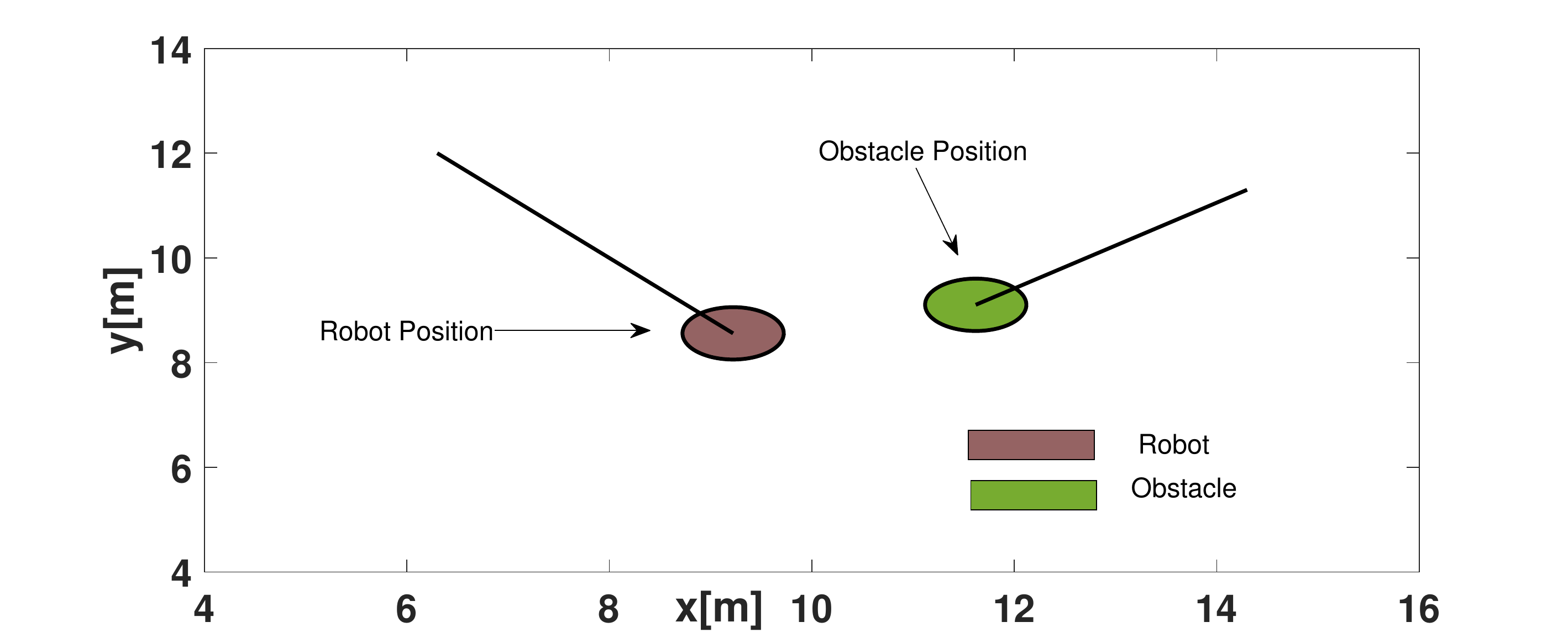}
        \label{one_obst_det}
        }\hspace{-0.7cm}
\subfigure[]{
\includegraphics[width= 4.0cm, height=3.0cm]{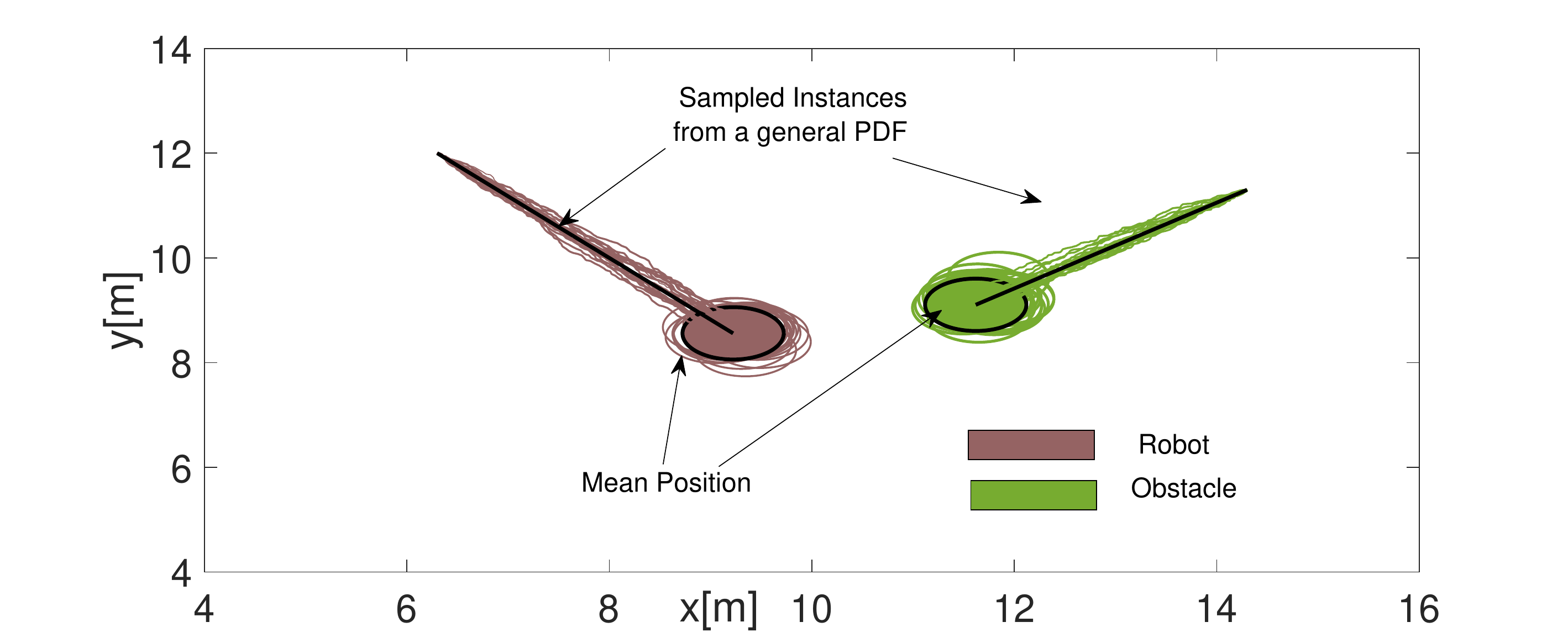}
        \label{one_obst_uncert}
        }
\caption{(a): Deterministic collision avoidance between a robot and a dynamic obstacle. (b): Stochastic variant of the dynamic collision avoidance. The robot has both perception and ego-motion uncertainty.}
\end{figure}

\noindent Here, we consider dynamic collision avoidance between a disk shaped robot and non-reactive moving obstacles with similar shapes (Fig. \ref{one_obst_det}). Both the robot and the obstacles are assumed to have a single integrator motion model, i.e they can instantaneously change their velocities. Further, we consider a variant of the problem where the path of the robot is fixed and the robot achieves collision avoidance simply by varying the magnitude of its forward velocity. As shown in our earlier works \cite{iros14_bharath}, \cite{iros17}, the more general collision avoidance like \cite{rvo}, \cite{nh_orca} can be conveniently built from this special case. 

Let, $(x, y)$ and $(\dot{x}, \dot{y})$ be the position and velocity vector of the robot at some specific time instant when the robot detects imminent collision with the obstacles. Similarly, let $(x_o, y_o)$ and $(\dot{x}_o, \dot{y}_o)$ represent similar vectors for the  moving obstacle. It is clear that if the velocity vector of the robot is modified as $(u\dot{x}, u\dot{y})$, then it continues to move along its current path although the magnitude of its forward velocity gets scaled by a factor $u$. For $u>1$, the robot would increase its forward velocity while for $u<1$, it would slow down, to avoid collisions. Therefore the dynamic collision avoidance constraint can be written in the following form (refer to \cite{iros17} for details).

\small
\begin{subequations}
\begin{align}
\frac{(\textbf{r}^T\textbf{v})^2}{\Vert \textbf{v}\Vert^2}-\Vert\textbf{r}\Vert^2+R^2\leq 0\label{collcone}\\
R=R+R_o\\
 \textbf{r}= \begin{bmatrix}
    x-x_o  \\
   y-y_o \\
\end{bmatrix} , \textbf{v} =   \begin{bmatrix}
    u\dot{x}-\dot{x}_o  \\
   u\dot{y}-\dot{y}_o \\
\end{bmatrix}.
\end{align}
\end{subequations}
\normalsize

\noindent Where, $R, R_o$ represent the radius of the footprint of the robot and the obstacle. Inequality (\ref{collcone}) can be put in the following more compact form, which resembles (\ref{general_chance}) with $l=2$. 

\begin{equation}
f(\textbf{w}_1, \textbf{w}_2, u): h_0(\textbf{w}_1, \textbf{w}_2)+h_1(\textbf{w}_1, \textbf{w}_2)u+h_2(\textbf{w}_1, \textbf{w}_2)u^2\leq 0
\label{collision_constraint}
\end{equation}

\noindent where, $\textbf{w}_1 = (x,y,\dot{x}, \dot{y})$ and $\textbf{w}_2 = (x_o,y_o,\dot{x}_o, \dot{y}_o)$. 

\noindent \textbf{Uncertainty}: Assume that the robot has both perception and ego-motion uncertainty (Fig. \ref{one_obst_uncert}). That is, 

\begin{subequations}
\begin{align*}
x \sim P(\mu ^{x}, \Sigma ^{x},\lambda_1^{x},\lambda_2^{x})\\
\dot{x} \sim P(\mu ^{\dot{x}}, \Sigma ^{\dot{x}},\lambda_1^{\dot{x}},\lambda_2^{\dot{x}})\\
y \sim P(\mu ^{y}, \Sigma ^{y},\lambda_1^{y},\lambda_2^{y})\\
\dot{y} \sim P(\mu ^{\dot{y}}, \Sigma ^{\dot{y}},\lambda_1^{\dot{y}},\lambda_2^{\dot{y}})\\
x_o \sim P(\mu ^{x_o}, \Sigma ^{x_o},\lambda_1^{x_o},\lambda_2^{x_o})\\
\dot{x}_o \sim P(\mu ^{\dot{x}_o}, \Sigma ^{\dot{x}_o},\lambda_1^{\dot{x}_o},\lambda_2^{\dot{x}_o})\\
y_o \sim P(\mu ^{y_o}, \Sigma ^{y_o},\lambda_1^{y_o},\lambda_2^{y_o})\\
\dot{y}_o \sim P(\mu^{\dot{y}_o}, \Sigma ^{\dot{y}_o},\lambda_1^{\dot{y}_o},\lambda_2^{\dot{y}_o})
\end{align*}
\label{distribution_general_form}
\end{subequations}

\noindent $P$  denotes a general PDF that is defined until its fourth order of moment.  $\lambda_1^{(.)} $ and $\lambda_2^{(.)}$   denote the  third(skewness) and fourth(kurtosis) moments respectively. If $\lambda_1^{x}, \lambda_1^{x}=0$, and $\lambda_2^{x}, \lambda_2^{x}=3$, then PDFs take the Gaussian form.

Both $\textbf{w}_1, \textbf{w}_2$  then become uncertain parameters and thus collision avoidance needs to be rephrased as chance constraints. The final chance constrained optimization for dynamic collision avoidance under uncertainty takes the following form.

%

\begin{subequations}
\begin{align}
\min J(u) = (u-1)^2 \label{cost_dynamic} \\
P(f(\textbf{w}_1, \textbf{w}_2, u)\leq 0)\geq \eta\label{chance_collavoid} \\
u\geq 0 \label{feasible_collavoid}
\end{align}
\end{subequations}

The cost (\ref{cost_dynamic}) minimizes the deviation from the current forward velocities. Optimization (\ref{cost_dynamic})-(\ref{feasible_collavoid}) fits in the form described by (\ref{cost})-(\ref{feasible_set}). After solving the above optimization problem or rather the RKHS embedding based reformulation of it, the robot draws a sample from its current velocity distribution $\dot{x}$, $\dot{y}$ and executes it after scaling by a factor $u$ to avoid collisions. 

\noindent \textbf{Multiple moving obstacles}: If there are multiple moving obstacles in the environment, then the parameter $\textbf{w}_2$ needs to be computed specifically for each moving obstacle. That is, we have:

\begin{equation*}
^{i}\textbf{w}_2 =  (^{i}x_o, ^{i}y_o, ^{i}\dot{x}_o, ^{i}\dot{y}_o)
\end{equation*}

\noindent Consequently, we will also have multiple collision avoidance constraints:

\begin{equation}
f_i(\textbf{w}_1, ^{i}\textbf{w}_2, u): h_0(\textbf{w}_1, ^{i}\textbf{w}_2)+h_1(\textbf{w}_1, ^{i}\textbf{w}_2)u+h_2(\textbf{w}_1, ^{i}\textbf{w}_2)u^2\leq 0
\label{collision_constraint_multi}
\end{equation}

\noindent The chance-constrained optimization would now have multiple chance constraints and take the following form.

\begin{subequations}
\begin{align}
\min J(u) = (u-1)^2 \label{cost_dynamic_multi} \\
P(f_i(\textbf{w}_1, ^{i}\textbf{w}_2, u)\leq 0)\geq \eta, \forall, i = 1,2..m\label{chance_collavoid_multi} \\
u\geq 0 \label{feasible_collavoid_multi}
\end{align}
\end{subequations}

\noindent Our RKHS embedding based reformulation would now have the following form:

\begin{subequations}
\begin{align}
\min \rho_1\sum\overbrace{\Vert \mu_{P_{f_i}}(u)-\mu_{P^{des}_{f_i}}\Vert^2}^{MMD}+\rho_2J(u)\label{cost_reform_collavoid}\\
u\in \mathcal{C}\label{feasible_reform_collavoid}
\end{align}
\end{subequations}

\noindent where, $\mu_{P_{f_i}}(u)$ represents the KME of the $i^{th}$ chance constraints and $\mu_{P^{des}_{f_i}}$ represents the KME of the desired distribution corresponding to the $i^{th}$ chance constraints. Note that the first term in (\ref{cost_reform_collavoid}) can be obtained using the derivations presented in Section \ref{simplification}.

\subsection{Inverse Dynamics based Path Tracking} \label{path_tarcking_section}
\begin{figure}[!tbh]
  \centering
\subfigure[]{
\includegraphics[width= 8.5cm, height=4.0cm]{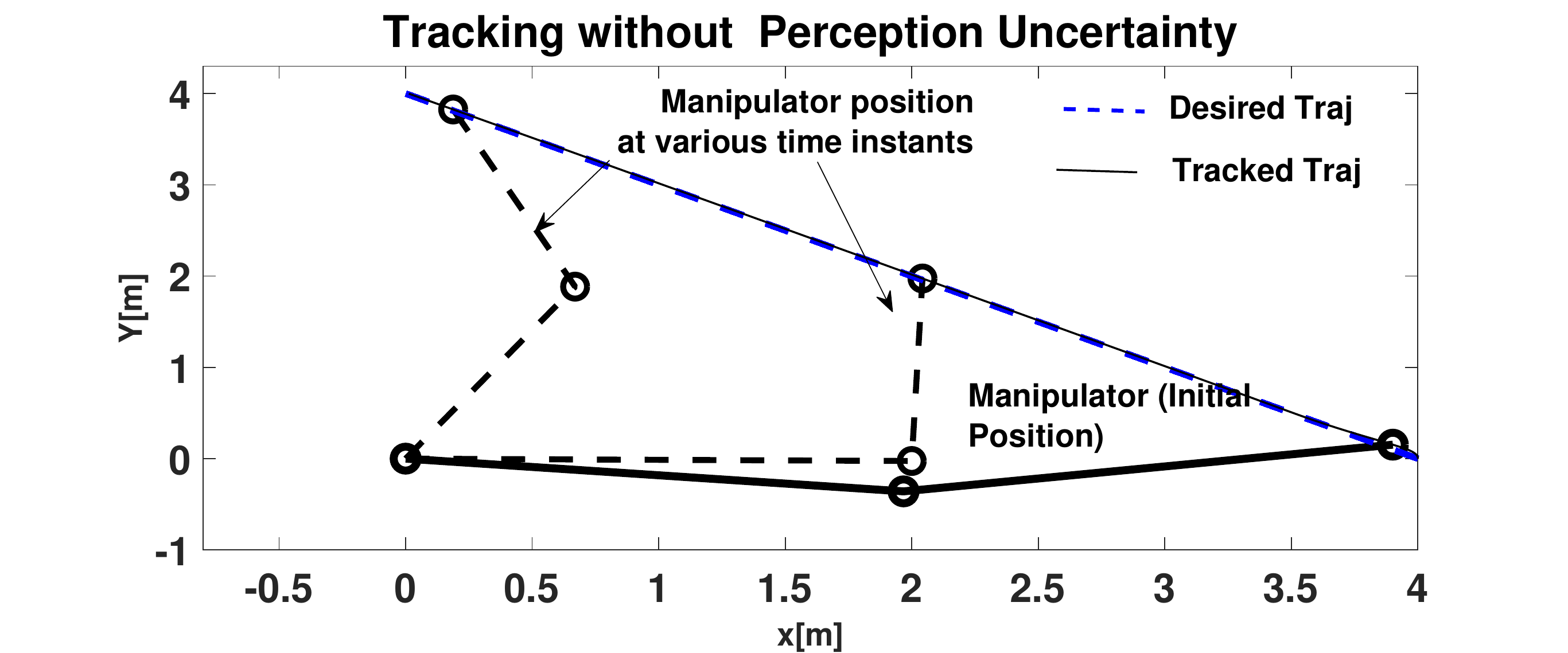}
        \label{man_det}
        }\hspace{-0.7cm}
\subfigure[]{
\includegraphics[width= 8.5cm, height=4.0cm]{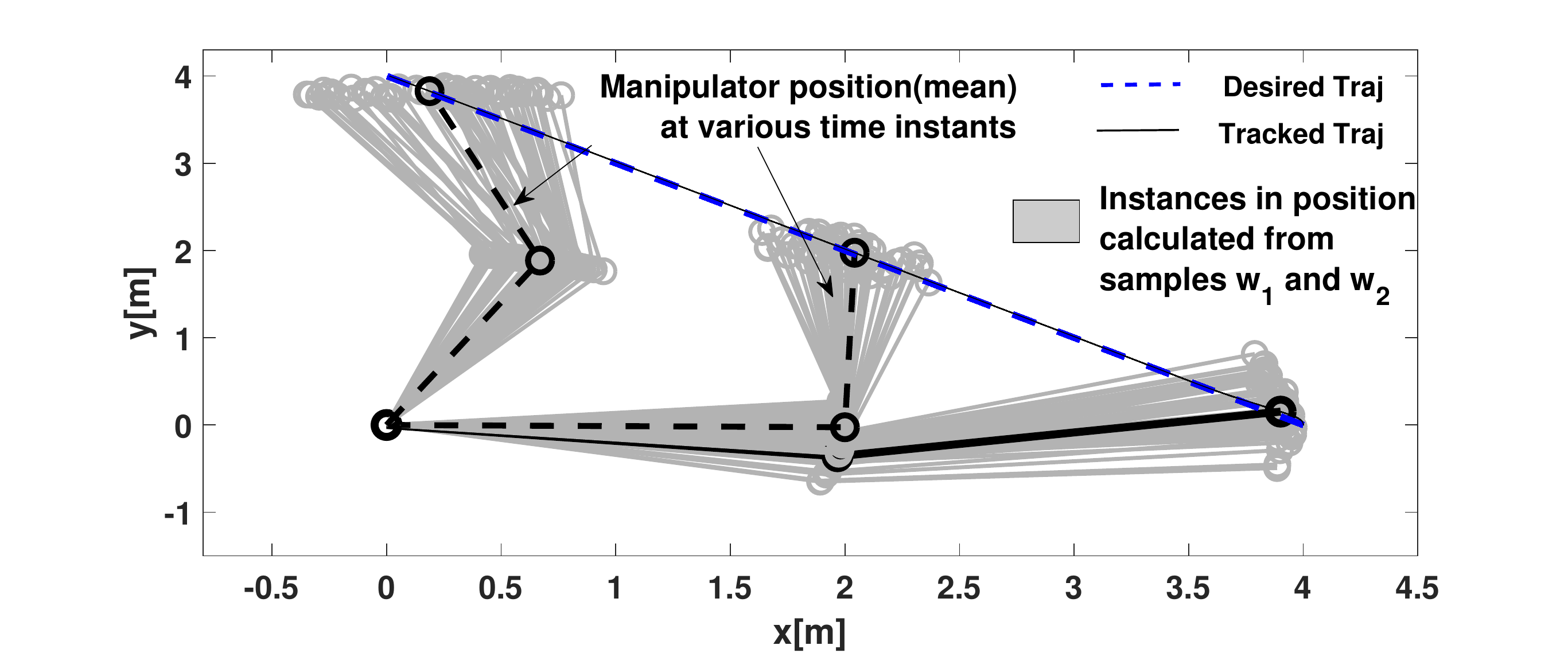}
        \label{man_uncert}
        }
\caption{(a): Problem set-up for inverse dynamics based path tracking for a two-link planar manipulator. (b): Inverse dynamics based path tracking under perception uncertainty leading to noisy estimates for joint position $\textbf{q}(t)$ and joint velocities $\dot{\textbf{q}}(t)$}
\end{figure}

\noindent In this application, we consider the task of tracking a reference trajectory $\textbf{x}_d(t)$ by a manipulator (Fig. \ref{man_det}), which can be framed as the following quadratic programming (QP) problem.


\begin{subequations}
\begin{align}
\arg\min_{\ddot{\textbf{q}}(t)} \frac{1}{2}\Vert \textbf{J}(\textbf{q}(t))\ddot{\textbf{q}}(t)+\dot{\textbf{J}}(\textbf{q}(t), \ddot{\textbf{q}}(t))\dot{\textbf{q}}(t)-\ddot{\textbf{x}}(t)\Vert_2^2 \label{cost_invdyn}  \\
\textbf{M}(\textbf{q}(t))\ddot{\textbf{q}}(t)+\textbf{C}(\textbf{q}(t), \dot{\textbf{q}}(t))\dot{\textbf{q}}(t)\leq \tau_{max} \label{ac1}  \\
\textbf{M}(\textbf{q}(t))\ddot{\textbf{q}}(t)+\textbf{C}(\textbf{q}(t), \dot{\textbf{q}}(t))\dot{\textbf{q}}(t)\geq -\tau_{max} \label{ac2} \\
\vert\ddot{\textbf{q}}(t)\vert \leq \ddot{\textbf{q}}_{max}\label{ac3}.
\end{align}
\end{subequations}

Where, $\ddot{\textbf{x}}(t)=k_p(\textbf{x}(t)-\textbf{x}_d(t))+2*\sqrt{k_p}(\ddot{\textbf{x}}(t)-\ddot{\textbf{x}}_d(t))+\ddot{\textbf{x}}_d(t)$ and $k_p$ is a constant feedback gain.  $\textbf{q}(t)$ and $\dot{\textbf{q}}(t)$ represents the joint angle and velocities at time $t$. Let the degree of freedom  of the manipulator be $m$, i.e, $\textbf{q}(t) = (q_1(t), q_2(t)...q_m(t))$. $\textbf{J}$ is the manipulator Jacobian matrix. The inequalities (\ref{ac1})-(\ref{ac2}) ensures that the resulting $\ddot{\textbf{q}}(t)$ is achievable without violating the torque bounds. The QP (\ref{cost_invdyn})-(\ref{ac3}) is solved in a one-step receding horizon setting for trajectory tracking. To be precise,the QP is solved for the joint accelerations at each instant considering the current joint position and velocities. The state is evolved with the current acceleration and the process is repeated for a specific time duration.

Constraints (\ref{ac1})-(\ref{ac2}) represent $2m$ affine inequalities  
each of which can be represented in the following familiar form:
%

\small
\begin{eqnarray}
f_i(\textbf{w}_1, \textbf{w}_2, u_1, u_2..u_n) = \sum_{j=1}^{j=m}h_{i}^j(\textbf{w}_1. \textbf{w}_2) u_{j}(t)+h_i(\textbf{w}_1, \textbf{w}_2) \leq 0\nonumber\\
\forall i = 1,2..2m \label{inv_dyn_general_const}
\end{eqnarray}
\normalsize

\begin{equation*}
\text{where}, \textbf{w}_1 = (q_1(t), q_2(t)...q_m(t)), \textbf{w}_2 = (\dot{q}_1(t), \dot{q}_2(t)...\dot{q}_m(t))
\end{equation*}
\begin{equation*}
(u_1, u_2..u_m) = (\ddot{q}_1(t), \ddot{q}_2(t)..\ddot{q}_m(t))
\end{equation*}

\noindent \textbf{Trajectory Tracking under Perception Uncertainty}
Assume that the manipulator has perfect motion capability but imperfect sensing for the joint angles $\textbf{q}(t)$ and velocity $\dot{\textbf{q}}(t)$ (Fig. \ref{man_uncert}). In such a case, $\textbf{q}(t)$, $\dot{\textbf{q}}(t)$  and functions $h_i^j(.)$ and $h_i(.)$ can be modeled as random variables. With this insight, we now formulate a stochastic variant of the inverse dynamics based path tracking problem as the following chance-constrained optimization:

\begin{subequations}
\begin{align}
\arg\min_{\ddot{\textbf{q}}(t)} \frac{1}{2}\Vert \textbf{J}(\overline{\textbf{q}}(t))\ddot{\textbf{q}}(t)+\dot{\textbf{J}}(\overline{\textbf{q}}(t), \overline{\dot{\textbf{q}}}(t))\dot{\textbf{q}}(t)-\ddot{\textbf{x}}(t)\Vert_2^2 \label{cost_chance_invdyn}\\
P(f_i(\textbf{w}_1, \textbf{w}_2, u_1, u_2..u_n)\leq 0)\geq \eta \label{chance_invdyn}\\
\vert\ddot{\textbf{q}}(t)\vert \leq \ddot{\textbf{q}}_{max}\label{accbound_chance_invdyn}
\end{align}
\end{subequations}

where, $J(\overline{\textbf{q}}(t))$ and $\dot{\textbf{J}}(\overline{\textbf{q}}(t), \overline{\dot{\textbf{q}}}(t))$ represents the Jacobian matrix formed with the mean variables $\overline{\textbf{q}}(t)$ and $\overline{\dot{\textbf{q}}}(t)$. The inequality (\ref{chance_invdyn}) ensures that the resulting $\ddot{\textbf{q}}(t)$ can be achieved without violating the torque bounds with atleast probability $\eta$. It can be seen that, (\ref{cost_chance_invdyn})-(\ref{accbound_chance_invdyn}) is an extended variant of the original chance constrained optimization (\ref{cost})-(\ref{feasible_set}). Specifically, we now have multiple decision variables along with multiple chance constraints.

\begin{remark}\label{multiple_chance_observation}
There is a subtle difference between the multiple chance constraints in optimization (\ref{cost_dynamic_multi})-(\ref{feasible_collavoid_multi}) and (\ref{cost_chance_invdyn})-(\ref{accbound_chance_invdyn}). In the former, multiple chance constraints arise because the parameters $^{i}\textbf{w}_2$ were different for each obstacle while the  function $f(.)$ remained the same for each constraint. In contrast, in the latter, the  functions $f_i(.)$ were different for each constraint but the parameters $\textbf{w}_1$, $\textbf{w}_2$ remained same across different constraints.
\end{remark}

The RKHS embedding based reformulation of (\ref{cost_chance_invdyn})-(\ref{accbound_chance_invdyn}) takes the following form.
\small
\begin{subequations}
\begin{align}
\min \rho_1\sum\overbrace{\Vert \mu_{P_{f_i}}(u_1, u_2..u_n)-\mu_{P^{des}_{f_i}}\Vert^2}^{MMD}+\rho_2J(u_1, u_2,..u_n)\label{cost_reform_invdyn}\\
\vert\ddot{\textbf{q}}(t)\vert \leq \ddot{\textbf{q}}_{max}\label{feasible_reform_invdyn}
\end{align}
\end{subequations}
\normalsize

\noindent where, $\mu_{P_{f_i}}(.)$ represents the KME of the $i^{th}$ chance constraints and $\mu_{P^{des}_{f_i}}$ represents the KME of the desired distribution corresponding to the $i^{th}$ chance constraint. 


\section{Results}
In this section, we present simulations obtained by applying our formulation to the examples derived in the previous section. During each application, we also separately benchmark our formulation with the some of the  existing approaches for chance constrained optimization.The simulation videos for the results can be found in \url{http://robotics.iiit.ac.in/uploads/Main/Publications/Bharath_journal/}.  

\begin{figure}[!h]
  \centering
 \subfigure[]{     
\includegraphics[width= 4.2cm, height=4.0cm]{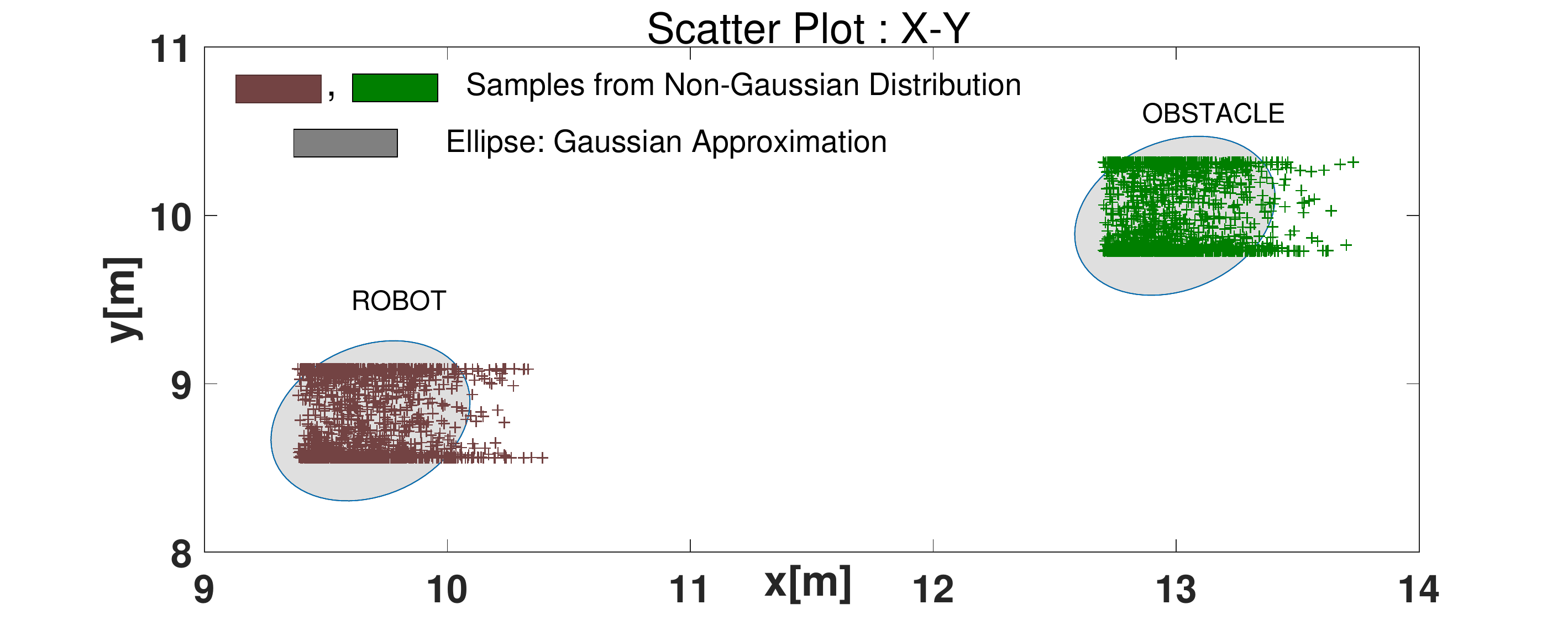}
        \label{samples_non_gaussian}
        }\hspace{-0.7cm}
 \subfigure[]{      
\includegraphics[width= 4.2cm, height=4.0cm]{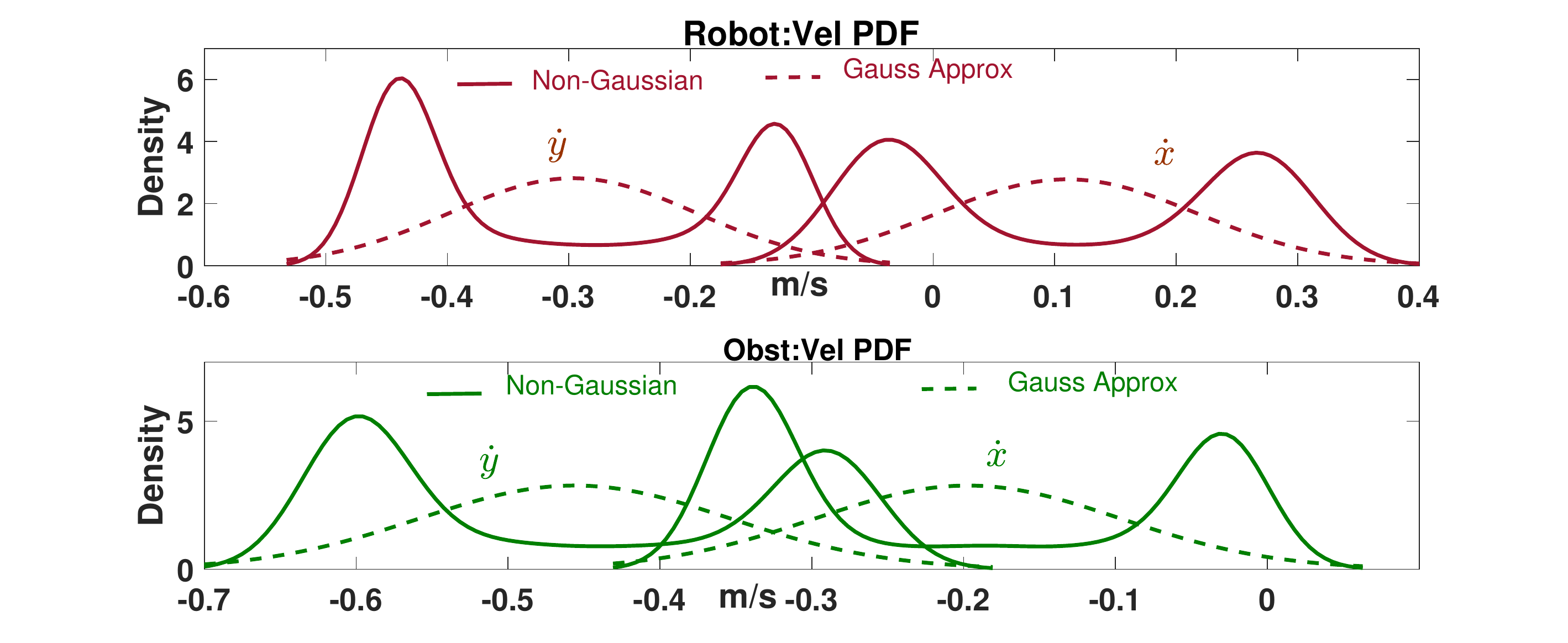}
        \label{kde_xdot_ydot_nongaussian}
        }        
 \caption{This figure corresponds to  collision avoidance with a single moving obstacle shown in Fig.\ref{one_obst_uncert}. The uncertainty considered here is non-Gaussian in nature. (a): The position samples of robot and obstacle at some specific instant when the robot detects the imminent collision and decides to modify its velocity. (b): The uncertainty in velocity of the robot and the obstacle. We used the Kernel Density Estimation technique to obtain a graphical representation of the uncertainty in the velocity. The Gaussian approximation for these non-Gaussian distributions are also shown in lighter shades.It is evident that the Gaussian approximation results in a poor inference of the actual probability. }
\end{figure} 

%
%
 
\begin{figure*}
\centering
\subfigure[]{
\includegraphics[width=6.0cm,height=4.0cm]{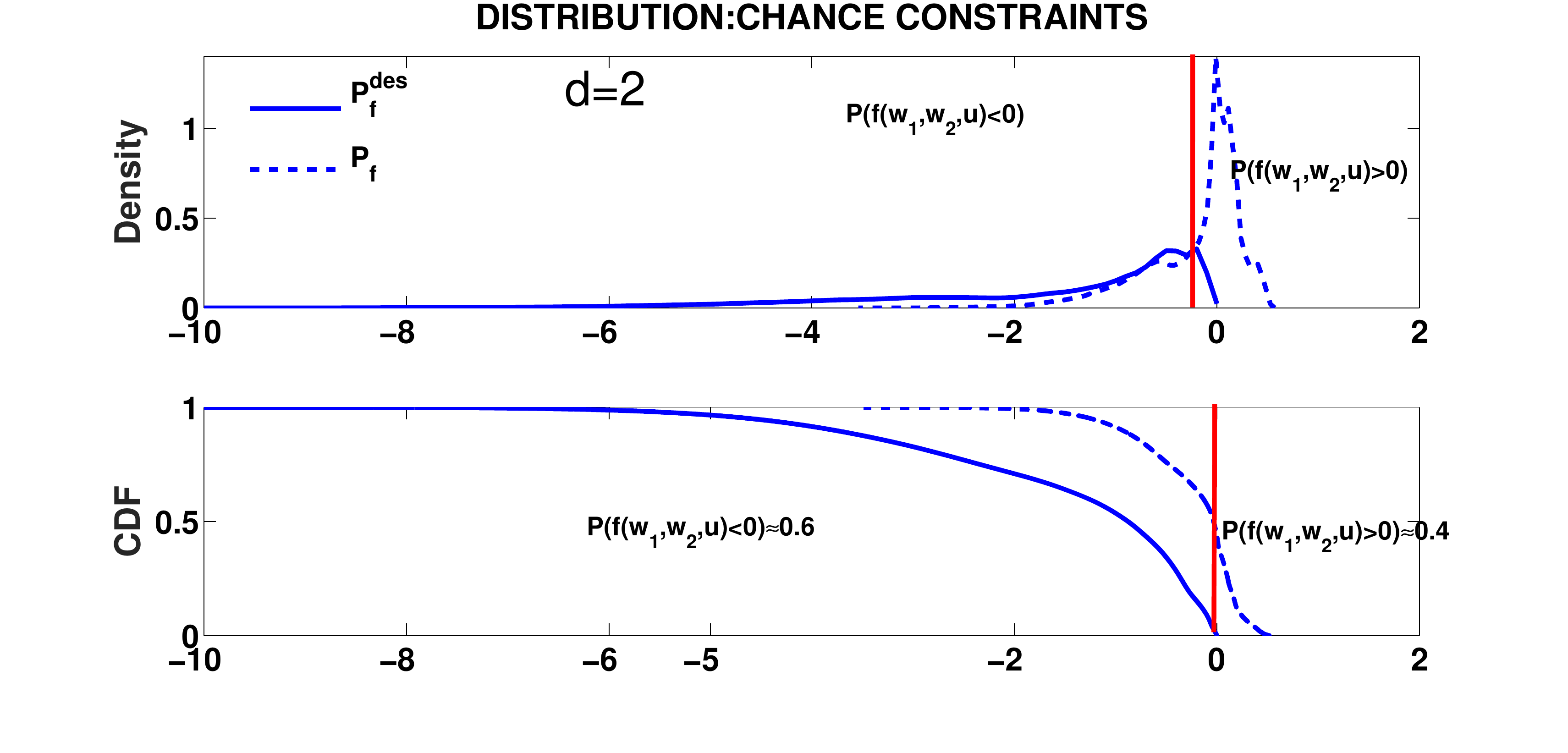}\label{density_nongauss_d2}
      }\hspace{-0.7cm}
 \subfigure[]{
\includegraphics[width=6.0cm,height=4.0cm]{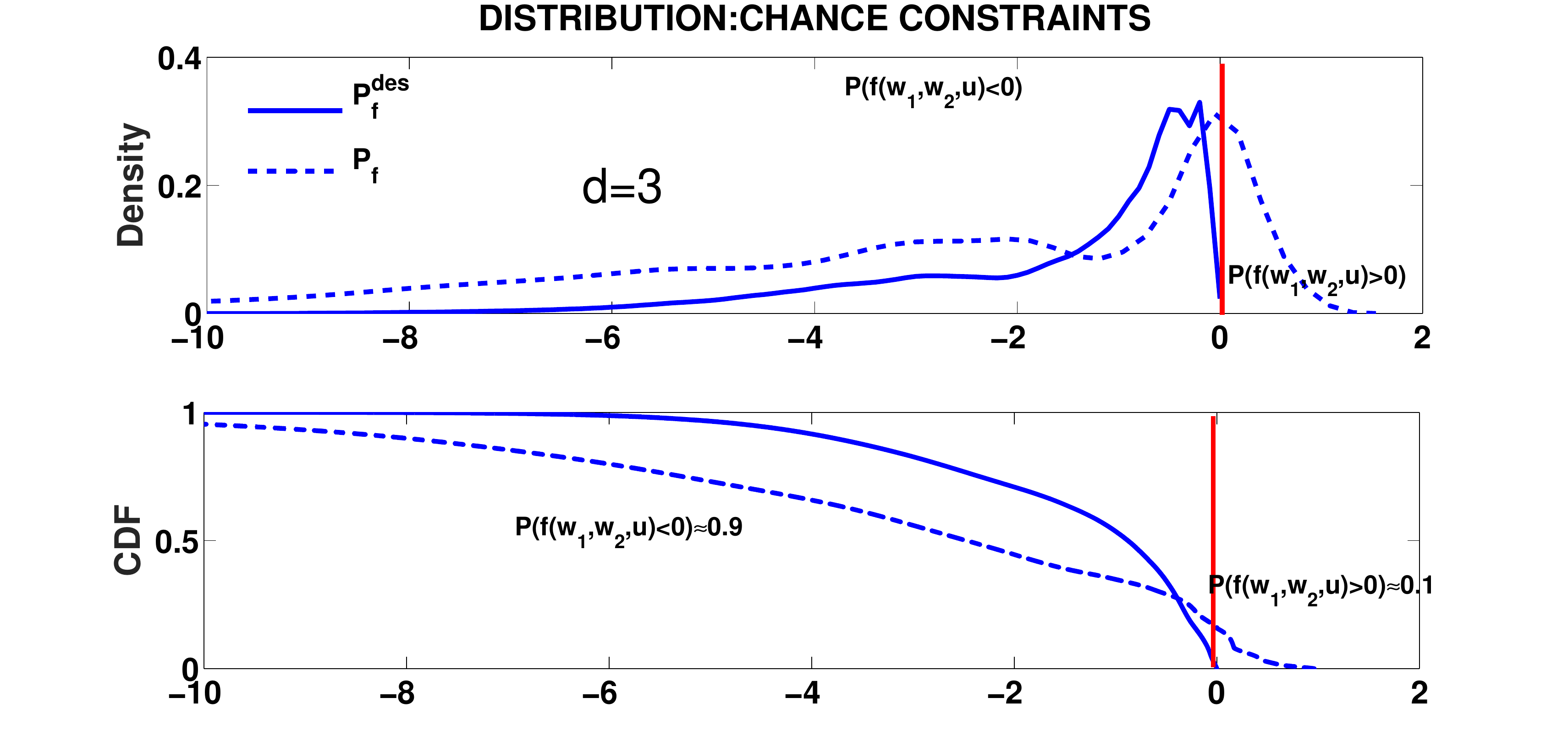}\label{density_nongauss_d3}
      }  \hspace{-0.7cm}
        \subfigure[]{
\includegraphics[width=6.0cm,height=4.0cm]{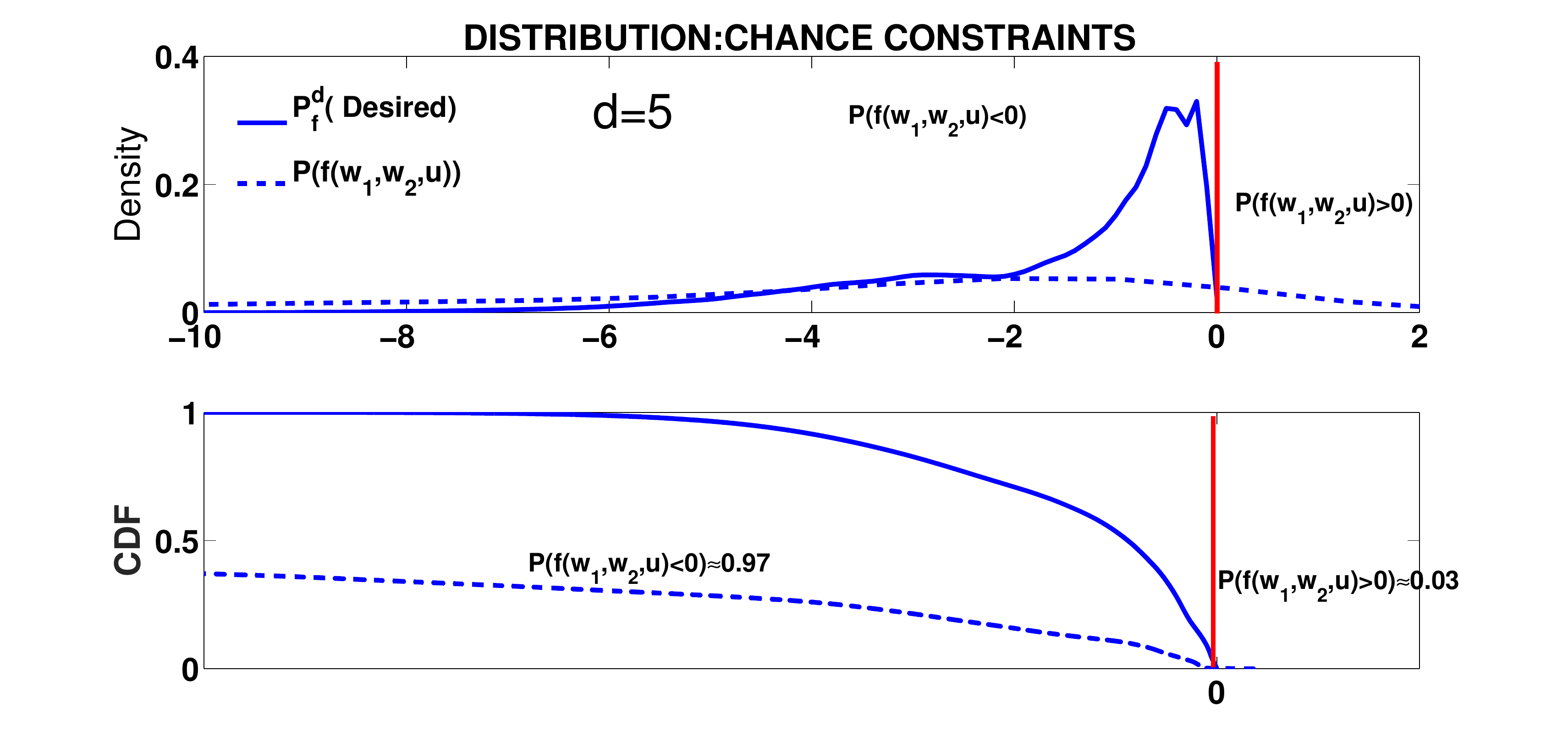}\label{density_nongauss_d5}
      }
      \subfigure[]{
\includegraphics[width=6.0cm,height=4.0cm]{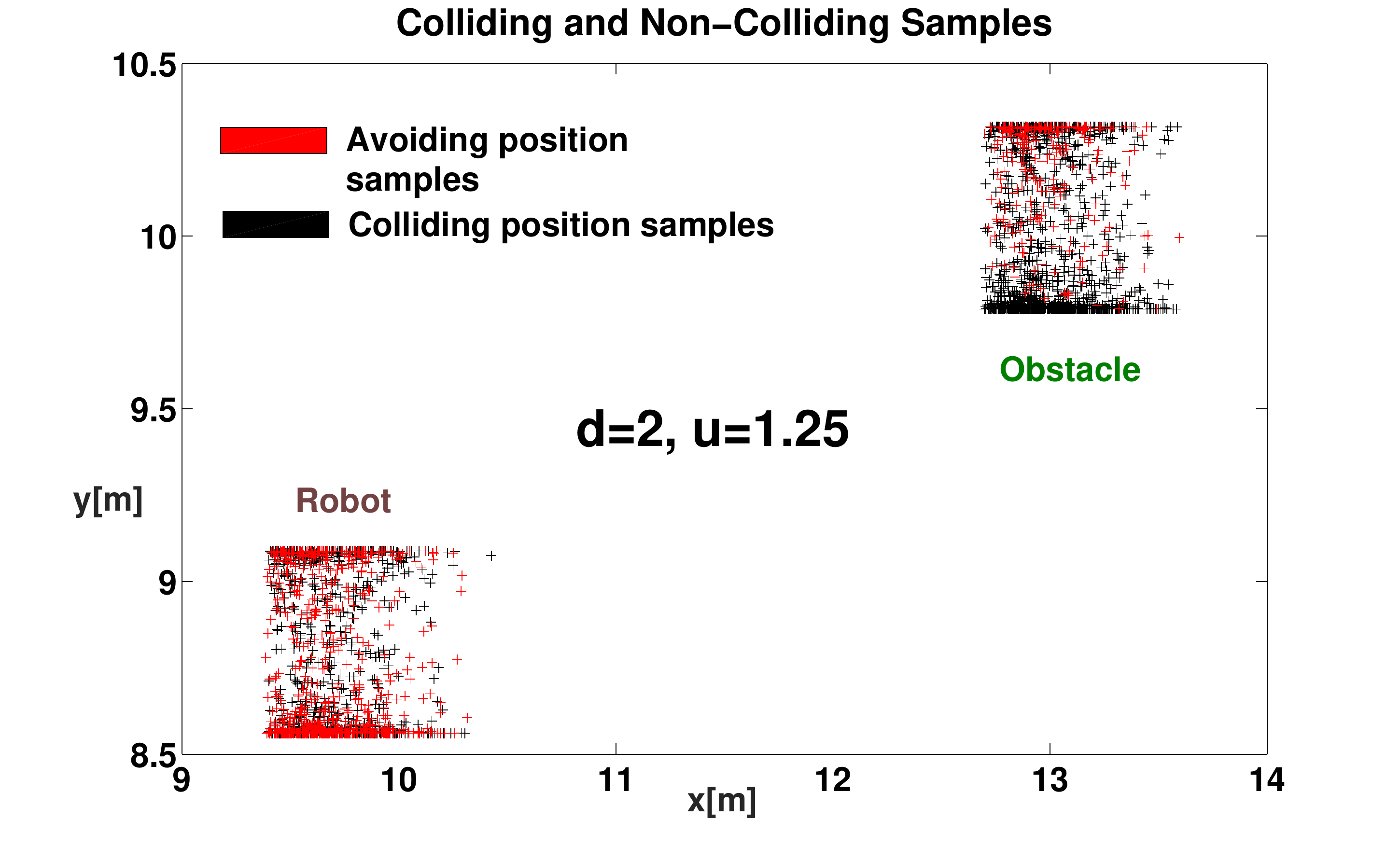}\label{d_2_samples_nongauss}
      }\hspace{-0.7cm}
           \subfigure[]{
\includegraphics[width=6cm,height=4.0cm]{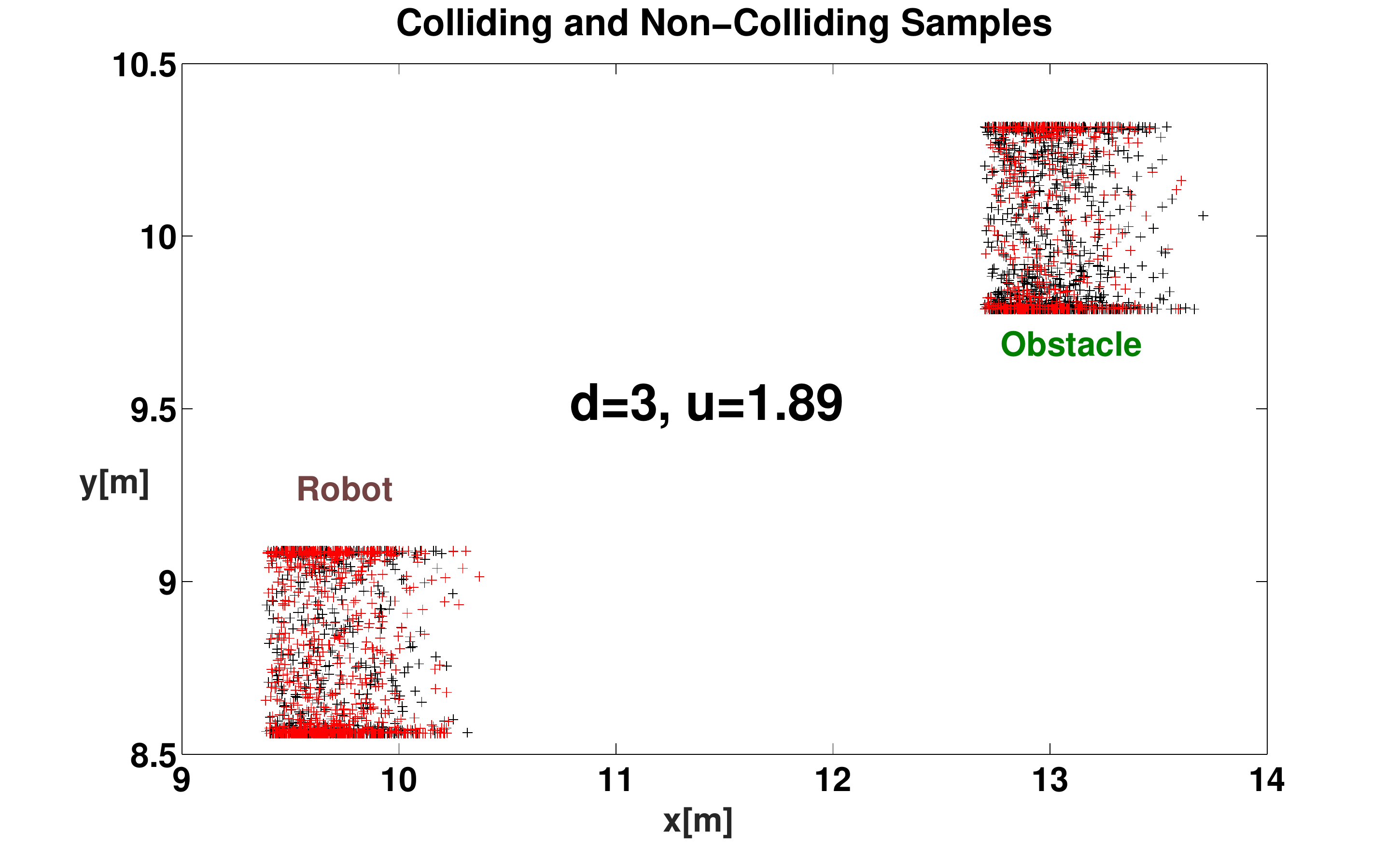}\label{d_3_samples_nongauss}
      }   \hspace{-0.7cm}
      \subfigure[]{
\includegraphics[width=6.0cm,height=4.0cm]{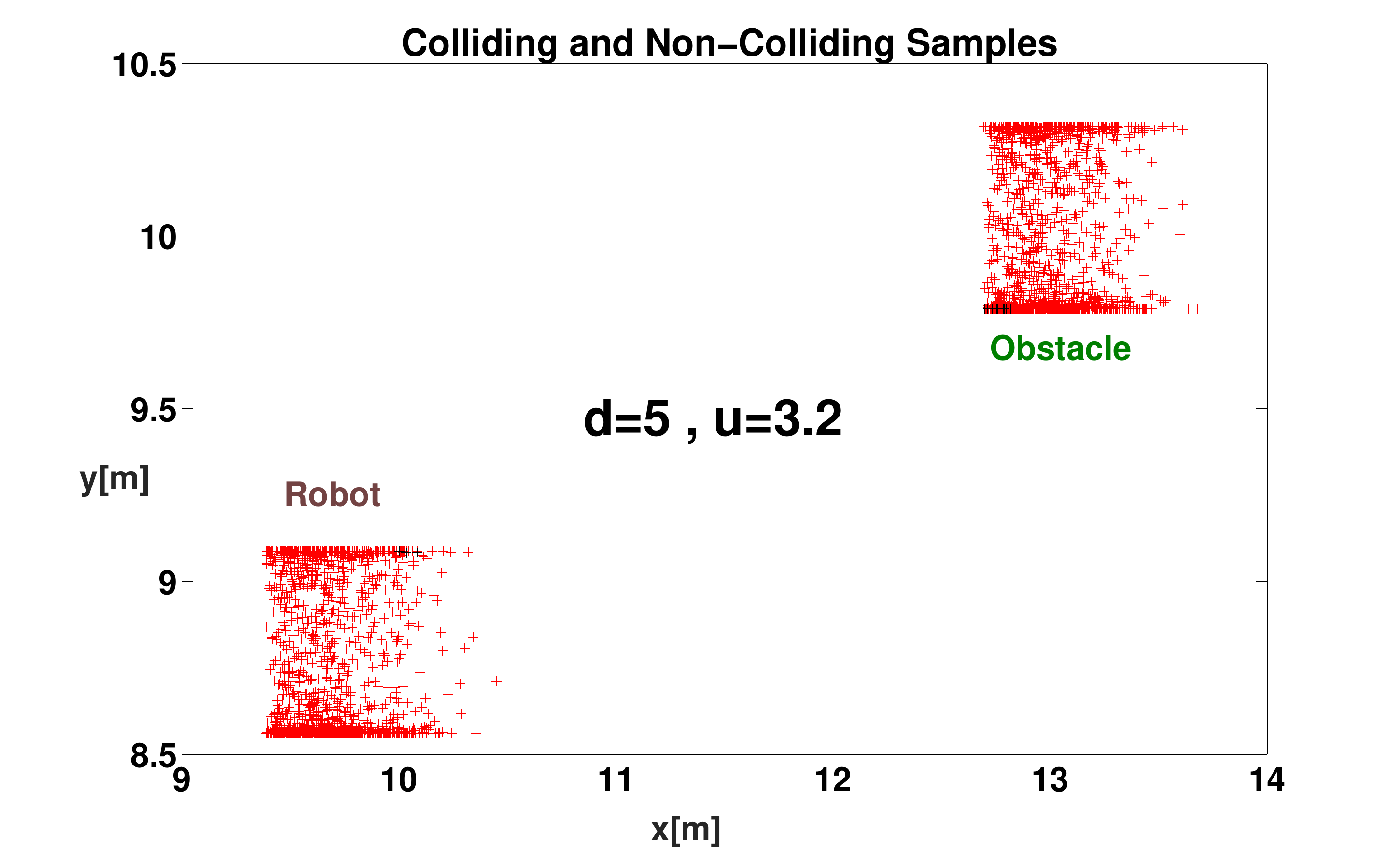}\label{d_5_samples_nongauss} 
      }
      \caption{The figures present the simulation results for collision avoidance with a single obstacle (Fig. \ref{one_obst_uncert}) under non-Gaussian uncertainty. Figures (a), (b), (c) show the constructed desired distribution $P_f^{des}$ and the distribution of $P_{f}(u)$. The increase in the degree of the polynomial kernel $d$ can be easily correlated with the decrease in the colliding samples shown in figures (d), (e), (f). }
\end{figure*}

\begin{figure*}[!h]
\subfigure[]{
\includegraphics[width=6cm,height=5cm]{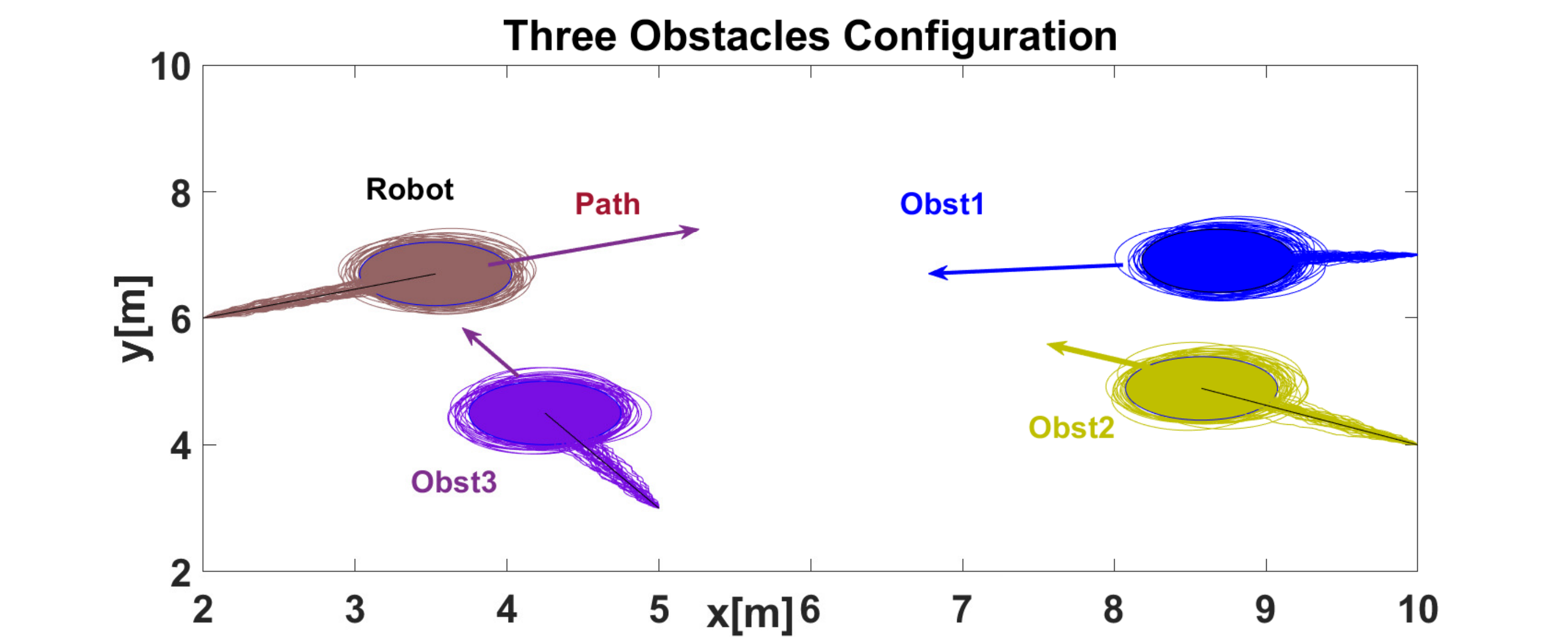}\hspace{-0.7cm}
     \label{Three_obstacles_config} }
 \subfigure[]{
\includegraphics[width=6cm,height=5cm]{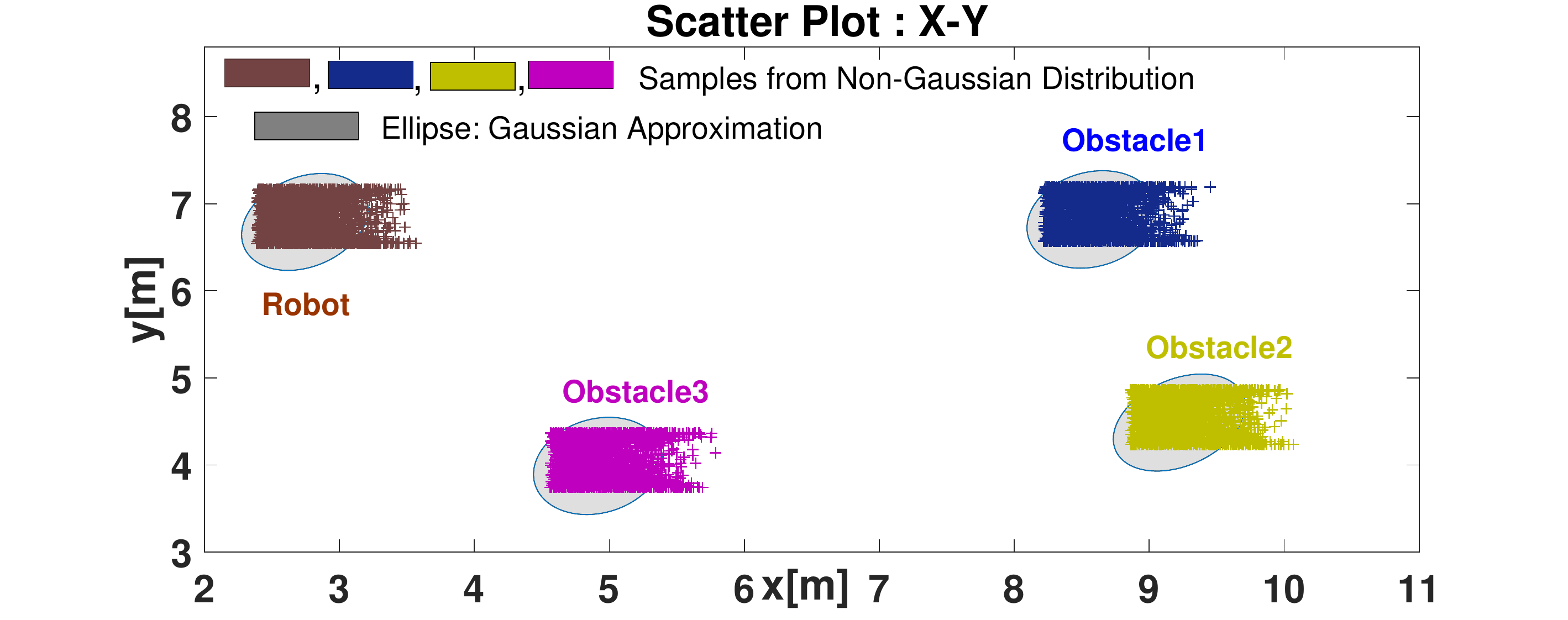}
     \label{three_obstacles_colliding} }\hspace{-0.7cm} 
       \subfigure[]{
\includegraphics[width=6.9cm,height=5cm]{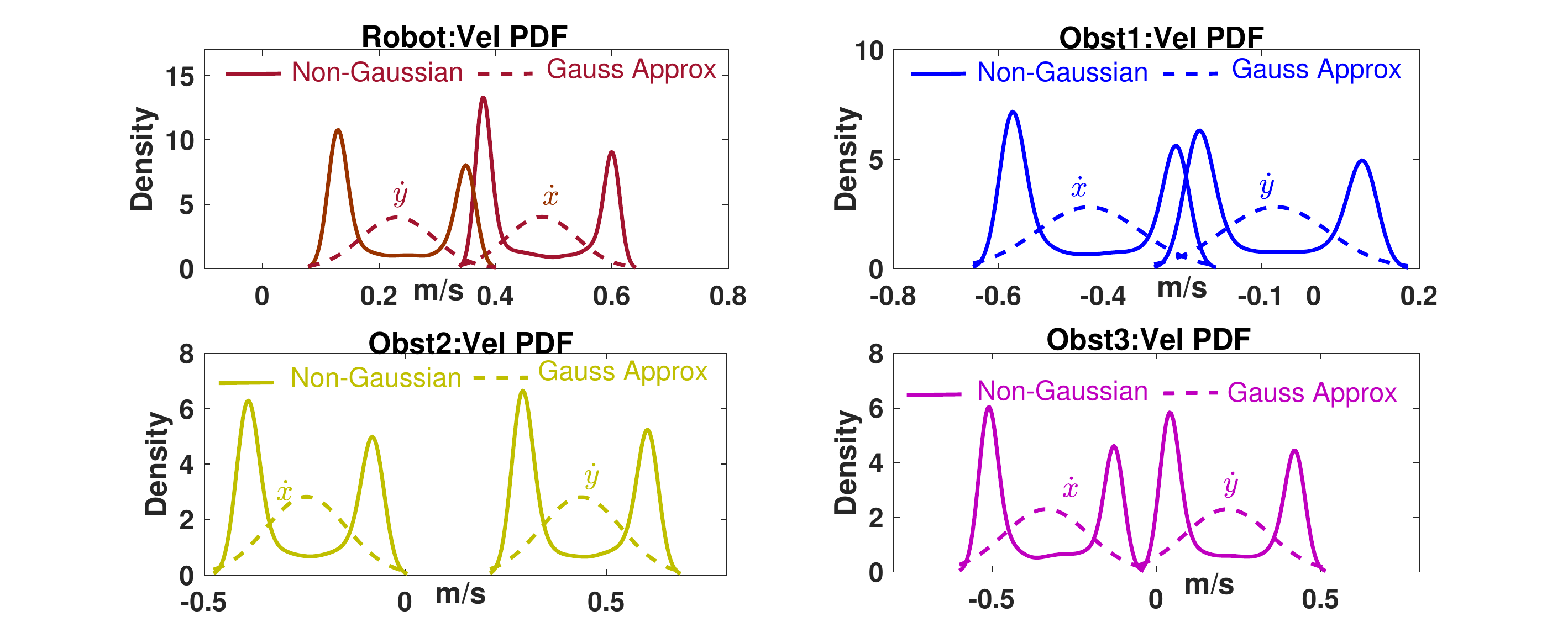}
     \label{three_obstacles_vel_kde} }
      \caption{(a): Collision avoidance scenario where the robot needs to avoid collision with three moving obstacles. (b): The position samples of robot and obstacles at some specific time instant when the robot detects imminent collision with the obstacles. (c): The uncertainty in velocities of robot and obstacles. It is clear from the plots that they are non-Gaussian in nature, the Gaussian approximations of these distributions are also shown. The main intent of displaying the Gaussian distributions is to show how poorly they approximate the original non-Gaussian distributions. }
\end{figure*}

\subsection{Collision Avoidance Results}
\subsubsection{One Obstacle Benchmark with Non-Gaussian Uncertainty}
In this benchmark, a robot tries to avoid a single moving obstacle while considering the uncertainty in its own motion and its perception of the obstacle, both of which are assumed to be Non-Gaussian with unknown distribution. The configuration of the robot and the obstacle is shown in Fig. \ref{one_obst_uncert}. The brown and the green colored samples in this figure, indicate the uncertainty in the robot and obstacle trajectories. At some specific instant, the position and velocity uncertainty are as shown in Figs.\ref{samples_non_gaussian}, \ref{kde_xdot_ydot_nongaussian} respectively. Note that the distribution indicate a typical non-Gaussian nature. 
As shown in Section \ref{obst_avoid}, the uncertainty in position and velocity can be mapped to uncertain parameters $\textbf{w}_1, \textbf{w}_2$ and consequently to functions $h_0(\textbf{w}_1, \textbf{w}_2)$, $h_1(\textbf{w}_1, \textbf{w}_2)$ and $h_2(\textbf{w}_1, \textbf{w}_2)$. We subsequently use this information to compute the collision avoidance velocity for the robot.

The solution process and results are summarized in Figs.\ref{density_nongauss_d2}-\ref{d_5_samples_nongauss}. As described previously, the solution process starts with the construction of the desired distribution $P_f^{des}$ \footnote{ Recall that the parametric form for the desired distribution or even $P_f(u)$ is not known. But for illustration purposes, we can use the 
 Kernel Density Estimation and empirical CDF methods to  graphically represent the distribution in our plots.  
}. Subsequently, we ensure that the distribution of $P_f(u)$ is similar to $P_f^{des}$ (atleast near the tail end) by choosing an appropriate $u$ and the degree of the polynomial kernel $d$. The following key points should be particularly noted from the plots. Figs.\ref{density_nongauss_d2}, \ref{density_nongauss_d3}, \ref{density_nongauss_d5} clearly show that as $d$ increases, the distributions $P_f(u)$ and $P_f^{des}$ become more alike (atleast near the tail end) and at the same time a higher portion of the mass of $P_f(u)$ gets pushed to the left of $f(.)=0$.

The increase in similarity between the two distributions is correlated with actual collision avoidance in Figs.\ref{d_2_samples_nongauss}-\ref{d_5_samples_nongauss}, wherein the position samples shown in black 
correspond to the samples of distribution $P_f(u)$ which are to the right of $f(.)=0$. The position samples shown in red correspond to samples which are to the left of $f(.)=0$. Physically, this means that if at the current instant, the robot and obstacle occupy any of the position shown in black then the robot is going to collide with the obstacle if it chooses any velocity from the distribution $(\dot{x}, \dot{y})$ and executes it after scaling it by a factor $u$. Similarly, the reverse holds for the position samples shown in red. Another important thing to note from Figs.\ref{d_2_samples_nongauss}-\ref{d_5_samples_nongauss} is that as the mass of the distribution  $P_f(u)$ gets pushed to the left of $f(.)=0$ due to an increase in $d$, the number of colliding position samples also reduces. Also, note that an increase in $d$ is observed with a simultaneous increase in $u$. This means that the robot needs to modify its forward velocity by a larger amount to maintain a high probability of collision avoidance.

\subsubsection{Three  Obstacle Benchmark with Non-Gaussian Uncertainty}
Here we consider a benchmark where the robot needs to avoid collisions with three obstacles under Non-Gaussian perception and motion uncertainty.  Fig.\ref{Three_obstacles_config} represents the configuration of the robot and the moving obstacles. At some specific time instant, Figs.\ref{three_obstacles_colliding}, \ref{three_obstacles_vel_kde} represent the uncertainty in the robot's and the obstacle's  positions and velocities.

\begin{figure*}[!tbh]
\centering
\subfigure[]{
\includegraphics[width=6.0cm,height=6cm]{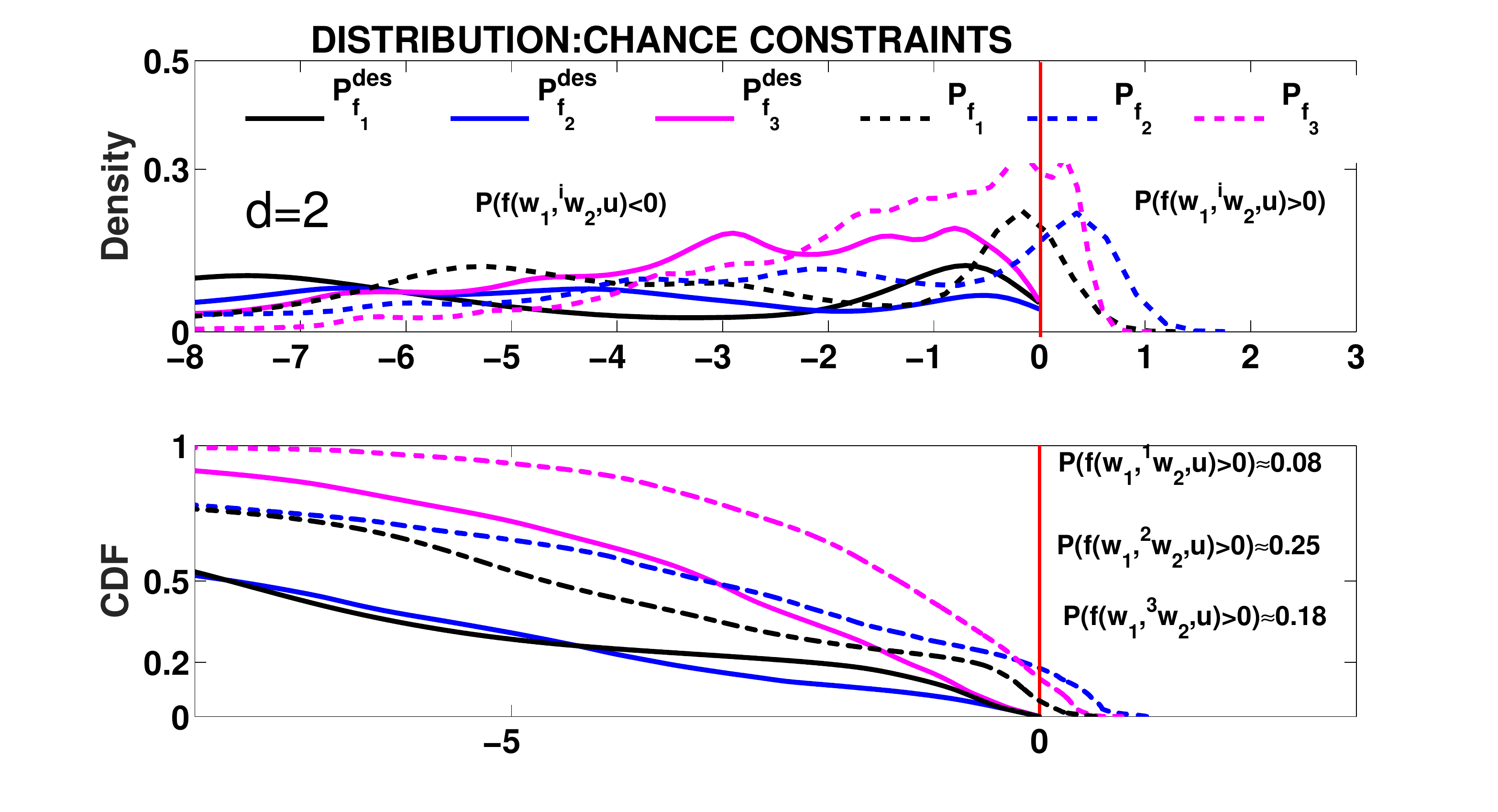}
\label{density_three_d2}
      }\hspace{-0.7cm}
\subfigure[]{
\includegraphics[width=6.0cm,height=6cm]{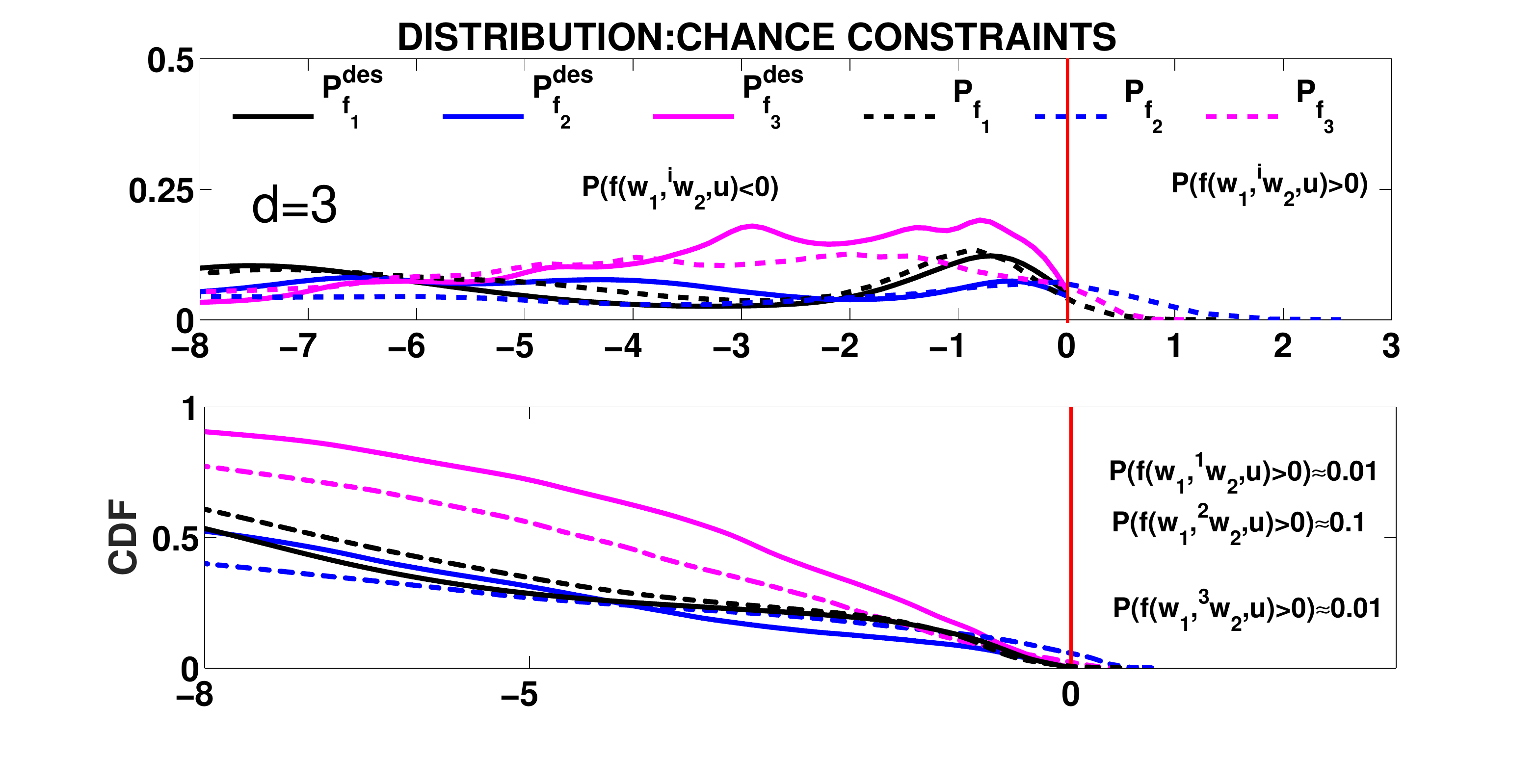}
\label{density_three_d3}
      } \hspace{-0.7cm}
      \subfigure[]{
\includegraphics[width=6.0cm,height=6cm]{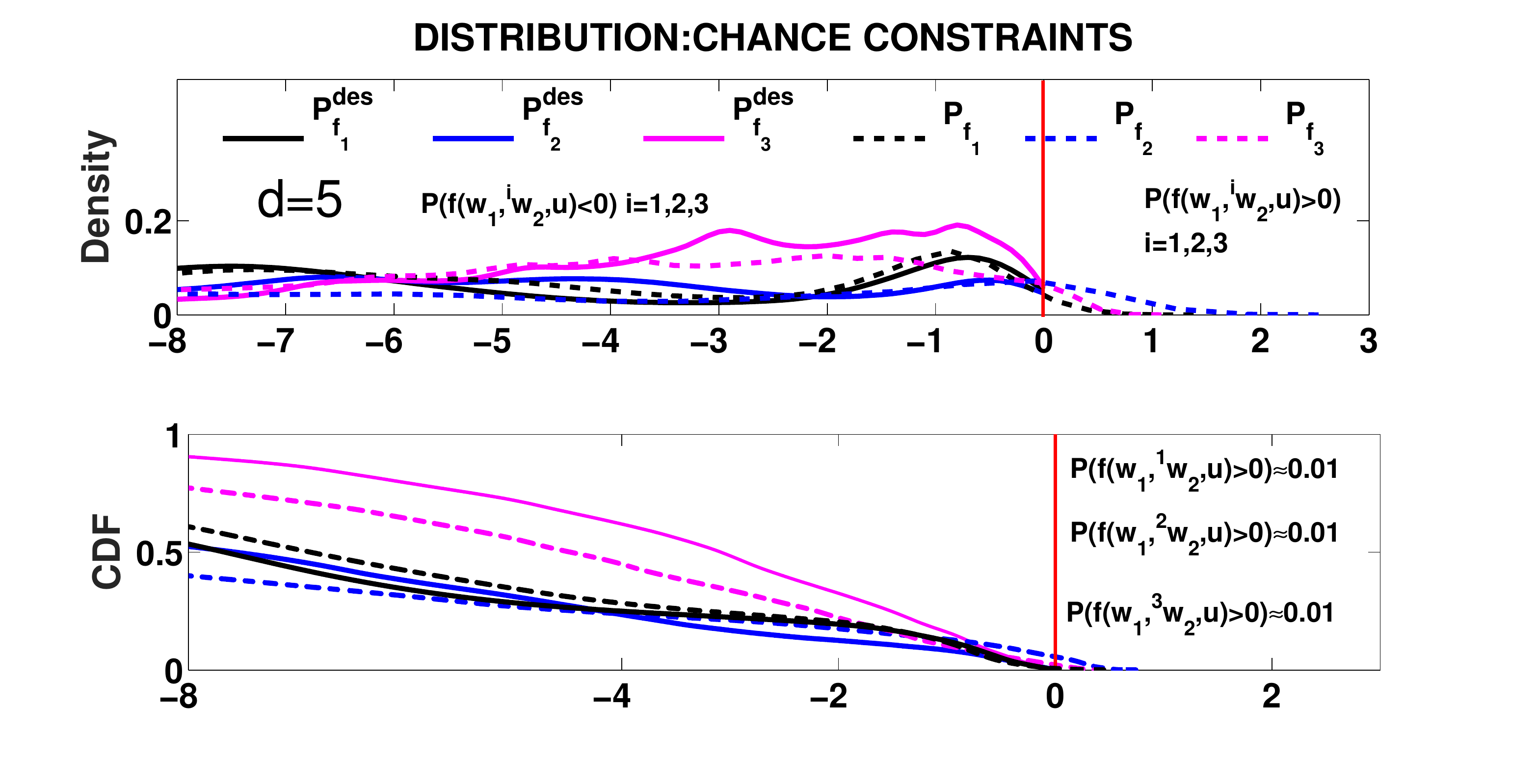}
\label{density_three_d5}
      }
       \subfigure[]{
\includegraphics[width=6.0cm,height=4.0cm]{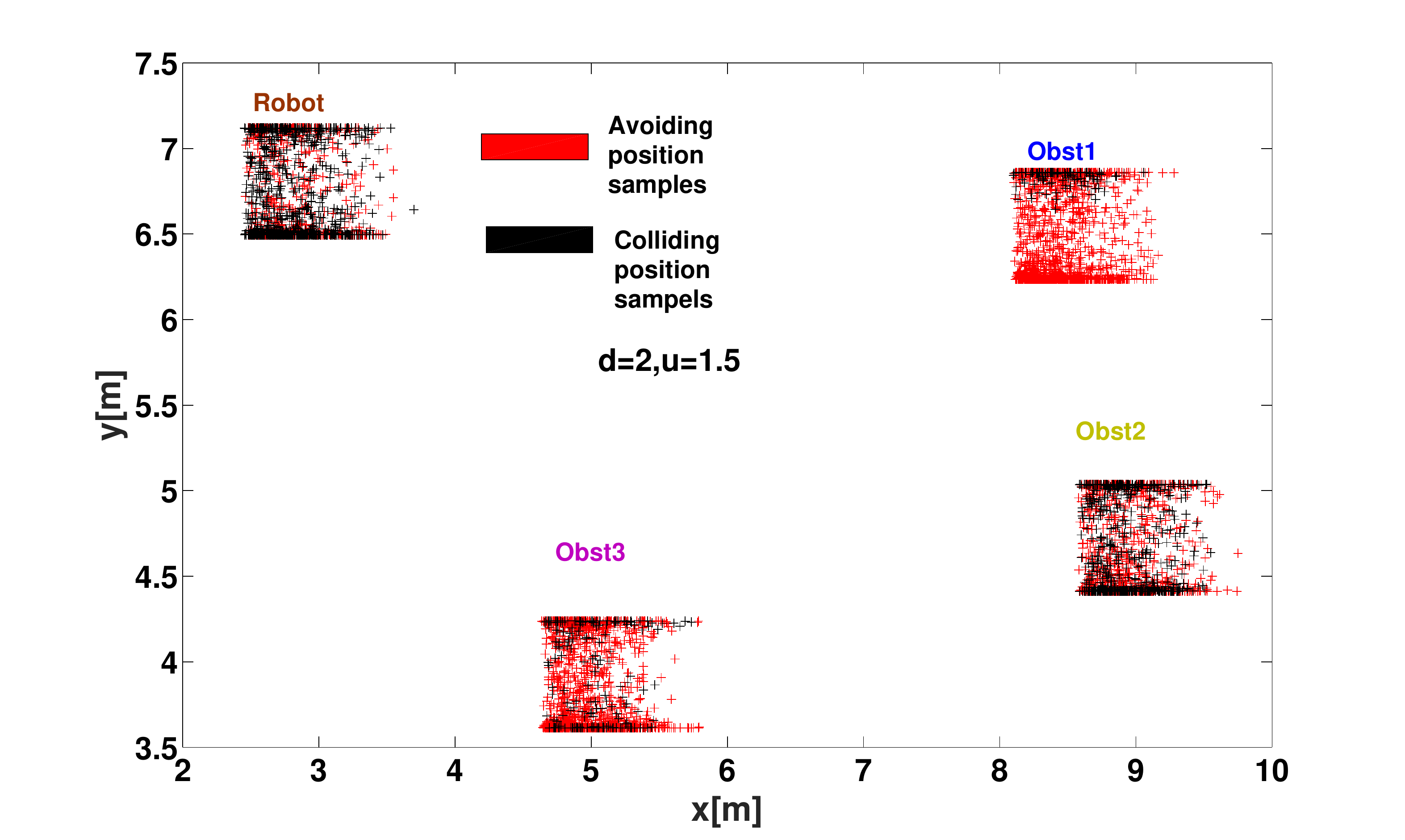}
\label{three_obstacles_d2}
      }  \hspace{-0.7cm}
       \subfigure[]{
\includegraphics[width=6.0cm,height=4cm]{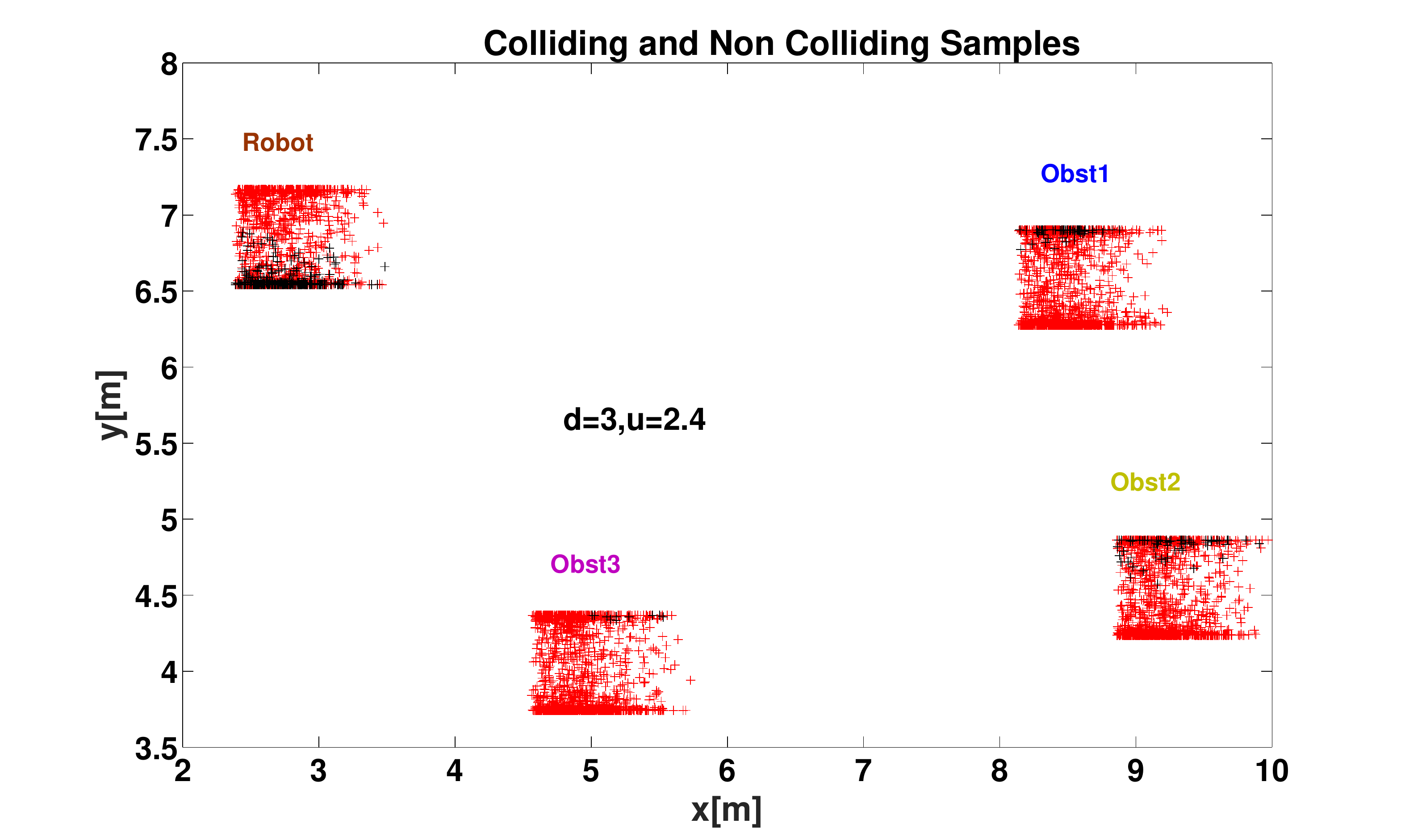}
\label{three_obstacles_d3} 
      }\hspace{-0.7cm}  
\subfigure[]{
\includegraphics[width=6cm,height=4cm]{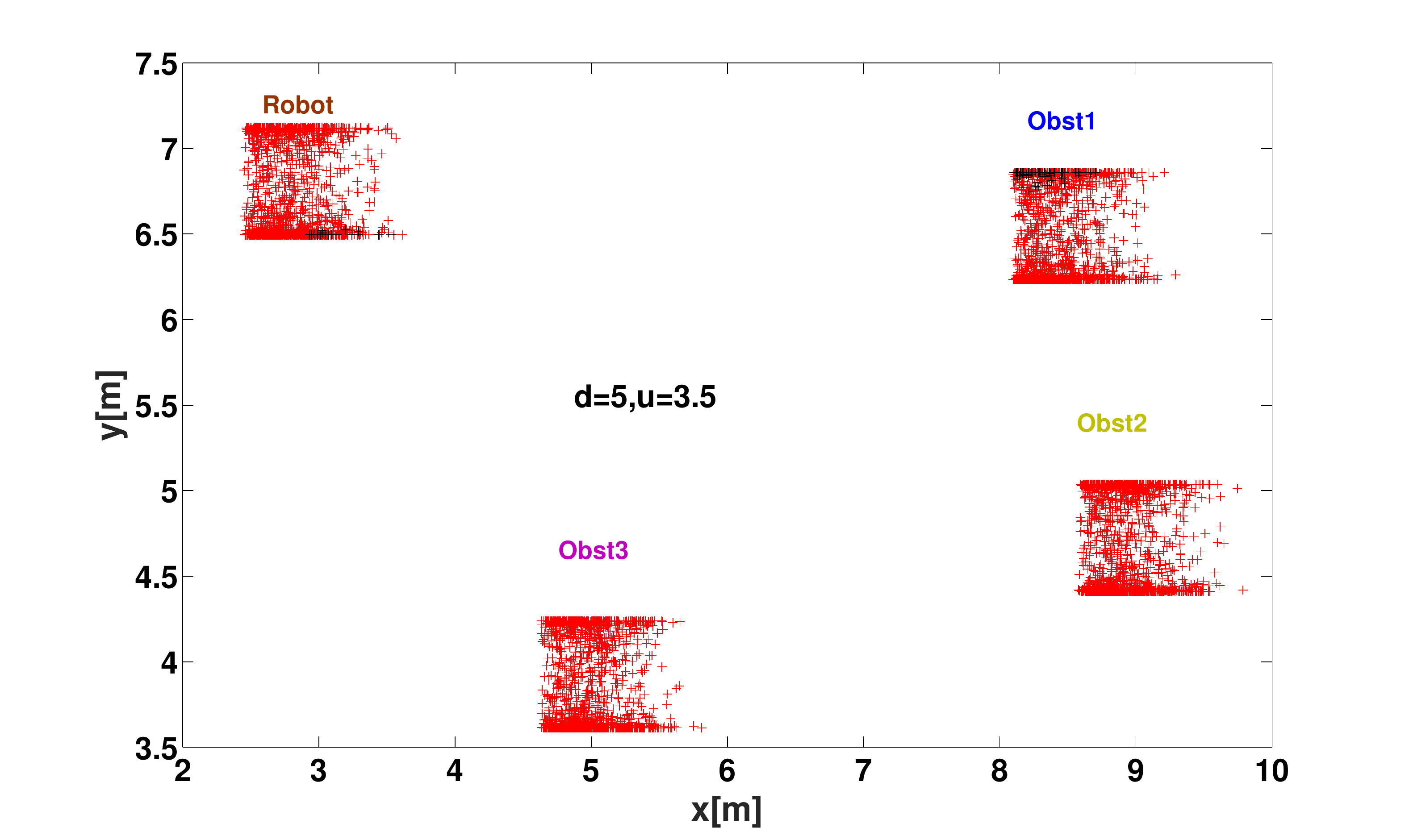}
\label{three_obstacles_d5}
      }
      \caption{Figures show the simulation results for collision avoidance with three moving obstacles shown in Fig.\ref{three_obstacles_colliding} under non-Gaussian uncertainty. In this example, we have multiple chance constraints $P(f_i(\textbf{w}_1, {^i}\textbf{w}_2, u )\leq 0)\geq \eta$ because the uncertain parameter $\textbf{w}_2$ was different for each obstacle. Thus, as shown in Figures (a), (b), (c), we need to construct three different desired distributions $P_{f_i}^{des}$ corresponding to chance constraints formulated with respect to each obstacle. As seen in previous examples, an increase in the degree of the polynomial kernel $d$ leads to the increase in the portion of the mass of $P_{f_i}$ to the left of $f(.)=0$. Figures (d), (e), (f) validate the reduction of the colliding samples with an increase in $d$.}      
\end{figure*}

\begin{figure*}
\subfigure[]{
\includegraphics[width=4.5cm,height=4.1cm]{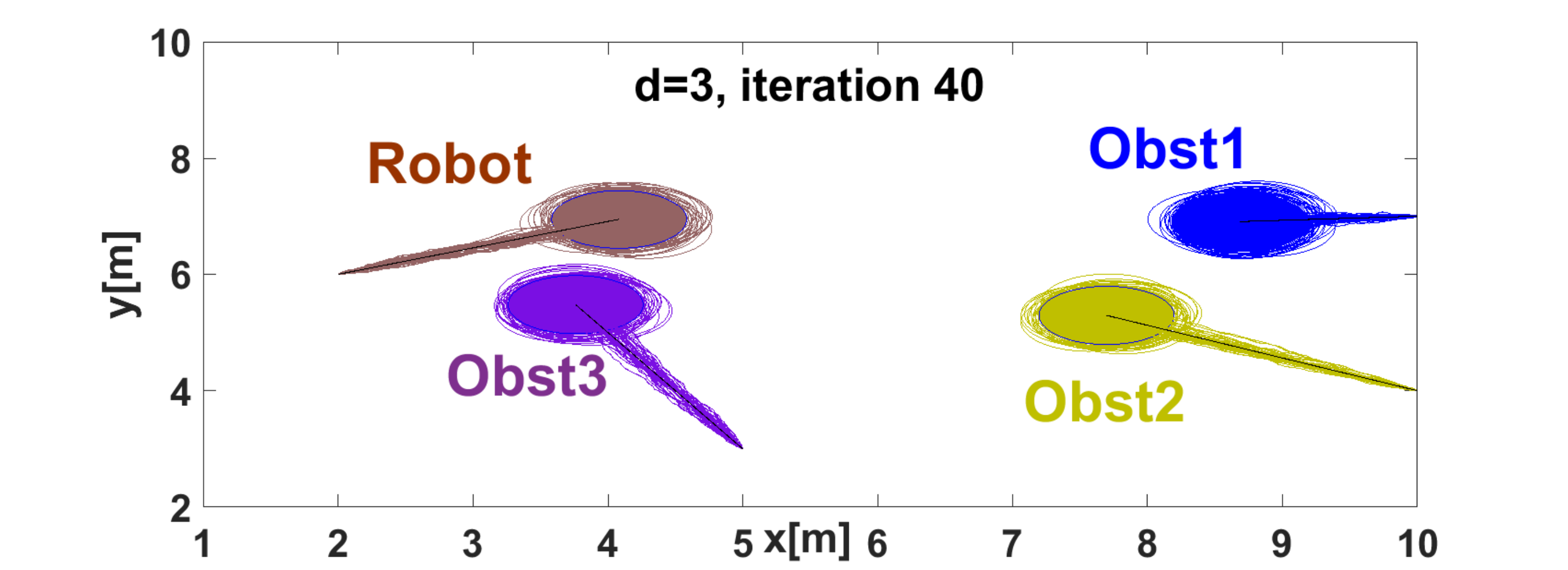}
\label{d_3_snapshot1}
      } \hspace{-0.5cm}
 \subfigure[]{
\includegraphics[width=4.5cm,height=4.1cm]{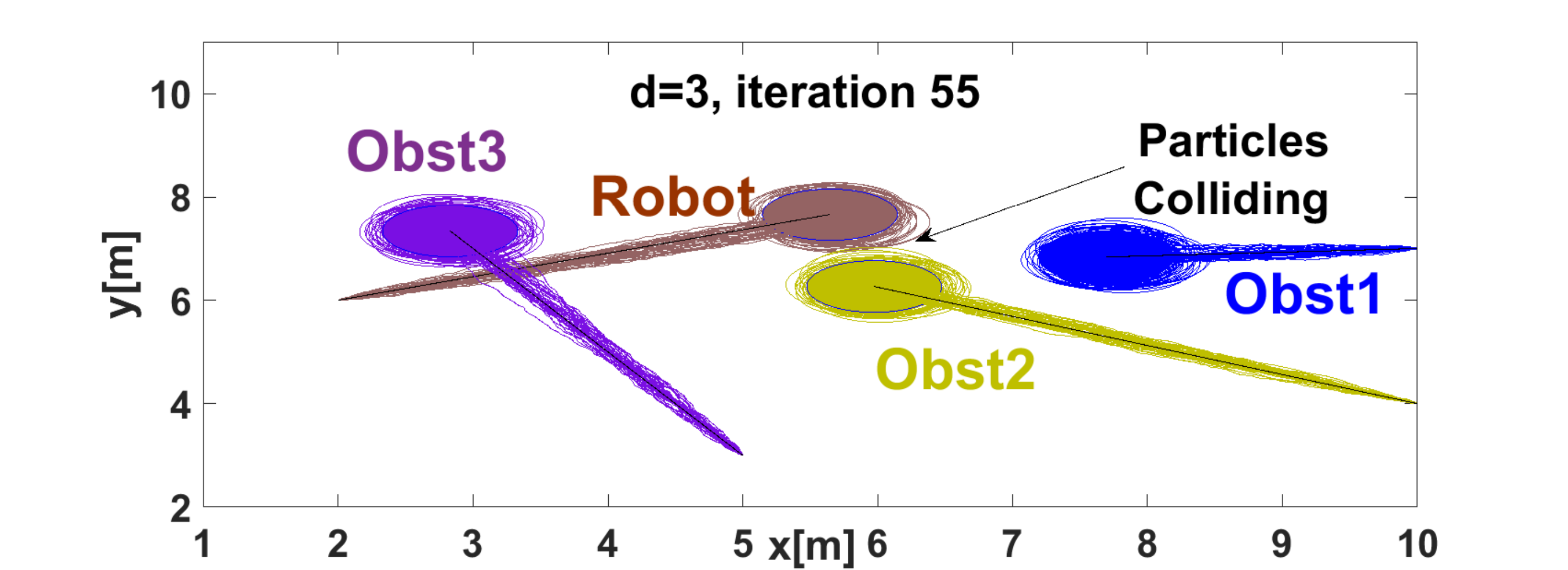}
\label{d_3_snapshot2}
      } \hspace{-0.5cm}
\subfigure[]{
\includegraphics[width=4.5cm,height=4.1cm]{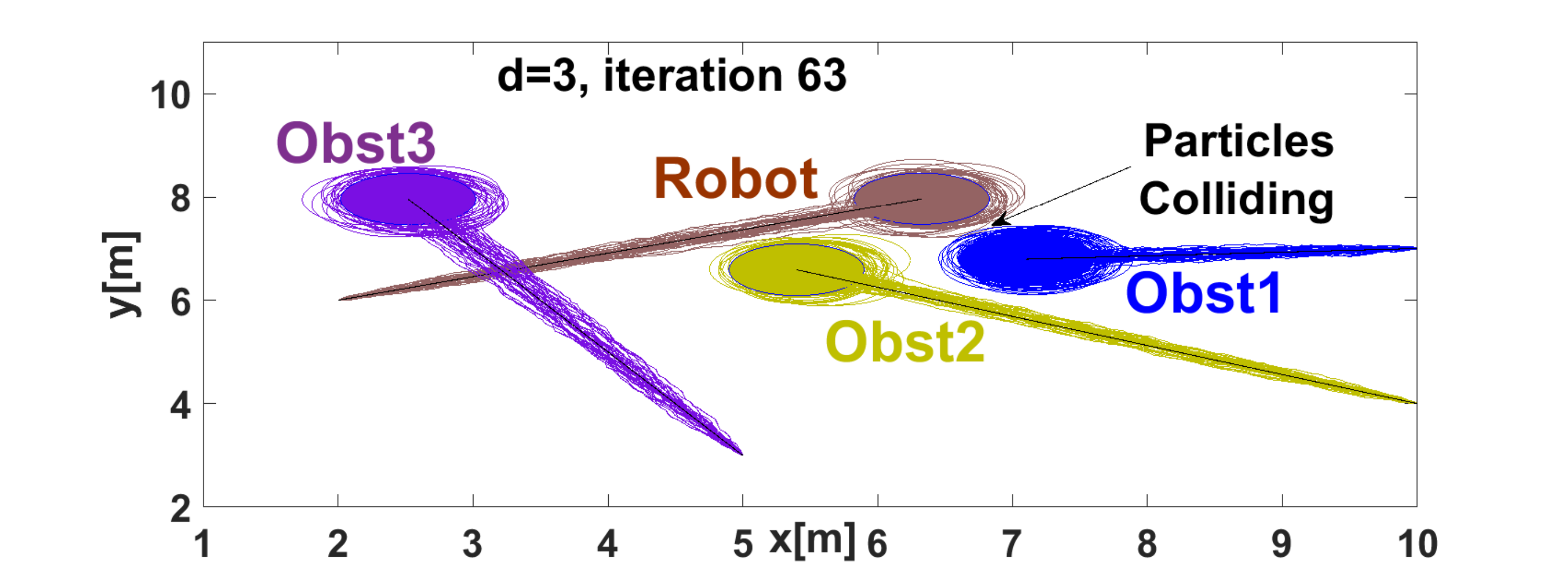}
\label{d_3_snapshot3}
      }  \hspace{-0.5cm}
 \subfigure[]{
\includegraphics[width=4.5cm,height=4.1cm]{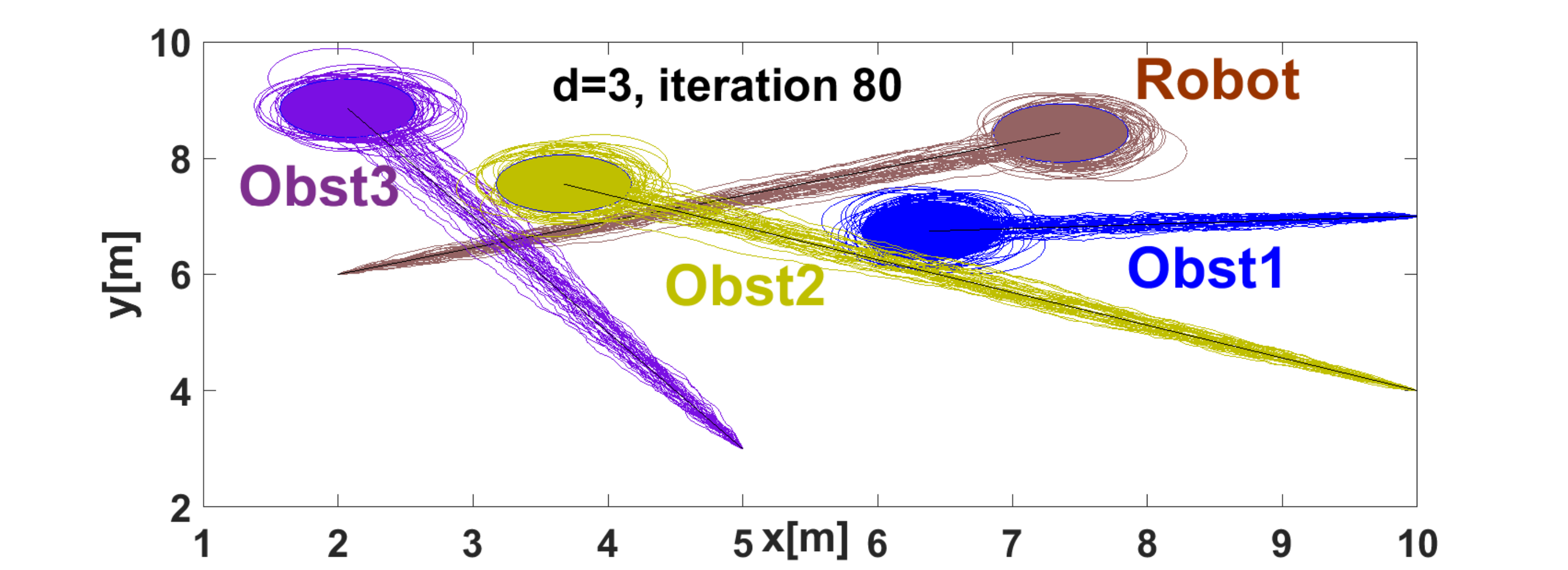}
\label{d_3_snapshot4} 
      }
\subfigure[]{
\includegraphics[width=4.5cm,height=4.1cm]{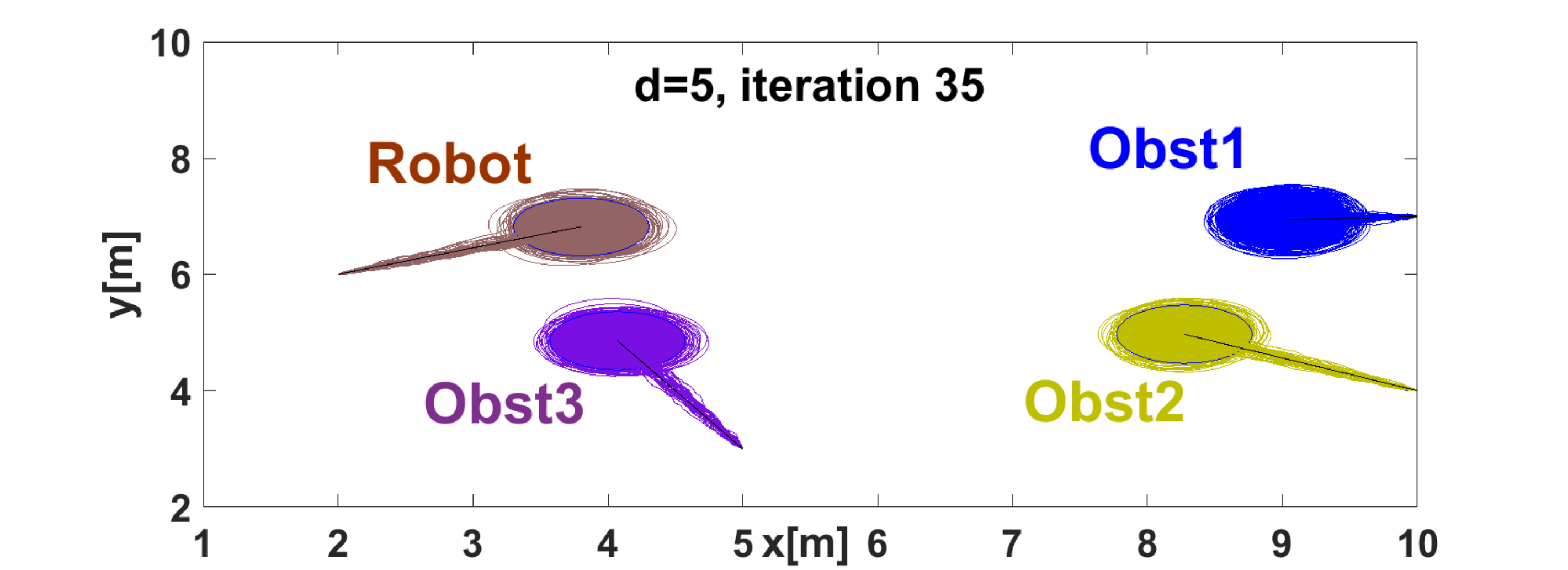}
\label{d_5_snapshot1}
      } \hspace{-0.5cm}
 \subfigure[]{
\includegraphics[width=4.5cm,height=4.1cm]{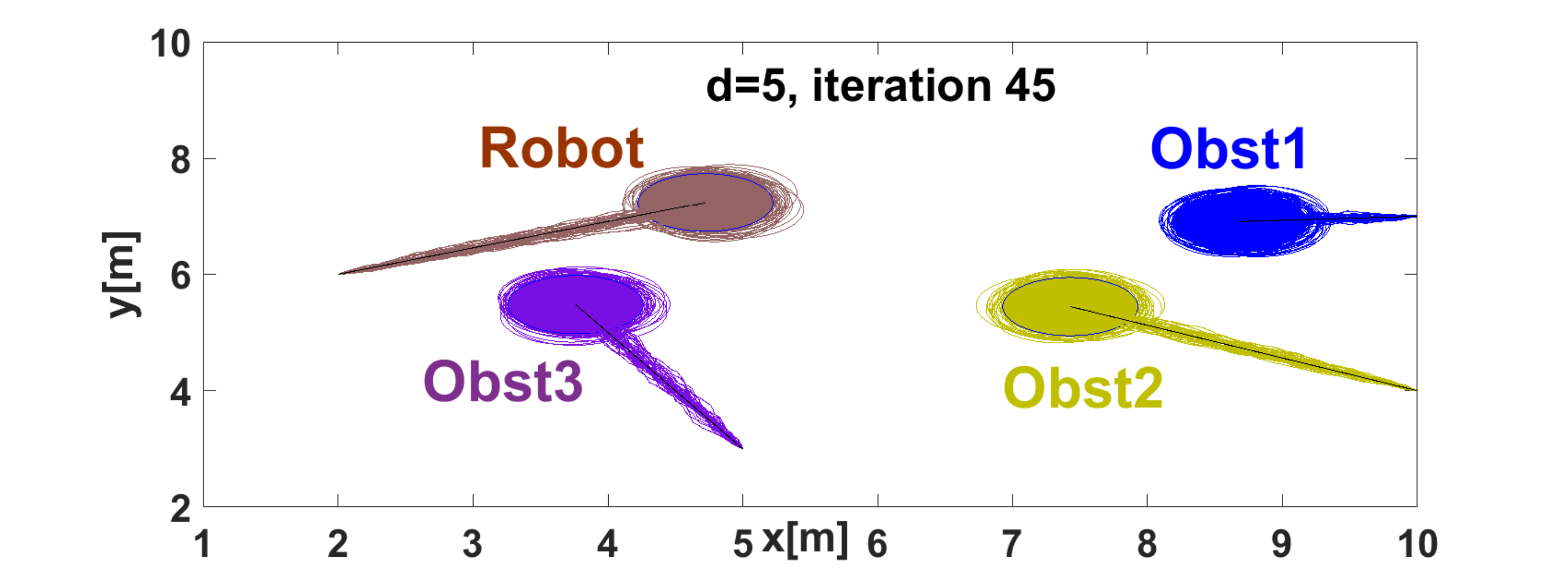}
\label{d_5_snapshot2}
      } \hspace{-0.5cm}
\subfigure[]{
\includegraphics[width=4.5cm,height=4.1cm]{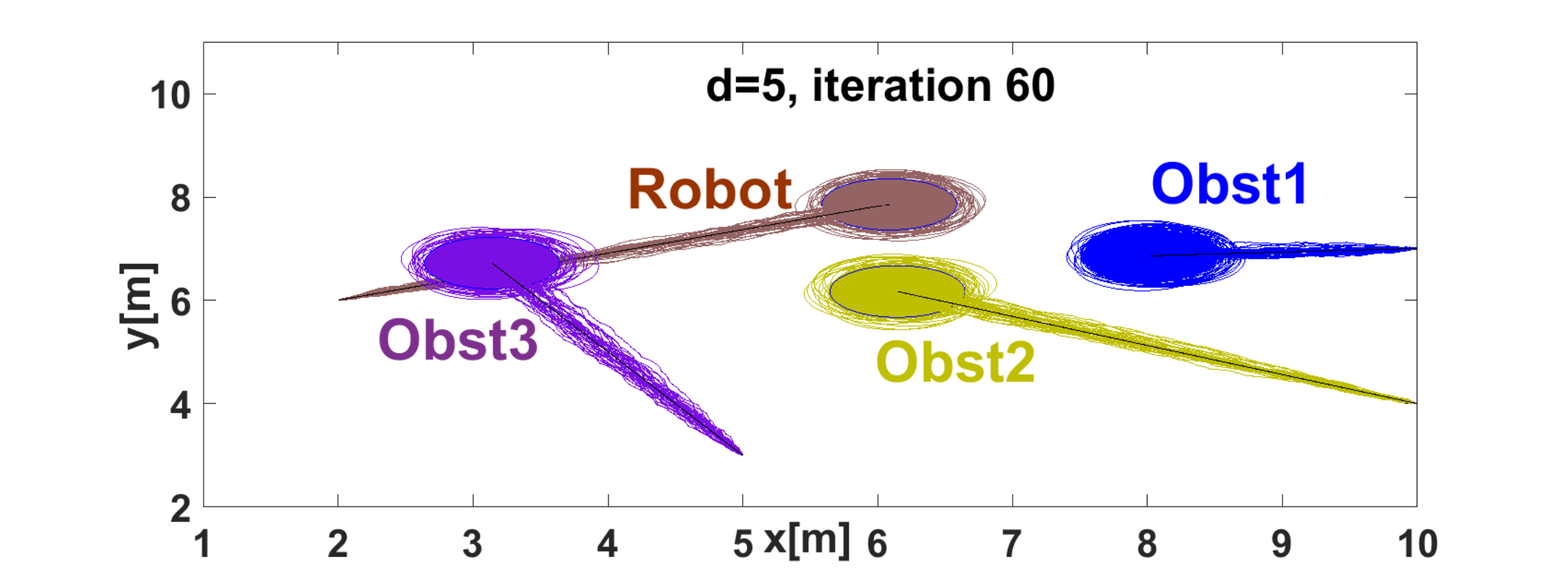}
\label{d_5_snapshot3}
      } \hspace{-0.5cm}
 \subfigure[]{
\includegraphics[width=4.5cm,height=4.1cm]{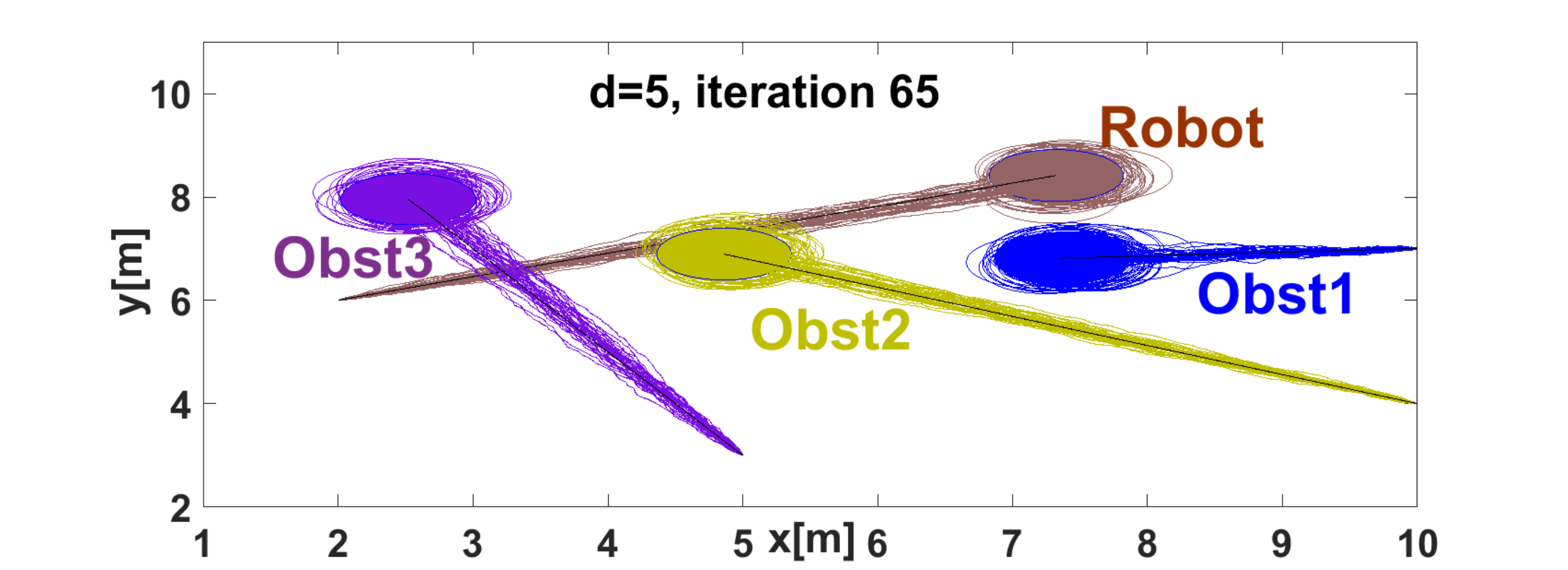}
\label{d_5_snapshot4} 
      }                 
\caption{Snapshots of collision avoidance simulation for $d=3,5$. Note how increase in $d$ results in increase in clearance between the robot and the obstacles. The increased clearance translates to improvement in probability of collision avoidance.}
\end{figure*}

Figs. \ref{density_three_d2}, \ref{density_three_d3}, \ref{density_three_d5} show the desired distribution $P_{f_i}^{des}$ constructed corresponding to chance constraints formulated with respect to each obstacle. The figures also show the distribution of $P_{f_i}(u)$ for various values of $d$. The improvement in collision avoidance probability with an increasing value of $d$ is further validated in Figs.\ref{three_obstacles_d2}, \ref{three_obstacles_d3}, \ref{three_obstacles_d5} via a comparison of the position samples from where the robot can either collide with (black) or avoid (red) the obstacles. Snapshots from collision avoidance simulations are shown in Figures \ref{d_3_snapshot1}-\ref{d_5_snapshot4}. It is easy to relate these snapshots to the position samples from figures \ref{three_obstacles_d2}, \ref{three_obstacles_d3}, \ref{three_obstacles_d5}. As the value of $d$ increases, the robot chooses a velocity that results in more and more clearance with the obstacles. This is what results in reduction of colliding samples in Figs.\ref{three_obstacles_d2}, \ref{three_obstacles_d3}, \ref{three_obstacles_d5}.

\begin{table*}[!h]
\centering
\caption{Table summarizing sample complexity for collision avoidance application. }
\begin{tabular}{|p{4cm}|p{5cm}|p{5cm}|}\hline
Approach & $P(f(\textbf{w}_1,\textbf{w}_2,u)<0)\approx 0.7$ & $P(f(\textbf{w}_1, \textbf{w}_2,u)<0) \approx 0.9$ \\ \hline
Scenario & $\textbf{w}_1, \textbf{w}_2=200$ & $\textbf{w}_1, \textbf{w}_2=500$      \\ \hline
  SAA \cite{sample_average_shapiro}   & $\textbf{w}_1, \textbf{w}_2 = 100$ & $\textbf{w}_1, \textbf{w}_2=200$  \\ \hline
  $E[f(\textbf{w}_1, \textbf{w}_2, u)]+\epsilon \sqrt{Var[f(\textbf{w}_1, \textbf{w}_2, u)]}\leq0$
 &$\textbf{w}_1, \textbf{w}_2 = 800$ & $ \textbf{w}_1, \textbf{w}_2=800$ \\ \hline
 Proposed RKHS embedding    & $\textbf{w}_1, \textbf{w}_2 =20$, $\widetilde{\textbf{w}}_1, \widetilde{\textbf{w}}_2 = 20$ & $\textbf{w}_1, \textbf{w}_2 = 40$, $\widetilde{\textbf{w}}_1, \widetilde{\textbf{w}}_2 = 20$ \\ \hline
\end{tabular}
\label{sample_complexity_0.9collavoid}
\end{table*}

\subsubsection{Comparative Results on Collision Avoidance}

Table \ref{sample_complexity_0.9collavoid} shows a comparison of the number of samples required by different approaches to compute an optimal solution such that the chance constraints are satisfied with a specified $\eta$.The following points can be noted from the table

\begin{itemize}
\item As expected, a naive implementation of the scenario approach shows the worst sample complexity. For $\eta \approx 0.7$, we required $200$ samples each of $\textbf{w}_1, \textbf{w}_2$ leading to a grid of size $4*10^4$.
For $\eta \approx 0.9$, we required $500$ samples each of $\textbf{w}_1, \textbf{w}_2$.

\item The SAA approximation proposed in \cite{sample_average_shapiro} required a sample size  almost half of that required by the scenario approach.  For $\eta \approx 0.7$, we needed $100$ samples each of $\textbf{w}_1, \textbf{w}_2$. This requirement increased to $200$ for $\eta \approx 0.9$.

\item The approach of \cite{musigma1}, \cite{musigma2} which is based on surrogate constraints \ref{meanvar} shows an interesting trend. The sample complexity is worse than scenario approach for $\eta \approx 0.7$. However, the sample size does not vary with $\eta$. This is because the samples of the uncertain parameters are used to obtain an estimate of $E[f(\textbf{w}_1, \textbf{w}_2, u)]$ and $\sqrt{Var[f(\textbf{w}_1, \textbf{w}_2, u)]}$ and importantly, this estimation is independent of $\eta$.

\item As can be seen from Table \ref{sample_complexity_0.9collavoid}, our proposed formulation based on RKHS embedding has significantly better sample complexity than all the above discussed approaches. It required $20$ samples each of $\textbf{w}_1, \textbf{w}_2$ to construct a reasonable estimate of the desired distribution. An additional $20$, $40$ samples were required to construct the RKHS embedding based reformulations at $\eta \approx 0.7$ and $\eta \approx 0.9$ respectively.  

\end{itemize}

\begin{figure*}[!h]
\centering
\subfigure[]{
\includegraphics[width=9.0cm,height=5cm]{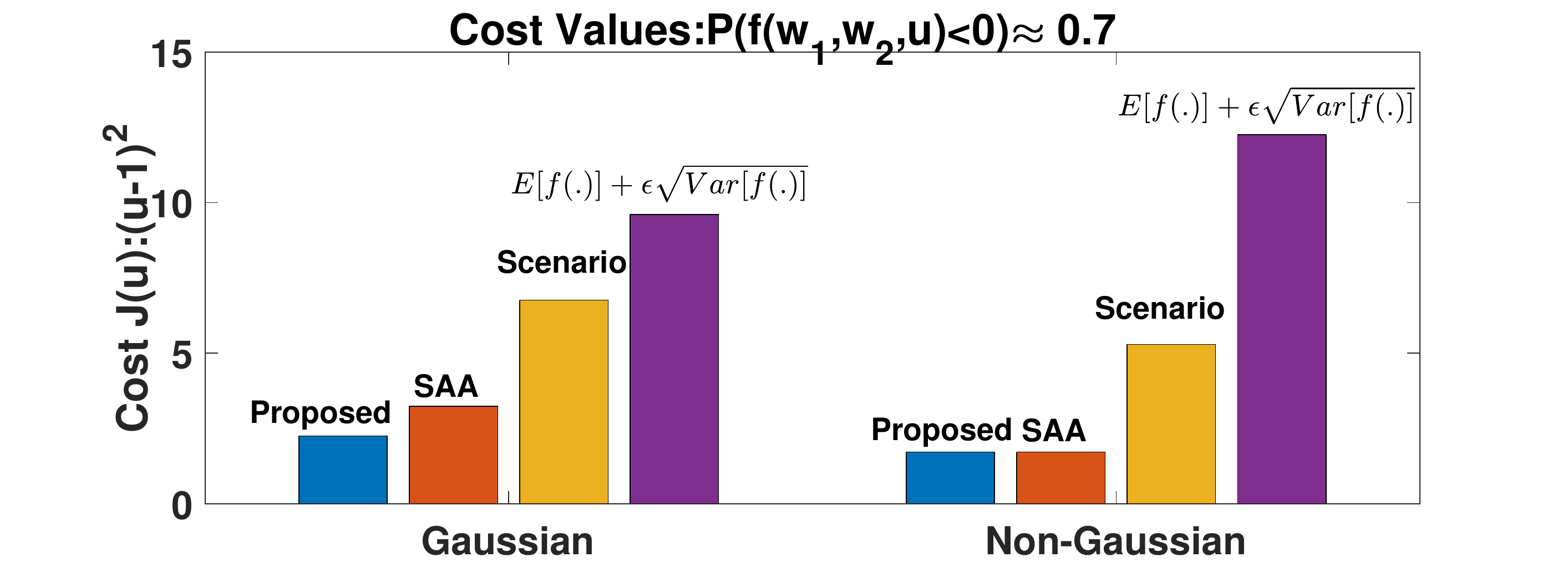}
\label{cost_comp_collavoid_0.7}
      }\hspace{-0.7cm}
 \subfigure[]{
\includegraphics[width=9.0cm,height=5cm]{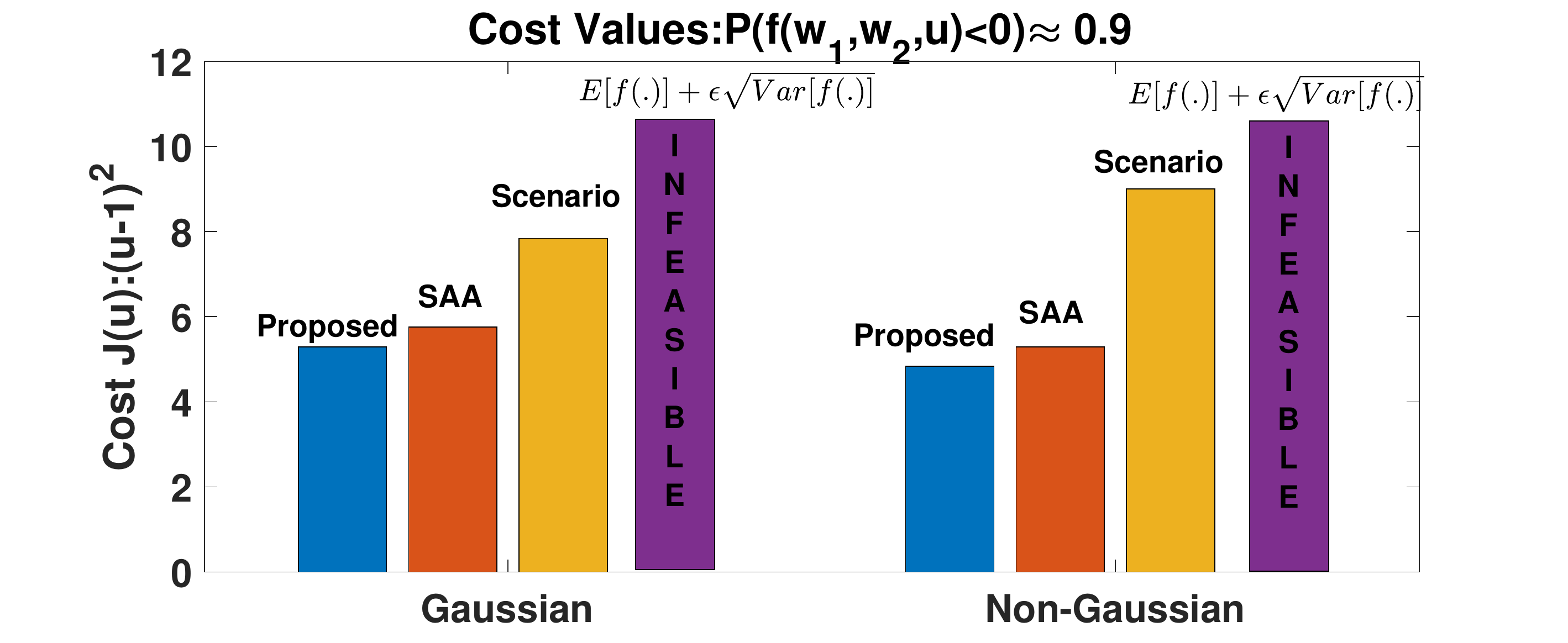}
\label{cost_comp_collavoid_0.8}
      }
    \caption{Average Optimal cost obtained with different methods for collision avoidance application observed across 20 different problem instances. Our RKHS formulation consistently results in lower cost solutions. Furthermore, the approach based on surrogate constraints (\ref{meanvar}) often runs into infeasibility at higher $\eta$. }
\end{figure*}

Figs.\ref{cost_comp_collavoid_0.7}, \ref{cost_comp_collavoid_0.8} compare the optimal cost obtained through different formulations. The following important observations can be drawn from it

\begin{itemize}
\item Our proposed formulation results in lower cost solutions than approaches based on scenario approximation and surrogate constraints (\ref{meanvar}) \cite{musigma1}, \cite{musigma2}. The difference is more pronounced for non-Gaussian uncertainty and at higher $\eta$. In fact, at a higher $\eta$, approach based on (\ref{meanvar}) often runs into infeasibility.

\item Interestingly, the SAA approach of \cite{sample_average_shapiro} result in very similar costs to those of our proposed formulation for both Gaussian and non-Gaussian uncertainty. This is not surprising as SAA proposed in \cite{sample_average_shapiro} is indeed a very tight approximation of the chance constraints.

\end{itemize}

\subsection{Path Tracking  Results for a 2 link Manipulator}
\begin{figure*}
\centering
\subfigure[]{     
\includegraphics[width=8.5cm,height=5cm]{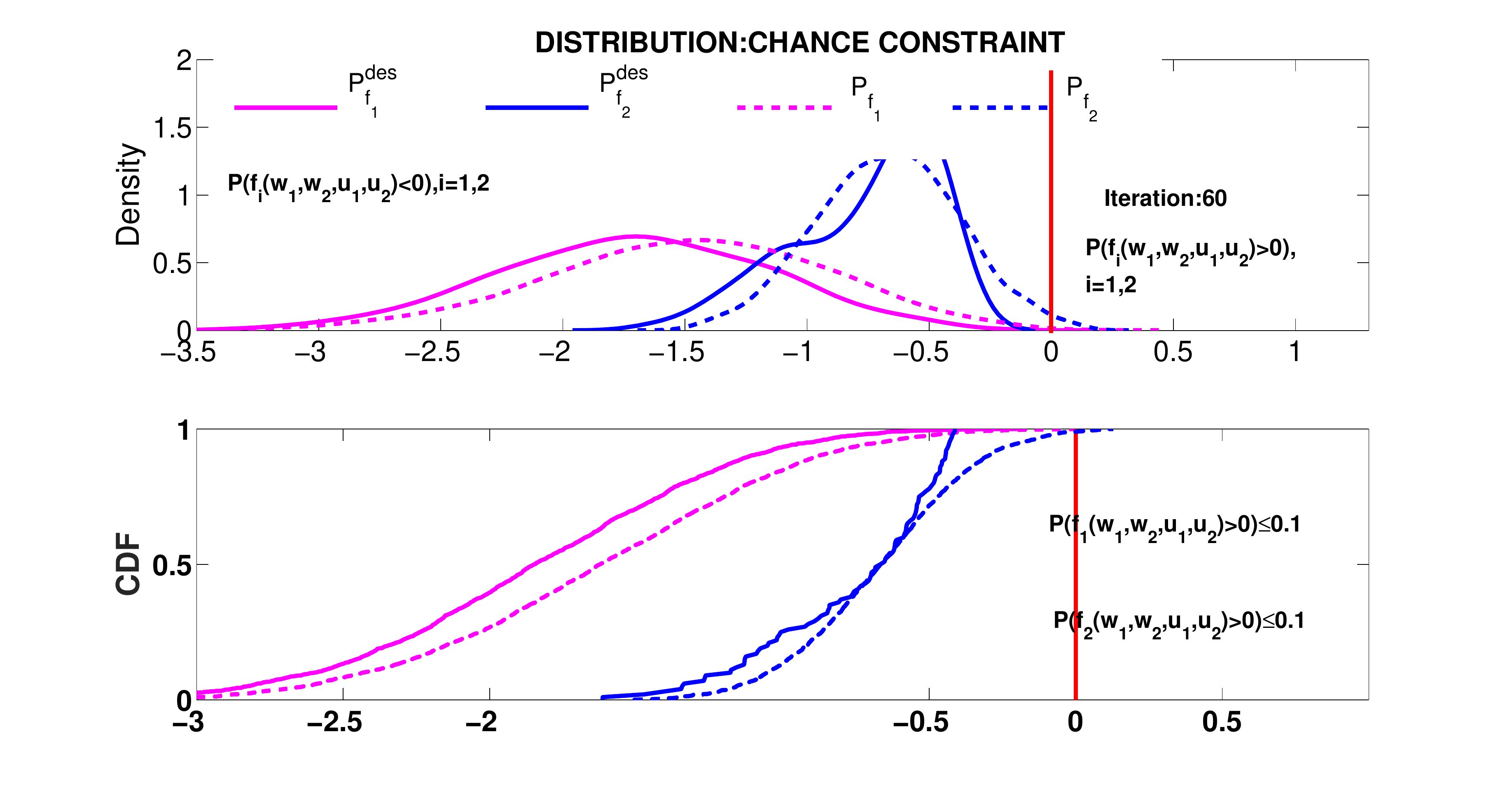}\label{kde_medium_kme1}
}\hspace{-0.7cm}
\subfigure[]{     
\includegraphics[width=8.5cm,height=5cm]{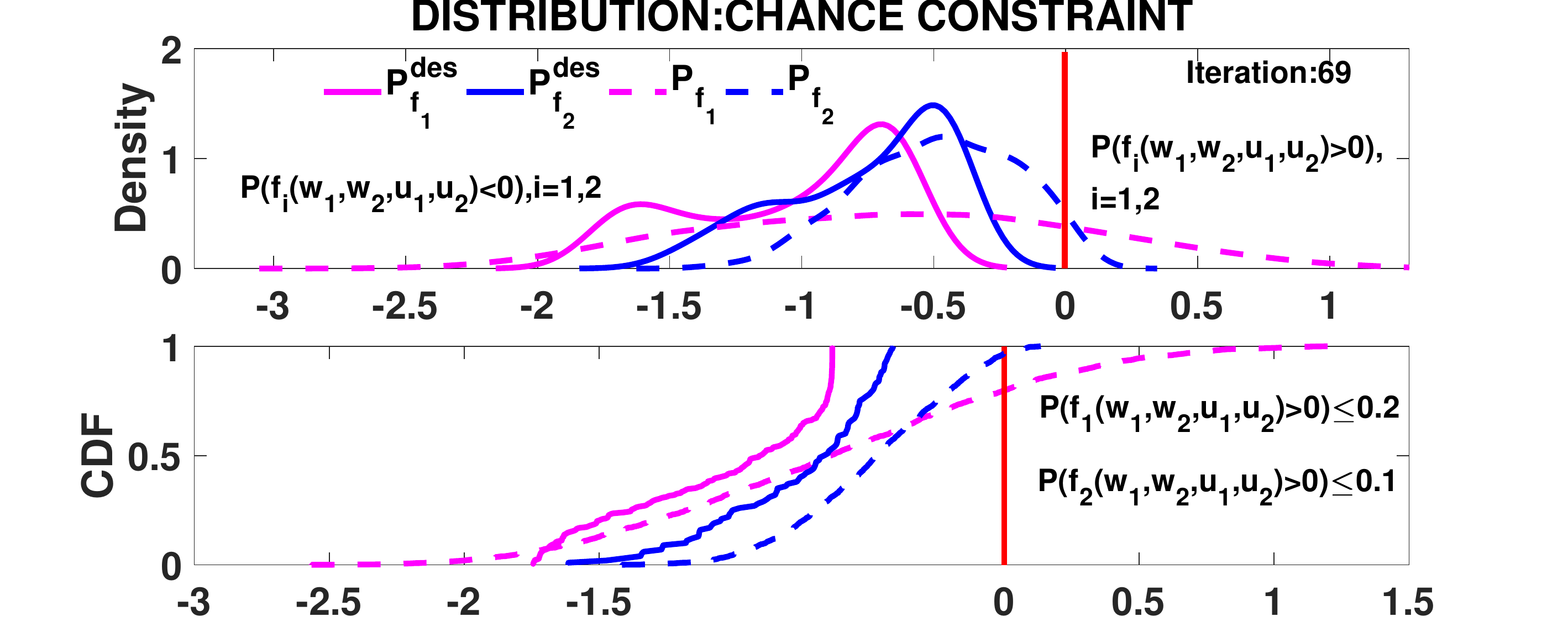}\label{kde_medium_kme}
}
\subfigure[]{
\includegraphics[width=9.0cm,height=5cm]{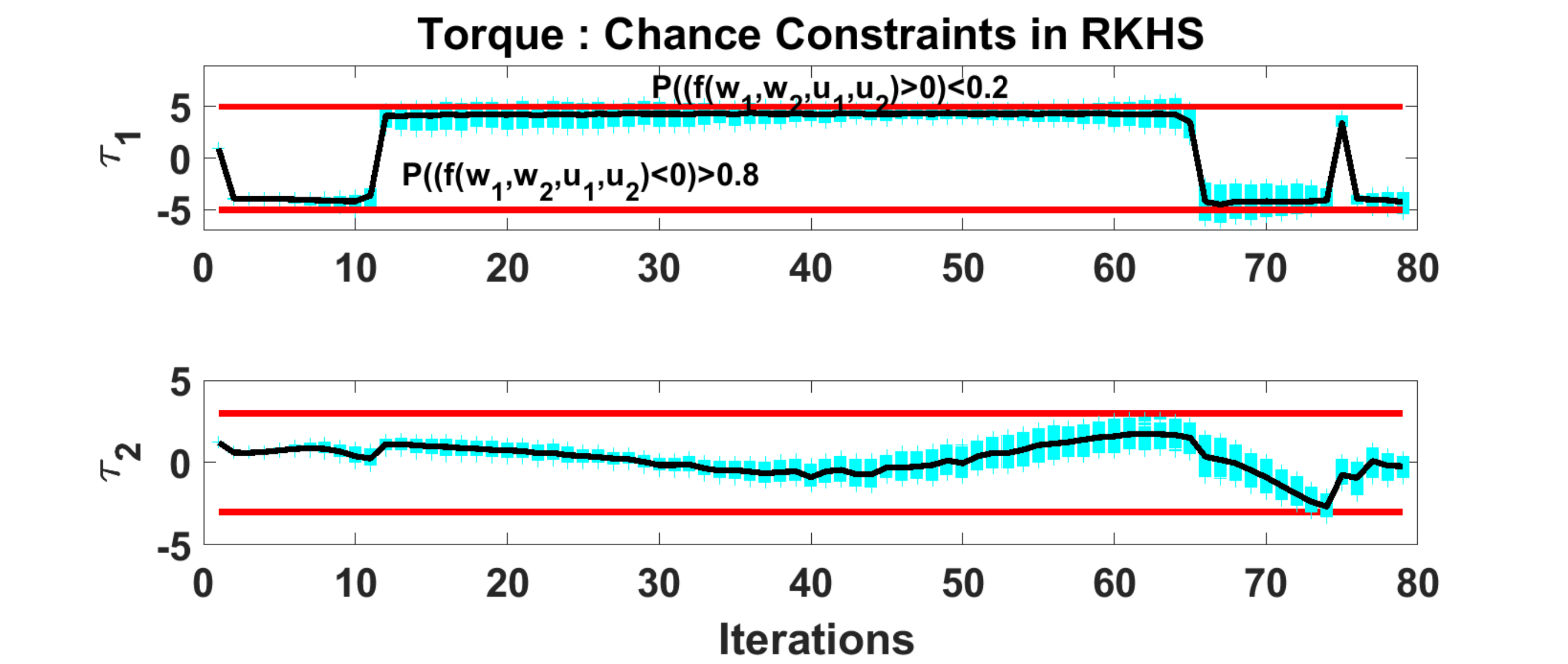}\label{tor_medium_kme}
       }\hspace{-0.7cm}
\subfigure[]{
\includegraphics[width=8.5cm,height=5cm]{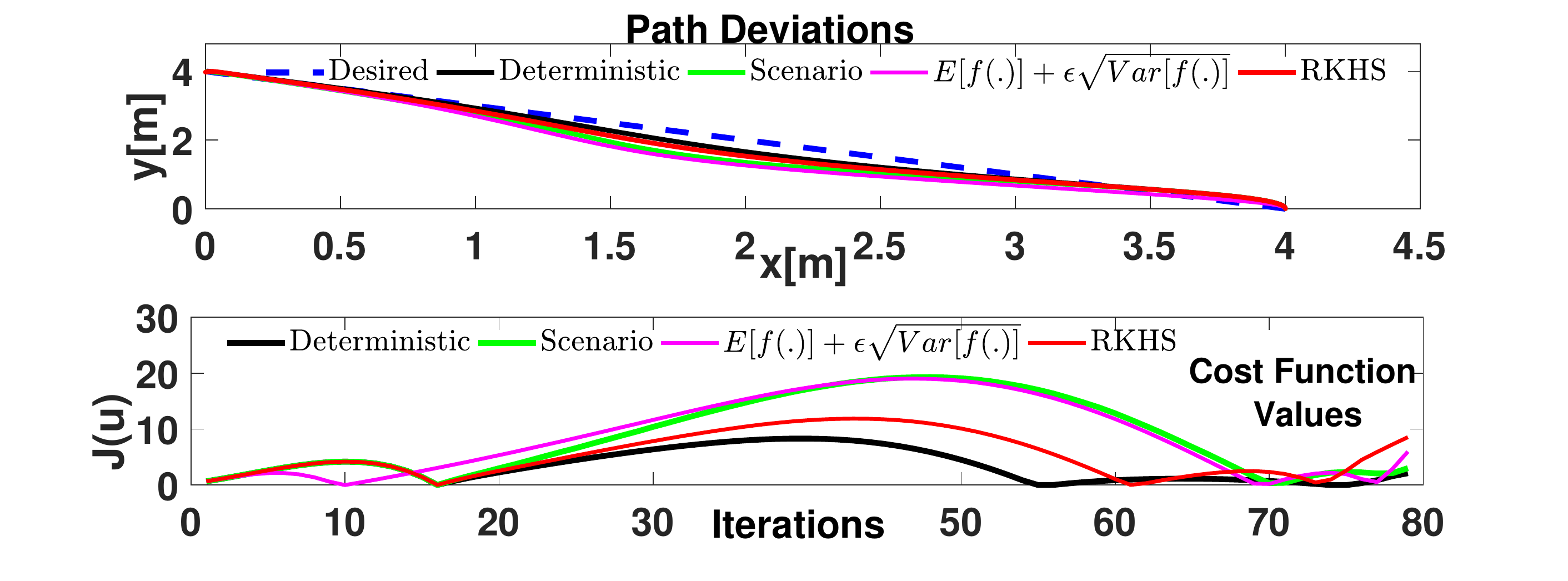}\label{cost_plots_medium}
      }
\caption{The simulation results for inverse dynamics based path tracking under non-Gaussian uncertainty (Fig.\ref{man_uncert}). In this example, we repeatedly solve the optimization (\ref{cost_reform_invdyn})-(\ref{feasible_reform_invdyn}), formulated with polynomial kernel with $d=2$. At each iteration, we need to construct a desired distribution corresponding to each chance constraint. Figures (a) and (b) show the desired distribution constructed at iteration 60 and 69. The figures also show the distribution of $P_{f_i}(.)$ for $u_1, u_2$ obtained as a solution to (\ref{cost_reform_invdyn})-(\ref{feasible_reform_invdyn}). Figure (c) shows the torque plots. The solid black lines represent the mean torque values while the cyan lines show the uncertainty around it. Figure (d) shows the tracking performance in terms of path deviation and cost plot. Refer to the text for further insight.  }
\end{figure*}

\noindent 
Recall that in this application, we repeatedly solve the chance constrained  optimization (\ref{cost_chance_invdyn})-(\ref{accbound_chance_invdyn}) or rather the reformulation of it (\ref{cost_reform_invdyn})-(\ref{feasible_reform_invdyn})
and evolve the joint angles and velocities according to the computed acceleration control input at each iteration. Moreover, we have multiple chance constraints $P(f_i(\textbf{w}_1, \textbf{w}_2, u_1, u_2)\leq 0)\geq \eta$ and thus, a desired distribution $P^{des}_{f_i}$ needs to be constructed corresponding to each of them. Fig.\ref{kde_medium_kme1}, \ref{kde_medium_kme} show the  distributions $P^{des}_{f_i}$ and $P_{f_i}(.)$ (for one of the chance constraints) at iteration 60 and 69 for $d=2$. Fig.\ref{tor_medium_kme} shows the torque values obtained at each iteration. The lines in black represent the mean torque values while the cyan shows the uncertainty around it in the form of samples. Fig.\ref{cost_plots_medium} shows the tracking performance in terms of path deviation and optimal cost values at each iteration.

Comparing $P_{f_i}(.)$ and $P_{f_i}^{des}$ at both the iterations, it can be seen  at iteration 60, the tails of the two distributions are more closely matched and as a result, a larger portion of $P_{f_i}(.)$ lies to the left of $f(.)=0$. A direct consequence of this can be observed in the torque plots. At iteration 60, we observe fewer samples of torque that violate the torque bounds compared to what we observe at iteration 69.  

\subsubsection{Comparative Results for Path Tracking}
\begin{figure}[!t]
\centering
\includegraphics[width= 8.35cm, height=4.3cm]{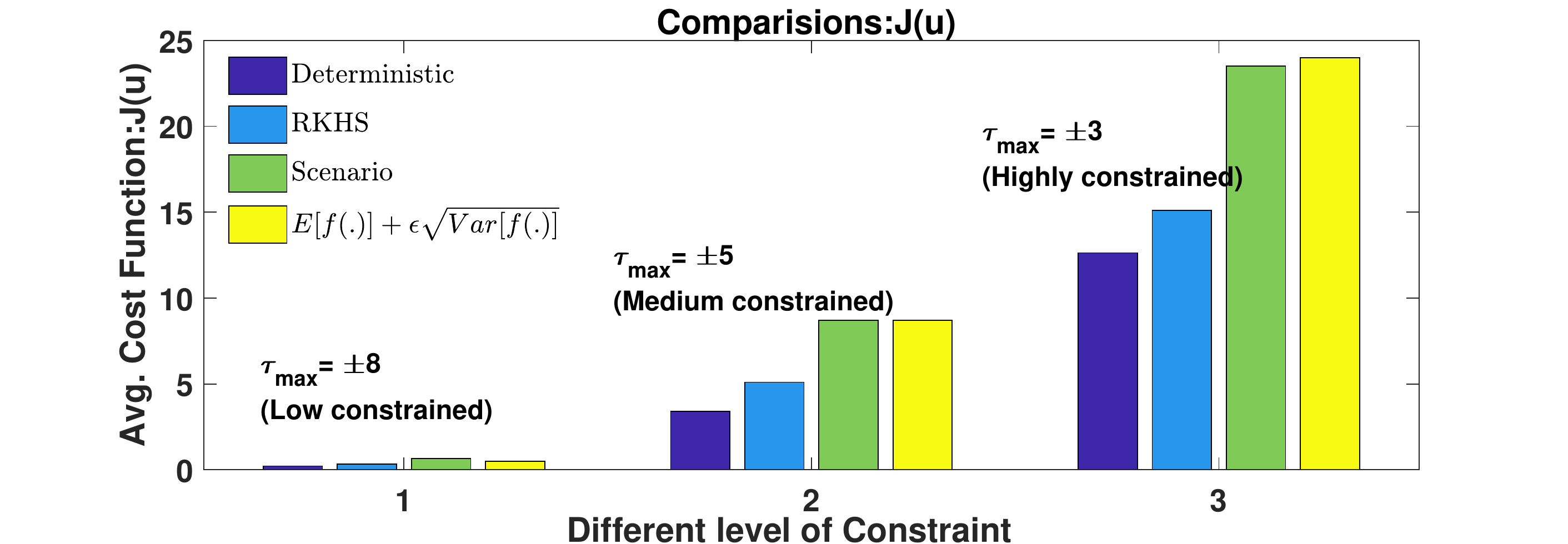}
\caption{Comparisons of optimal cost obtained for path tracking application with different approaches. A lower cost directly translates to better tracking.}
\label{cost_compare_manipulator}
\end{figure}

\begin{figure*}[!h]
\begin{center}
   \begin{tabular}{|c|c|}
   \hline
        Low constrained($\tau_{Max}=\pm 8$) & High Constrained ($\tau_{Max}=\pm 3$) \\ \hline
       \includegraphics[width=8.5cm,height=5.0cm]{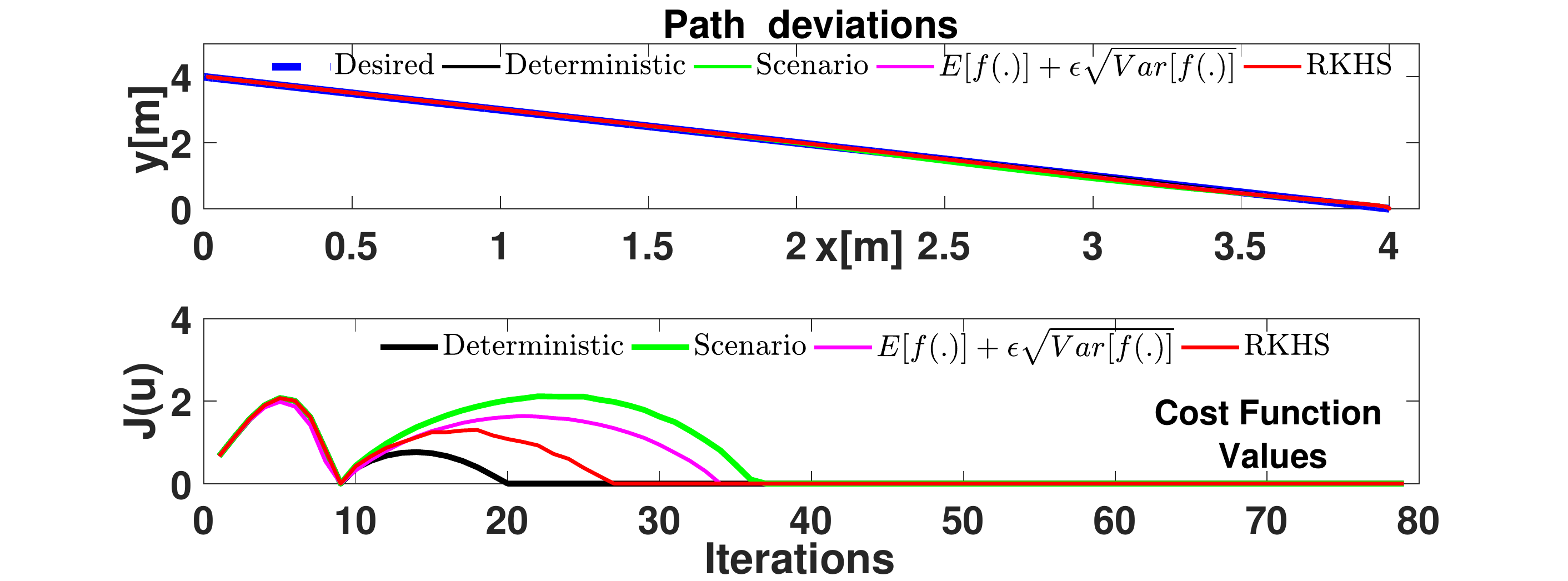}&
        \includegraphics[width=8.5cm,height=5.0cm]{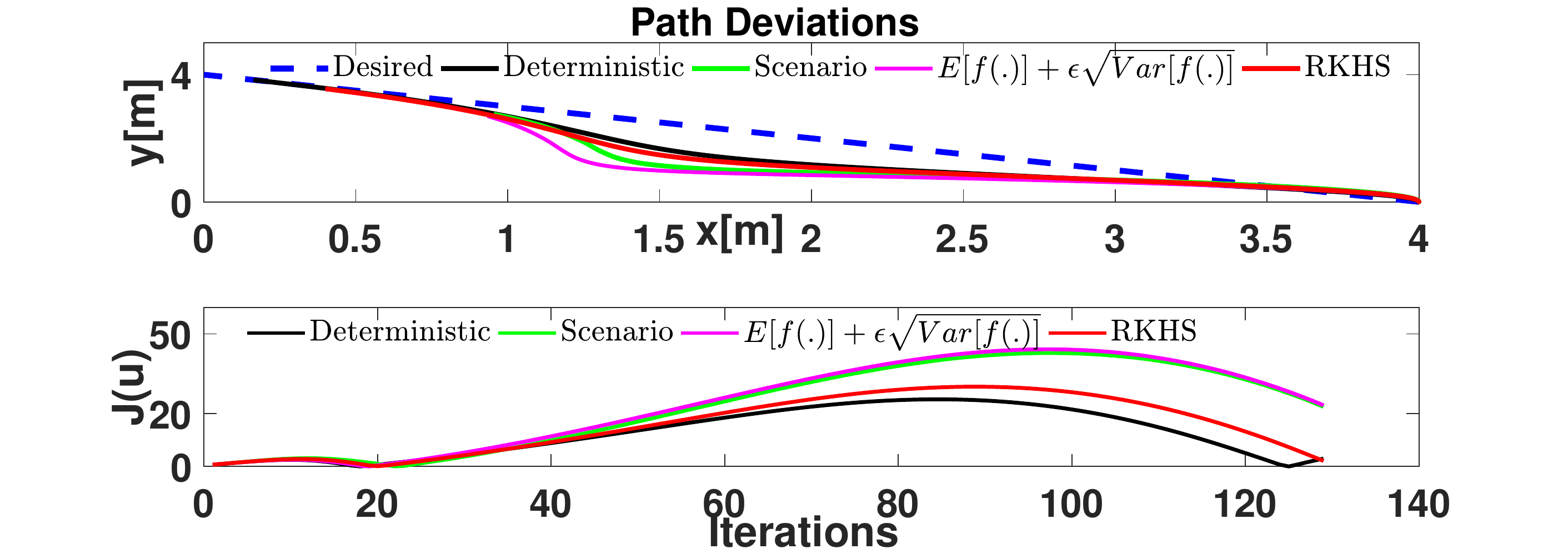}\\
        (a) & (e)  \\ \hline
        \includegraphics[width=8.5cm,height=5.0cm]{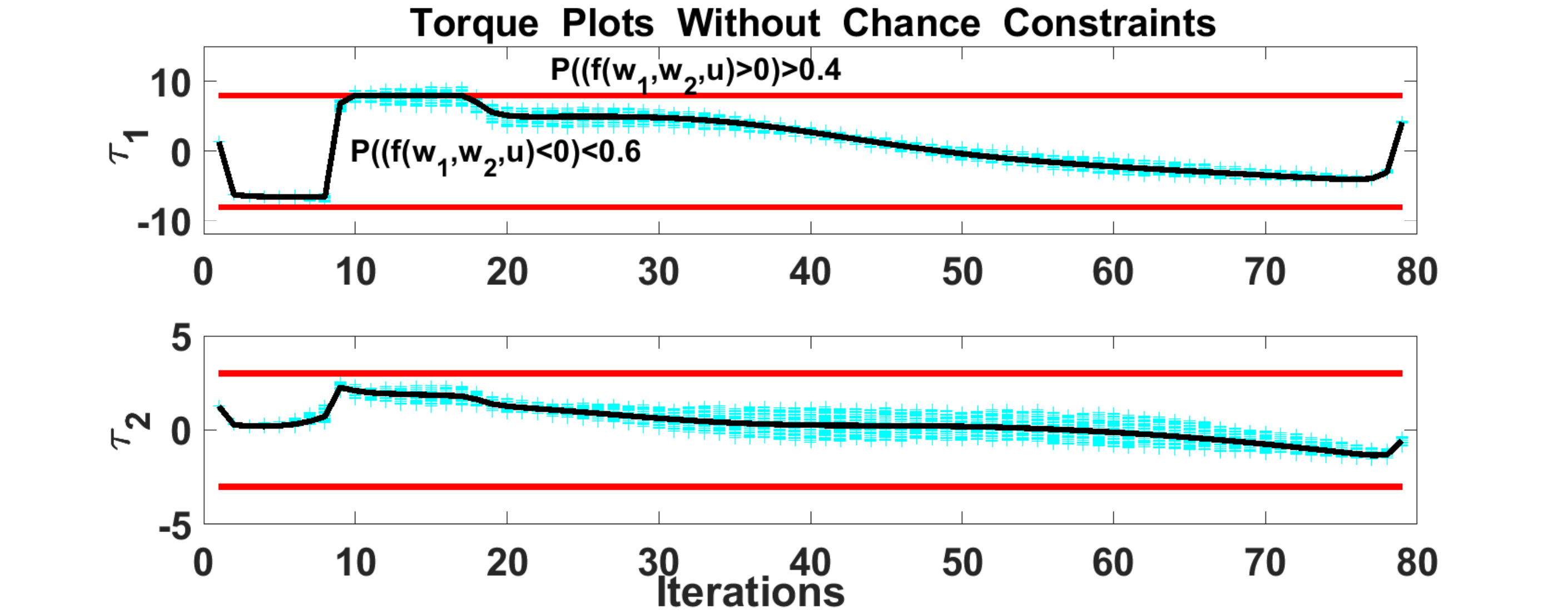}&
        \includegraphics[width=8.5cm,height=5.0cm]{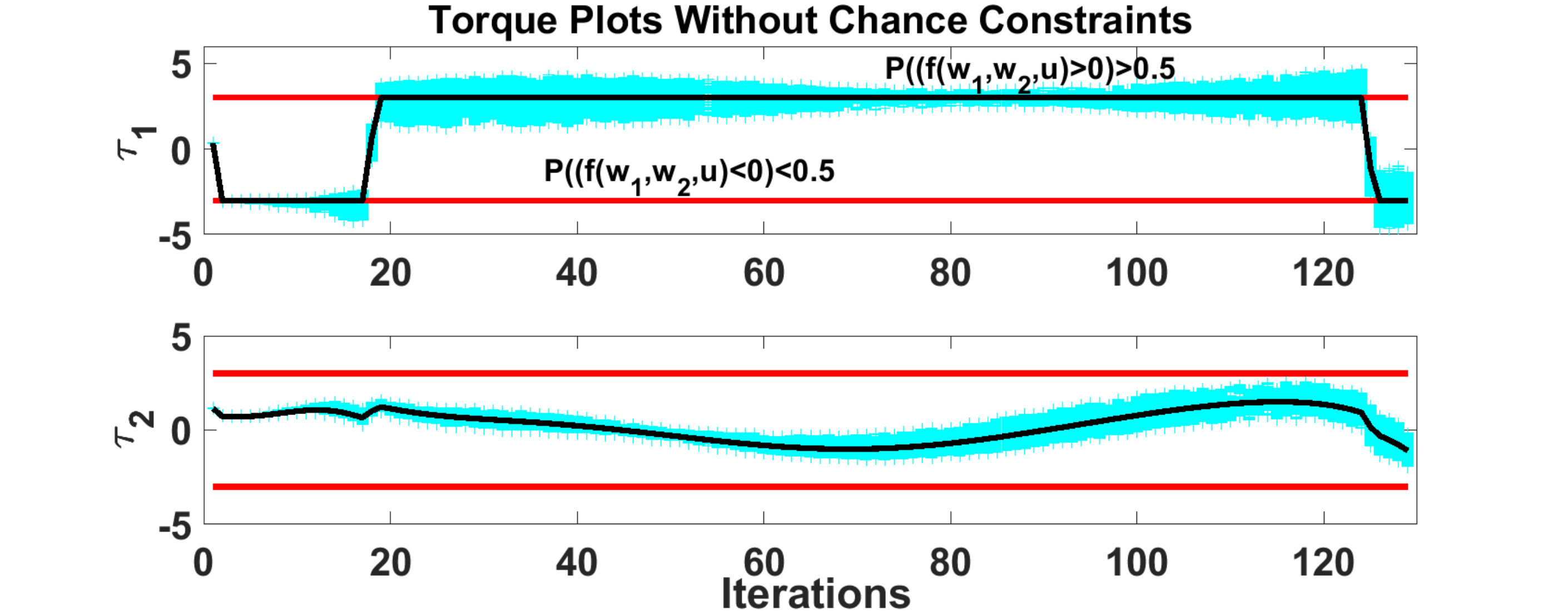}\\
        (b)&(f)\\ \hline
        \includegraphics[width=8.5cm,height=5.0cm]{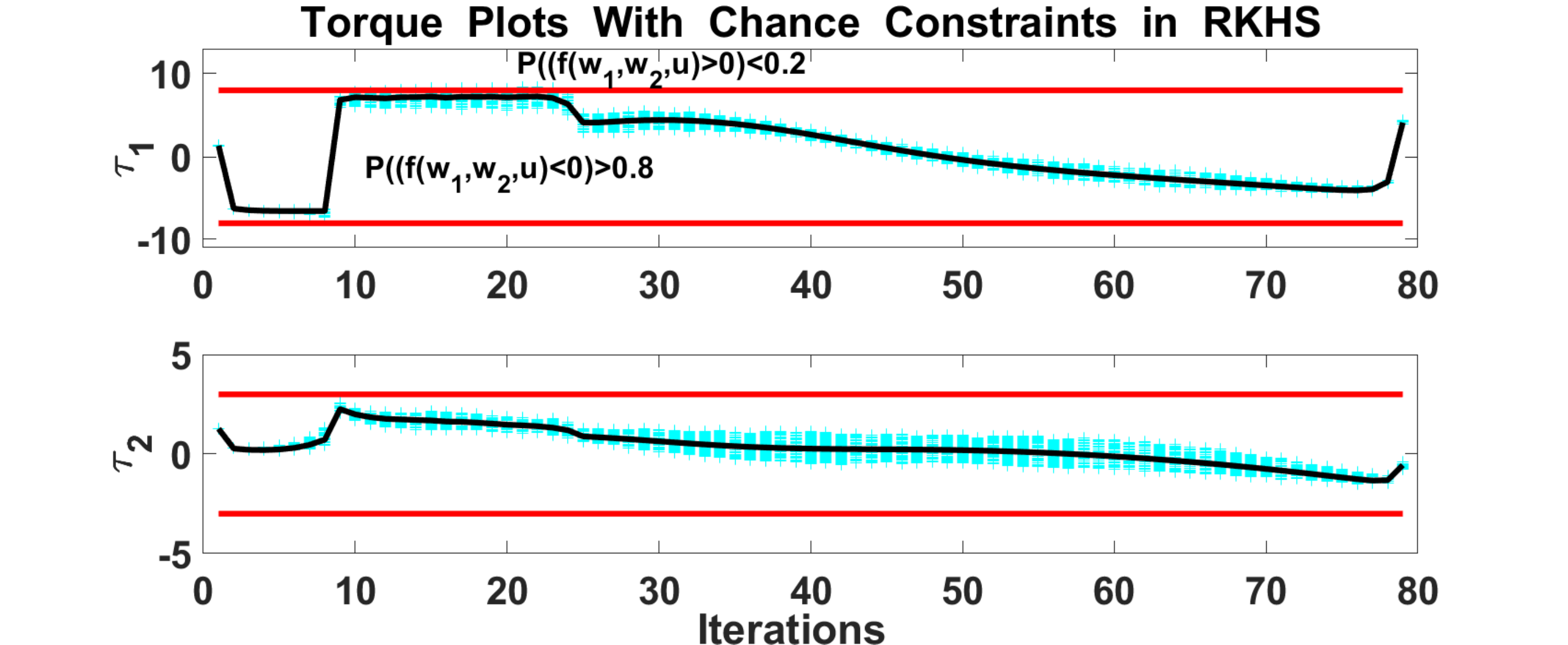}&
        \includegraphics[width=8.5cm,height=5.0cm]{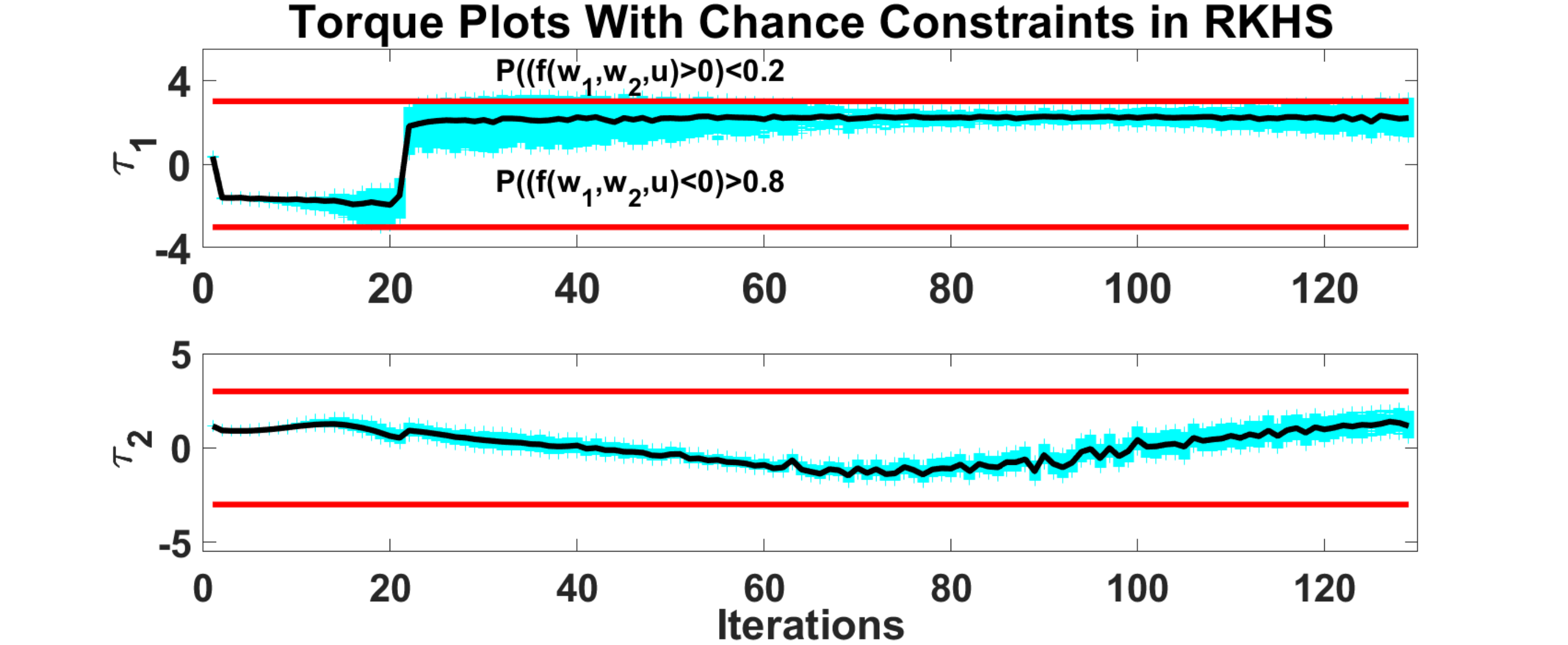}\\
        (c)&(g)\\ \hline
        \includegraphics[width=8.5cm,height=5.0cm]{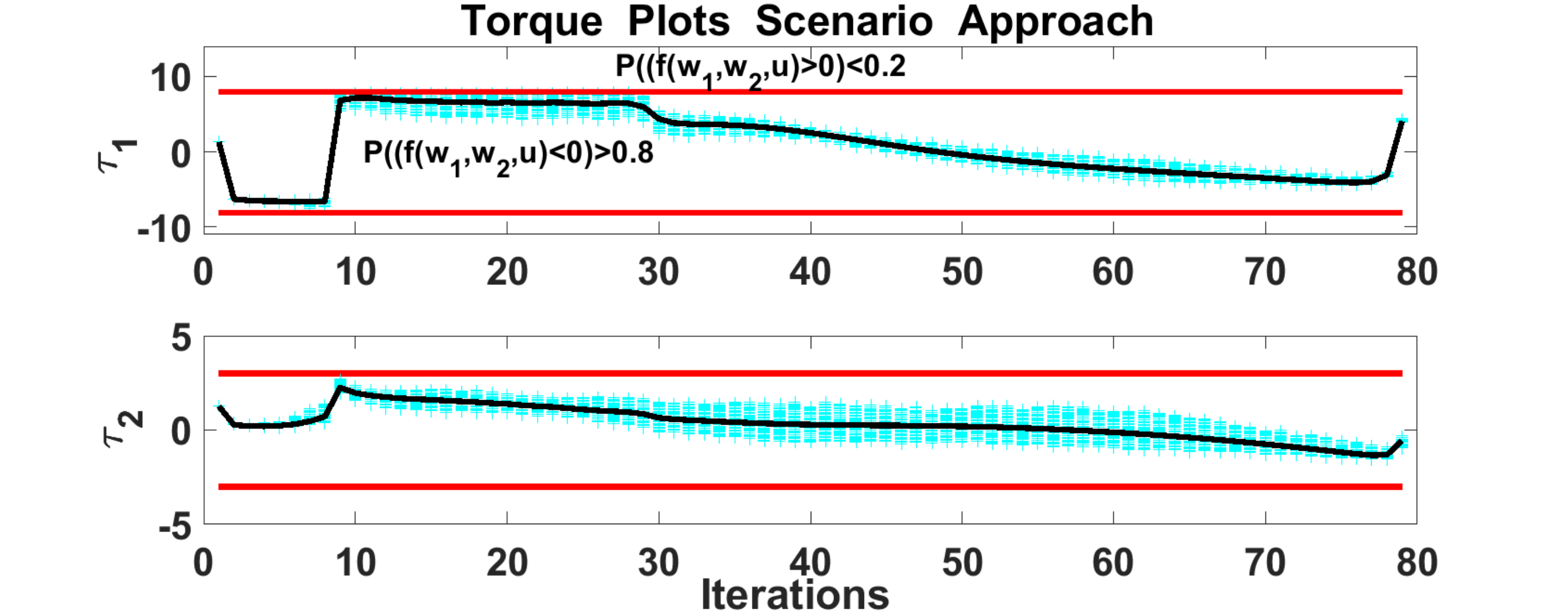}&
        \includegraphics[width=8.5cm,height=5.0cm]{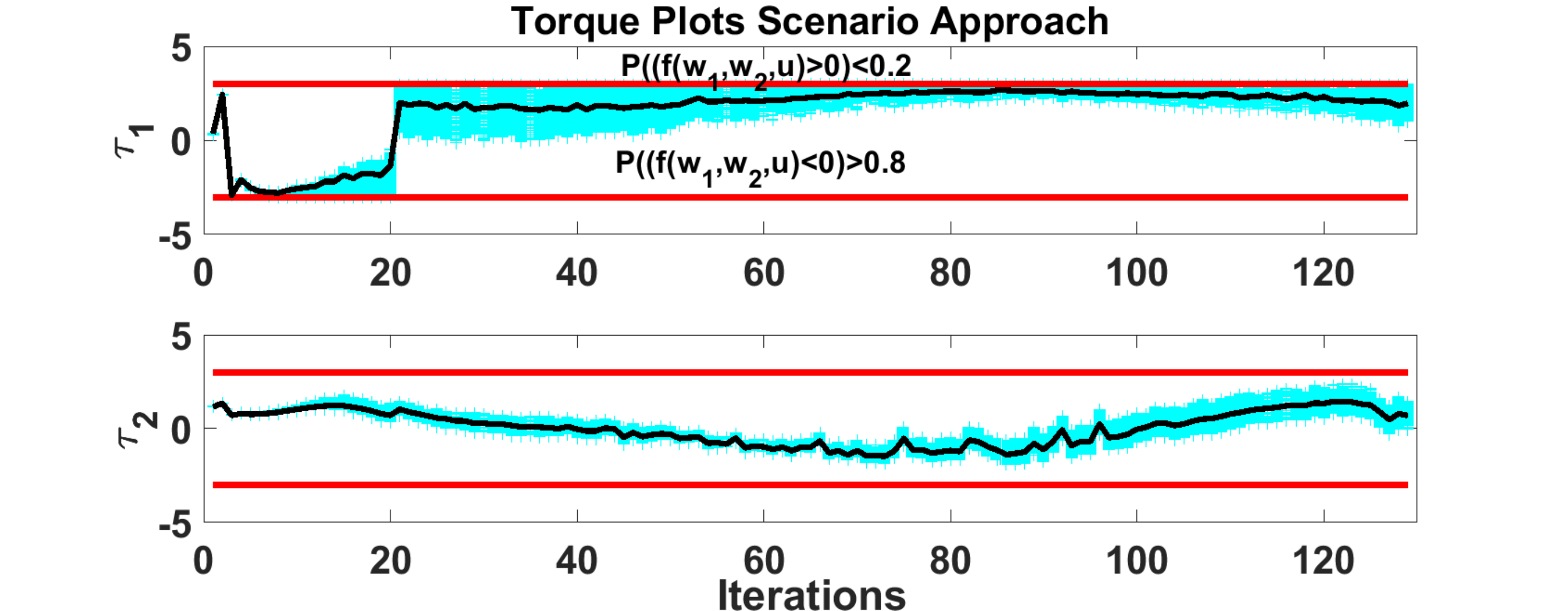}\\
        (d)&(h)\\ \hline
       \end{tabular}
\end{center}
\caption{The comparison between our proposed RKHS based formulation and the scenario approach. The solid black lines in the torque plot represent the mean values while samples colored in cyan show the uncertainty around it. Note that the torque profiles obtained from our formulation remain closer to saturation which in turn directly translates to better tracking performance as depicted in figures (a) and (e).}
\label{manip_comp_results}
\end{figure*}

We now compare our proposed RKHS based formulation with the scenario approach and the approach based on surrogate constraint (\ref{meanvar}) in the context of the path tracking application. \footnote{We do not compare with the SAA approach of \cite{sample_average_shapiro} here because its computational complexity on this application becomes too prohibitive. The collision avoidance application involved only one decision variable and thus, we could do a brute force search to solve the SAA formulated problem. However, such an approach would not be suitable for the path tracking application. Authors in \cite{sample_average_shapiro} suggest a mixed integer reformulation, wherein the number of integer variables would be equal to the number of samples of the uncertain parameters. But, we remark that such a reformulation would be prohibitive for high dimensional robotic systems like manipulators.}. Table \ref{sample_complexity_0.7_manip} and \ref{sample_complexity_0.9_manip} summarizes the sample complexity for $\eta \approx 0.7$ and $\eta \approx 0.9$ respectively for different values of $\tau_{max}$. As can be seen, our RKHS based formulation enjoys better sample complexity than both the compared approaches in this application too. The order of improvement increases with $\eta$. Moreover, at $\eta \approx 0.7$ an additional trend can be observed: the order of improvement also improves as the chance-constrained optimization becomes tighter due to a decrease in $\tau_{max}$. At $\eta \approx 0.9$, the order of improvement remains almost the same for various $\tau_{max}$. It can also be noted that the sample complexity  in this application is significantly lower than that observed in the previous collision avoidance application. We attribute this to the fact that $f_i(.)$ for path tracking application is affine in terms of decision variable (see (\ref{inv_dyn_general_const})) while in collision avoidance application, it is a non-convex quadratic. Fig.\ref{cost_compare_manipulator} shows the comparison of average optimal costs observed across 20 different problem instances. As can be seen, our RKHS based formulation produces significantly lower cost solutions and the order of improvement increases with a decrease in $\tau_{max}$. To provide  more insight, we present one sample comparison in Fig.\ref{manip_comp_results}. Note that the  torque profile obtained through our formulation remains closer to the saturation for more iterations as compared to the torque profiles obtained from the scenario approach. The difference is more pronounced for $\tau_{max} =3$. The higher torque naturally provides more control authority to track the reference path and velocity profiles as verified in the path deviation and cost plots of Fig.\ref{cost_compare_manipulator}.

\begin{table*}[!h]
\centering
\caption{Sample complexity for path trackign application corresponding to $P(f(\textbf{w}_1,\textbf{w}_2,u_1,u_2)<0)\approx 0.7$.}
\begin{tabular}{|p{4cm}|p{4cm}|p{4cm}|p{4cm}|}\hline
Approach & $\tau_{max}=\pm8$ & $\tau_{max}=\pm5$ &  $\tau_{max}=\pm3$   \\ \hline
Scenario & $\textbf{w}_1, \textbf{w}_2= 15$ & $\textbf{w}_1, \textbf{w}_2=25$ & $\textbf{w}_1, \textbf{w}_2=30$   \\ \hline
$E[f(\textbf{w}_1, \textbf{w}_2,u_1,u_2)]+\epsilon \sqrt{Var[f(\textbf{w}_1, \textbf{w}_2,u_1,u_2)]}\leq0$ & $\textbf{w}_1, \textbf{w}_2 = 120$ & $\textbf{w}_1, \textbf{w}_2=120$ & $\textbf{w}_1, \textbf{w}_2 = 120$\\ \hline
 Proposed RKHS formulation   & $\textbf{w}_1, \textbf{w}_2 = 10,  \widetilde{\textbf{w}}_1, \widetilde{\textbf{w}}_2 = 5$ & $\textbf{w}_1, \textbf{w}_2 = 10, \widetilde{\textbf{w}}_1, \widetilde{\textbf{w}}_2 = 5$& $\textbf{w}_1, \textbf{w}_2 = 10, \widetilde{\textbf{w}}_1, \widetilde{\textbf{w}}_2 = 5$ \\ \hline
\end{tabular}
\label{sample_complexity_0.7_manip}
\end{table*}

\begin{table*}[!h]
\centering
\caption{Sample complexity for path trackign application corresponding to $P(f(\textbf{w}_1,\textbf{w}_2,u_1,u_2)<0)\approx 0.9$.}
\begin{tabular}{|p{4cm}|p{4cm}|p{4cm}|p{4cm}|}\hline
Approach & $\tau_{max}=\pm8$ & $\tau_{max}=\pm5$ &  $\tau_{max}=\pm3$   \\ \hline
Scenario & $\textbf{w}_1, \textbf{w}_2 = 30$ & $\textbf{w}_1, \textbf{w}_2 = 40$ & $\textbf{w}_1, \textbf{w}_2 = 50$    \\ \hline
$E[f(\textbf{w}_1, \textbf{w}_2,u_1,u_2)]+\epsilon \sqrt{Var[f(\textbf{w}_1, \textbf{w}_2,u_1,u_2)]}\leq0$ & $\textbf{w}_1, \textbf{w}_2 = 120$ & $\textbf{w}_1, \textbf{w}_2 = 120$ & $\textbf{w}_1, \textbf{w}_2 =120$ \\ \hline
 Proposed RKHS formulation  & $\textbf{w}_1, \textbf{w}_2 = 10, \widetilde{\textbf{w}}_1, \widetilde{\textbf{w}}_2 = 5 $ & $\textbf{w}_1, \textbf{w}_2 = 15, \widetilde{\textbf{w}}_1, \widetilde{\textbf{w}}_2 = 8$& $\textbf{w}_1, \textbf{w}_2 = 20, \widetilde{\textbf{w}}_1, \widetilde{\textbf{w}}_2 = 8$ \\ \hline
\end{tabular}
\label{sample_complexity_0.9_manip}
\end{table*}

\section{Conclusion and Future Work}
Mathematical operations in RKHS, has been the back bone for many of the modern machine learning algorithms. Examples of these span from kernel SVM to Gaussian Process. Recent trends in data science and programming languages widely advocate the use of probabilistic programming. Among the many existing approaches used in probabilistic programming, Hilbert space embedding of distributions has recently gained a lot of popularity. In fact literature along the lines of \cite{scholkopf} even call it as Kernel Probabilistic Programming. We have adopted a series of recent  papers \cite{scholkopf}, \cite{scholkopf2}, \cite{scholkopf3} in this field that describes what a hilbert space embedding of a function of random variables would actually mean. One of the key aspect of our work is connecting the theory of RKHS embedding of distributions to a widely studied problem of chance-constrained optimization, which has applications in both robotics and control. 

We formulated chance-constrained optimization as a problem of matching higher order moments of two distributions. The eventual structure that our formulation takes is that of a non-linear optimization problem, which can be easily solved with the help of off-the-shelf solvers. We validated our formulation on application like dynamic collision avoidance of mobile robots and path tracking of manipulators under torque bounds. Our benchmarking clearly establishes the improvement that our formulation provides over existing approaches in terms of sample complexity and optimal cost.

At the moment, our formulation has some limitations, which we would be looking to rectify in our future works. Firstly, the cost function in our formulation is assumed to be deterministic, i.e they do not contain the uncertain parameters. One simple way of rectifying this would be to formulate stochastic cost as constraints using some slack variables. We are currently evaluating the scalability of this idea. Secondly, we are working on benchmarking our formulation with approaches, which first fits some distribution to the non-parametric uncertainty and then performs the subsequent analysis. Examples of such fitting techniques include include Gaussian Mixture Model, Kernel Density Estimator, Gaussian Process etc. Finally, we are also looking at more complex applications like multi agent navigation, reinforcement learning, etc.


\bibliographystyle{IEEEtran}  
\bibliography{ref} 

\end{document}